\documentclass[10pt,letterpaper]{article}
\usepackage[top=0.85in,left=2.75in,footskip=0.75in]{geometry}

\usepackage{amsmath,amssymb}

\usepackage{changepage}

\usepackage{textcomp,marvosym}

\usepackage{cite}

\usepackage{nameref,hyperref}

\usepackage[nopatch=eqnum]{microtype}
\DisableLigatures[f]{encoding = *, family = * }

\usepackage[table]{xcolor}

\usepackage{array}

\newcolumntype{+}{!{\vrule width 2pt}}

\newlength\savedwidth

\raggedright
\usepackage[aboveskip=1pt,labelfont=bf,labelsep=period,justification=raggedright,singlelinecheck=off]{caption}

\makeatletter
\renewcommand{\@biblabel}[1]{\quad#1.}
\makeatother

\usepackage{lastpage,fancyhdr,graphicx}
\usepackage{epstopdf}
\fancyheadoffset[L]{2.25in}
\fancyfootoffset[L]{2.25in}
\usepackage{multirow}
\usepackage{xcolor,colortbl}
\usepackage{caption}
\usepackage{subcaption}
\usepackage{comment}

\definecolor{APFGood}{HTML}{9AFF99}
\definecolor{APFBad}{HTML}{FFCE93}

\begin{document}
\vspace*{0.2in}

\begin{flushleft}
{\Large
\textbf\newline{APF+: Boosting adaptive-potential function reinforcement learning methods with a W-shaped network for high-dimensional games} 
}
\newline
\\
Yifei Chen\textsuperscript{1,2*},
Lambert Schomaker\textsuperscript{1},
\\
\bigskip
\textbf{1} Bernoulli Institute, University of Groningen, Groningen, The Netherlands
\\
\textbf{2} Data61 of CSIRO, Sydney, NSW, Australia
\\
\bigskip

* yifei.chen@data61.csiro.au

\end{flushleft}
\section*{Abstract}
Studies in reward shaping for reinforcement learning (RL) have flourished in recent years due to its ability to speed up training. 
Based on the potential-based reward shaping framework, our previous work proposed an adaptive potential function (APF)~\cite{Chen2021} that is used as a component module to the regular value estimation in RL algorithms.
There, we showed that APF can accelerate the Q-learning with a Multi-layer Perceptron algorithm in the low-dimensional domain. 
This paper proposes to extend APF with an encoder (APF+) for RL state representation, allowing applying APF to the pixel-based Atari games using a state-encoding method that projects high-dimensional game's pixel frames to low-dimensional embeddings.
We approach by designing the state-representation encoder as a W-shaped network (W-Net), by using which we are able to encode both the background as well as the moving entities in the game frames.
Specifically, the embeddings derived from the pre-trained W-Net consist of two latent vectors: One represents the input state, and the other represents the deviation of the input state's representation from itself.
We then incorporate W-Net into APF to train a downstream Dueling Deep Q-Network (DDQN), obtain the APF-WNet-DDQN, and demonstrate its effectiveness in Atari game-playing tasks.
To evaluate the APF+W-Net module in such high-dimensional tasks, we compare with two types of baseline methods: (i) the basic DDQN; and (ii) two encoder-replaced APF-DDQN methods where we replace W-Net by (a) an unsupervised state representation method called Spatiotemporal Deep Infomax (ST-DIM)~\cite{anand2019unsupervised} and (b) a ground truth state representation provided by the Atari Annotated RAM Interface (ARI)~\cite{anand2019unsupervised}.
The experiment results show that out of $20$ Atari games, APF-WNet-DDQN outperforms DDQN ($14/20$ games) and APF-STDIM-DDQN ($13/20$ games) significantly. In comparison against the APF-ARI-DDQN which employs embeddings directly of the detailed game-internal state information, the APF-WNet-DDQN achieves a comparable performance.

\section*{Introduction}
\label{sec:intro}

Reinforcement learning (RL) has achieved increasing attention in many fields, such as games~\cite{Mnih2015}, aviation~\cite{razzaghi2022,Ruan2021,Lee2022}, and robotics~\cite{openai2019solving,BaierLowenstein2007,millan2021robust}. In the development of RL algorithms, a proper reward signal is an inevitable component to work on. The design of the reward signal is critical and greatly determines the success of the training procedure.

Among the research on reward signals, reward shaping~\cite{mataric1994reward,randlov1998learning} is a practical approach to speed up the RL agent's learning process. However, it may also lead to undesired or unsafe behaviors~\cite{randlov1998learning}.
Therefore, Ng et al.~\cite{ng1999policy} proposed the potential-based reward shaping (PBRS) framework to guarantee policy invariance after reward shaping. 
This means the underlying optimal policy can still be achieved when applying the reward shaping method. In the PBRS framework, a potential function $\Phi$, which can be any function over states, was introduced to compose the reward-shaping function $F$.

Based on the PBRS framework, our former work~\cite{Chen2021} proposed an adaptive potential function (APF) in the sense that the agent learns the potential function concurrently with the policy training, based on the RL agent's experienced trajectories. Intuitively, we aim to extract information from the RL agent's past experiences to shape the reward function. This procedure acknowledges the fact that a reward needs to be perceived by an agent from the current state, and this perception needs to be learned.

While our previous application of APF concerned low-dimensional discrete state-space problems in the Q-learning paradigm, we intend to generalize its use to high-dimensional problems such as benchmark suits of video games.
Ultimately, the APF method can be combined with any RL algorithms in various applications.
However, due to the limitation of computing and data storage, storing many trajectories of high-dimensional states for the RL agent to learn from is not feasible. 
Therefore, this paper proposes the format of APF pluses an encoder, namely APF+, where the encoder projects high-dimensional frames to low-dimensional embeddings as the APF's input.
To effectively encode high-dimensional states, we propose a new state representation method, the W-shaped Network (W-Net), which extracts both static and dynamic features of states.
By incorporating W-Net into the APF component to augment the training of a downstream Dueling Deep Q-Network (DDQN)~\cite{wang16dueling}, we obtain the APF-WNet-DDQN method.

\begin{figure}[tb]
     \includegraphics[width=.9\linewidth]{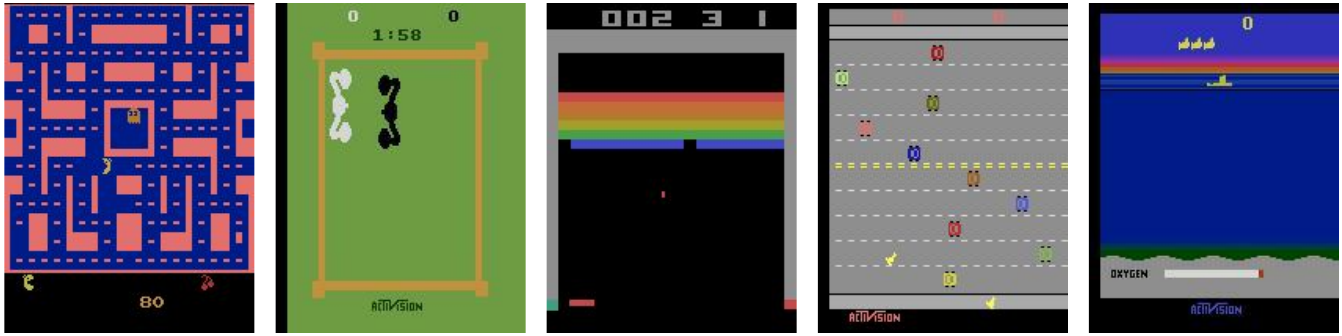}
     \caption{{\bf Screenshots of five Atari games.} From left to right: Ms Pacman, Boxing, Breakout, Freeway, and Seaquest.}
     \label{fig:atari_ss}
\end{figure}

We then empirically study APF-WNet-DDQN in 20 Atari game-playing tasks in the OpenAI Gym environment~\cite{brockman2016openai}.
Fig~\ref{fig:atari_ss} shows the screenshots of five Atari games as examples. 
To evaluate W-Net, we choose two other state representation encoders as our baselines: (1) an unsupervised representation method called Spatiotemporal Deep Infomax (ST-DIM)~\cite{anand2019unsupervised} and (2) the Atari Annotated RAM Interface (ARI)~\cite{anand2019unsupervised} which directly utilizes the ground truth RAM information, and obtain two baseline methods, APF-STDIM-DDQN and APF-ARI-DDQN, respectively.
Additionally, we also compare the APF+ augmented DDQN algorithms with the bare (non-APF) baseline, the DDQN. 
The experimental results show that
APF-WNet-DDQN agent outperforms the DDQN (14/20 games) and the APF-STDIM-DDQN (13/20 games) agents significantly.
Furthermore, the APF-WNet-DDQN agent achieves indifferent performances with the APF-ARI-DDQN agent, which generates embeddings by directly consulting the ground-truth Atari game RAM information.

Note that in this paper, the goal is not to beat the charts as
regards the RL performance on Atari gameplay. We use the Atari 
benchmark to test whether the APF, a specialized transform to improve value estimation in RL, also works in these high-dimensional, pixel-image-based games.
We use the Atari game set as an evaluation method to compare
the viability of game-state representation learning for optimized exploitation
of the APF transform in RL.

{\bf Paper outline.} Section~\nameref{sec:review} reviews related research works on PBRS and state representation methods. Section~\nameref{sec:back} introduces preliminaries of RL, PBRS, and the baseline RL algorithm. Section~\nameref{sec:method} revisits the APF method and describes our proposed W-Net and the APF-WNet-DDQN algorithm in detail. The experimental parameters and results are presented and discussed in Section~\nameref{sec:experiment}. Finally, we present a summary in Section~\nameref{sec:con}.


\section*{Related Work}
\label{sec:review}

In RL, reward shaping is an efficient way of incorporating external advice into a learning agent. Informative shaping rewards can accelerate the learning process of the RL agent. On the contrary, naive rewards like ad hoc rewards would lead to unwanted or unsafe control behaviors. Instead of handcrafted reward shaping, relying on a mechanism where the optimal computation of reward is learned, concurrent to learning the main task is advantageous.
Therefore, the PBRS framework, which guarantees policy invariance by using the difference in subsequent values of a potential function \cite{ng1999policy}, has shown to be effective in 
authonomous RL reward shaping.

Many RL reward-shaping research works are based on the PBRS framework. \cite{harutyunyan2015shaping} first demonstrated how to apply human feedback in reward shaping effectively based on PBRS. 
In \cite{Badnava2023}, the authors reinforced the reward signal under the PBRS framework by tracking the maximum and minimum episodic rewards.
\cite{Eck2015} utilized PBRS for online partially observable Markov decision processes to improve approximate planning.
Moreover, the PBRS framework is also helpful for reward shaping in multi-agent RL scenarios in a way that it reserves the Nash Equilibrium of multi-agent systems~\cite{marthi2007automatic,devlin2012dynamic,babes2008social,devlin2011empirical}.

\paragraph{Potential based reward shaping}
Following the PBRS, \cite{wiewiora2003principled} extended the potential function from the state domain to the state-action domain. In addition, they proved that using a static potential function is equivalent to initializing Q-values to the static potential function in a discrete domain. 

A concurrently learned, adaptive potential function will be more practical than using an a priori-defined static potential function. This is due to especially the problems of complex environments, where not enough is known about optimal reward shaping.
Though it is not guaranteed to speed up the learning process of an RL agent, the use of PBRS was demonstrated to warrant the convergence of the learning process even if the potential function is dynamic (i.e., time-variant) or misleading~\cite{devlin2012dynamic}. This result laid the groundwork for the idea of concurrently learning a potential function while learning a policy.

\paragraph{State representation in RL}
A major question to be answered concerns a state representation method that can capture the essence of high-dimensional state variables in complex and challenging RL scenarios. There are many studies on this~\cite{watter2015embed,lange2010deep,bohmer2015autonomous}. \cite{lesort2018state} summarizes various state representation learning approaches and evaluations.
In \cite{anand2019unsupervised}, the authors proposed a self-supervised state representation learning method called Spatiotemporal Deep Infomax (ST-DIM) that focuses on learning representations from spatiotemporal data. The ST-DIM model learns state representations by maximizing the mutual information across consecutive observations. To evaluate the representation, the authors implemented an easy-to-use gym wrapper called Atari Annotated RAM Interface (ARI) which can output a state label for each game image by consulting the source code.
In \cite{higgins2018towards}, the authors provided a theoretical framework of learning disentangled representations that separate the underlying factors of variation in the data. However, prior knowledge about the state space structure is needed here to balance the poor exploration by the agent which is caused by disentangled (i.e., isolated) representations.
Recently, \cite{wu2023read} demonstrated an unsupervised solution for grounding objects in Atari video games. However, due to the complicated game dynamics of Atari environments, the proposed solution intentionally overfitted to a single game, i.e., the Skiing scenario. In the current study, we will address a wide range of games.

\section*{Background}
\label{sec:back}

We first introduce the necessary RL background knowledge for our method, namely, the Markov decision process, the PBRS, and the DDQN.

\subsection*{Markov Decision Process and Q-values}
\label{subsec:mdp}

The RL problem can be mathematically formalized by the Markov Decision Process (MDP), which is defined as a five-element tuple $M = <S,A,P,\gamma,R>$, where $S$ is a set of environmental states, and $A$ is a set of actions that the agent can take.
The transition probability $P$ quantifies the probability that the agent transfers to state $s'$ when it takes action $a$ at state $s$ for each triple $s'\in S$, $a\in A$, and $s\in S$, and we denote this probability as $P(s'\mid s,a)$.
The environmental function $R: S \times A \times S \rightarrow \mathbb{R}$ returns a reward when the agent takes an action at a state and transfers to the next state.

The goal of an RL agent is to learn an optimal policy that maximizes the expected cumulative future rewards.
A policy $\pi: S \rightarrow A$ is a mapping from states to actions, depicting the RL agent's behavior, i.e., which action the agent should take at each state.
We call a policy deterministic if it guides the agent to choose a deterministic action on each state.
Eq~\ref{eq:vfunc} is the state-value function (a.k.a. V-function) which outputs the expected discounted sum of future rewards that the agent can obtain, taking state $s$.

\begin{eqnarray}\label{eq:vfunc}
    V(s) \doteq \mathbb{E}[\sum_{k=0}^{\infty} \gamma^k R_{t+k+1} \mid S_t=s]
\end{eqnarray}

Similarly, Eq~\ref{eq:qfunc} is the action-value function (a.k.a. Q-function) which specifies the expected discounted sum of future rewards when the agent takes an action $a$ at state $s$.

\begin{eqnarray}\label{eq:qfunc}
    Q(s,a) \doteq \mathbb{E}[\sum_{k=0}^{\infty} \gamma^k R_{t+k+1} \mid S_t=s, A_t=a]
\end{eqnarray}

In this paper, we study value-based RL agents, i.e., agents select actions based on action-value judgments (e.g., the Q-function).
Specifically, When the agent learns the optimal Q-function $Q^{\ast}(s,a) \doteq \max Q(s,a)$, the optimal policy is derived as $\pi^{\ast}(s) = \arg \max_a Q^{\ast}(s,a)$.

\subsection*{Potential-Based Reward Shaping}


Reward shaping is an efficient way to incorporate extra information in RL agents.
As described in the subsection~\nameref{subsec:mdp}, the original MDP is represented as $M = <S,A,P,\gamma,R>$. 
By shaping the environmental reward function $R$ with a reward shaping function $F$, the original MDP $M$ is transformed to $M' = <S,A,P,\gamma,R'>$, where $R'=R+F$.

Our work is theoretically based on the PBRS framework proposed in~\cite{ng1999policy}, which guarantees the invariance of the learned optimal policy against reward shaping.
In the PBRS framework, the reward-shaping function $F: S \times S \rightarrow \mathbb{R}$ is defined in Eq~\ref{eq:pbrs}. 

\begin{eqnarray}\label{eq:pbrs}
    F(s,s') = \gamma \Phi (s') - \Phi (s),
\end{eqnarray}

\noindent where $\gamma$ is the same discount factor as in MDP. $\Phi$ is the potential function that takes a state as input and outputs the potential value of the state.

In this paper, we aim at enabling the RL agent to learn an efficacious potential function based on the information collected during training.

\subsection*{Dueling Deep Q-Network}

\begin{figure}[!h]
     \includegraphics[width=.7\linewidth]{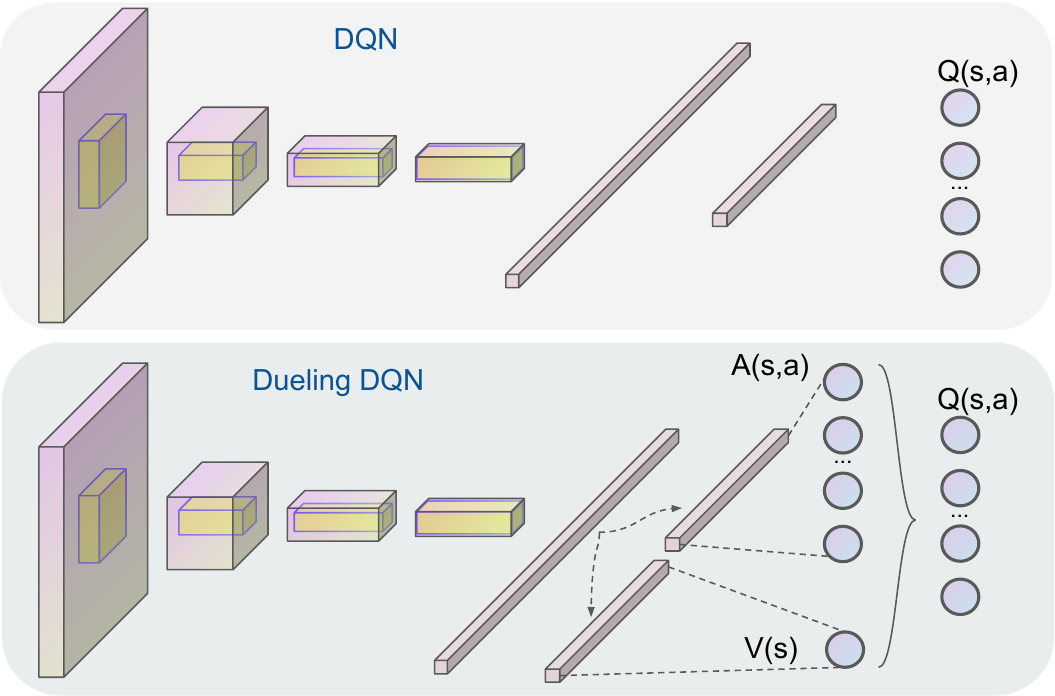}
     \caption{{\bf An illustration of DQN and DDQN.}}
     \label{fig:dqn_vs_ddqn}
\end{figure}

DDQN \cite{wang16dueling} is an off-policy RL algorithm, specializing in differentiating state/action quality. As illustrated in Fig~\ref{fig:dqn_vs_ddqn}, DDQN combines the dueling architecture and the well-known RL algorithm Deep Q-Network (DQN) \cite{Mnih2015}. 
Specifically, DDQN simultaneously learns a V-function and an action advantage function (a.k.a A-function) and aggregates the two streams to estimate the Q-function, instead of directly learning a Q-function as the DQN.

The A-function ($A$) is defined in Eq~\ref{eq:adv}, which intuitively represents the advantage of choosing action $a$ compared to other actions at state $s$.

\begin{eqnarray}\label{eq:adv}
    A(s,a) = Q(s,a) - V(s)
\end{eqnarray}

However, training the dueling network by simply adding the V-function and the A-function is problematic due to the identifiability issue. In other words, the V-value and A-value can't be recovered uniquely given a Q value. Therefore, as suggested in~\cite{wang16dueling}, we use Eq~\ref{eq:ddqn} to implement the forward mapping of the last module of DDQN.

\begin{eqnarray}\label{eq:ddqn}
    Q(s,a) = V(s) + \left(A(s,a) - \frac{1}{|A|}\sum_{a'}A(s,a')\right)
\end{eqnarray}

In this paper, we choose DDQN as our baseline algorithm because DDQN generates better policy evaluation in the presence of many similar-valued actions. This is important for environments with large action spaces, e.g., the Atari game environments used in this paper (with up to $18$ actions).




\section*{Methodology}
\label{sec:method}
Now, we are ready to introduce our method.
In our method, the RL agent learns the optimal policy while updating the APF, to enhance the learning efficiency and accuracy.
To achieve this, the APF network is updated simultaneously while learning a DDQN network.
As follows, we first introduce one of the key parts of our methods, namely the APF updating process.
Then, we show the core state representation technique, W-Net, with which we can handle high-dimensional RL tasks.
Finally, we illustrate how we combine the APF+ method with the DDQN network, with the help of W-Net, to form a novel RL algorithm.

\subsection*{Revisit Adaptive Potential Function}\label{sec:cnt_pf}

This subsection revisits the simple and effective adaptive potential function (APF) proposed in~\cite{Chen2021}.
Specifically, the APF is learned based on discriminating good and bad states by analyzing how often each state is visited by the RL agent in good and bad trajectories, respectively. 
We detail the APF learning process as follows.

We first formally define trajectories during training.
A trajectory $t_i$ is a sequence of states that are visited in the $i$-th episode, and the ranking of states is decided by the time that each state is visited.
Note that there may exist replicated states in a trajectory.
We denote the set of all trajectories during the training as $T$.
Then, each trajectory is stored in a trajectory replay buffer, denoted as $D_{traj}$, which is implemented by a priority queue.
Trajectories' priority in $D_{traj}$ is decided by each trajectory $t_i$'s episodic reward $R_i$, i.e., the un-discounted sum of rewards collected in episode $i$.

In $D_{traj}$, we define the set of good trajectories $T_g \subset T$ as the best $20\%$ among all trajectories in $T$, and the set of bad trajectories $T_b=T\setminus T_g$ as the rest of trajectories.

The calculation of the potential value of each state is as follows.
First, a batch size of trajectories, among which half are from $T_g$ and the other half are from $T_b$, is randomly sampled periodically during training.
Let $T^s_g$ be the state sequence that concatenates all sampled good trajectories, and $T_b^s$ be that of all sampled bad trajectories.
For each state $s$ included in $T^s_g$ or $T^s_b$, we denote its occurrence in $T^s_g$ as $N_g$, and its occurrence in $T^s_b$ is denoted as $N_g$.
Then, the potential value of state $s$, $P(s)$, is calculated as follows

\begin{eqnarray}\label{eq:cnt}
    P(s) = \frac{N_{g}[s] - N_{b}[s]}{N_{g}[s] + N_{b}[s]}.
\end{eqnarray}

Note that if a state is not included in any sampled trajectory, its potential value cannot be computed in the above method.
Note also that if a state has not been visited, its potential value cannot be computed either.
However, such problems can be mitigated by updating the potential value by a neural network.
Observe that, in Eq~\ref{eq:cnt}, the range of $P(s)$ is between -1 and 1.

In order to generalize the APF to unvisited and unsampled states, to tackle high-dimensional tasks, we use an APF neural network $\Phi$ which is implemented by a multi-layer perceptron to update all states' potential values.
We train the APF neural network by minimizing the loss function as follows:

\begin{eqnarray}\label{eq:loss_pf}
L_{\phi} = \mathbb{E}[(P(s) - \Phi(s; \phi))^2],
\end{eqnarray}

\noindent where $\Phi$ is a parametric function taking any state $s\in S$ in the environment.

Unlike the value and advantage evaluations in DDQN networks, which address the immediate state, action, and value per time step, the potential function is based on the expected potential predicted from a collection of complete states encountered doing trajectory $T_i$ with states {$s_1, s_2, ..., s_k$}.

In summary, the intuition of the APF method is to count the number of the sampled good and bad states in each APF update.
The counting is used to shape the potentials and consequently guide the agent to more desirable states.
Our designed APF guides the agent to follow the most successful trajectory history.

\subsection*{W-Net: Representation Using State and State-Expectancy Deviation}\label{sec:wnet}

In environments with a small-scale and low-dimensional state space, computing potential values using Eq~\ref{eq:cnt} is easy, such as using lookup tables. 
However, this is not feasible in large-scale and high-dimensional state spaces for two reasons: First, storing and counting high-dimensional states is memory-consuming and impractical. Second, most states in high-dimensional environments only occur once. Therefore, an efficient encoding method is needed to build a mapping from the high-dimensional space to a lower-dimensional space (an `embedding') which allows for a better generalization by capturing state-space neighborhoods.
It is reasonable to assume that an agent has the (perceptual/cognitive) ability to perform
this computation. However, this also requires that the agent collects perceptual experiences for the target task.

Upon trying several dimensionality-reduction networks (e.g., U-Nets) for the perceptual state in our preparatory research, we were only able to reconstruct the input pictures inaccurately through the embeddings.
The representations lacked an enhancement of small but highly informative changes or events in
the video sequences.
Consider that both the regular components of the state and the surprising components, e.g., the appearance of a ghost, are important for the estimation of the potential for reward given the current state.
We design a novel W-shaped Network (W-Net), a variant based on the U-Net structure, to encode high-dimensional states and represent deviations from the expected, at the same time. 

Compared to the classic U-Net structure~\cite{ronneberger2015u}, we make two changes to obtain our W-Net as follows:
\begin{enumerate}
\item Remove the skip layers. Instead of masking input images, we encode them using the bottom layer. Transferring features through connected skip layers acts as a bypass, preventing proper learning of an embedding in the bottom bottleneck layer. Therefore, we remove all skip layers in W-Net.
Note that in the following contents, without specification, we still call U-Nets for such skip layer removed U-Nets, but the term should be distinguished from the U-Nets in \cite{ronneberger2015u}.
\item Connect two U-Nets. We found that a U-Net alone is not useful because it represents the average expected state, which is not very informative for either value estimation or state prediction.
This also holds in the temporal difference case, where the next state at $t+1$ is predicted.
Therefore, a second U-Net is needed to zoom in on the details and unexpected events in the frame sequence. 
While the first U-Net can reconstruct the average parts among the game frames, the second U-Net captures unexpected parts, such as the moving player and ghosts.
\end{enumerate}

\begin{figure}[!h]
     \includegraphics[width=.9\linewidth]{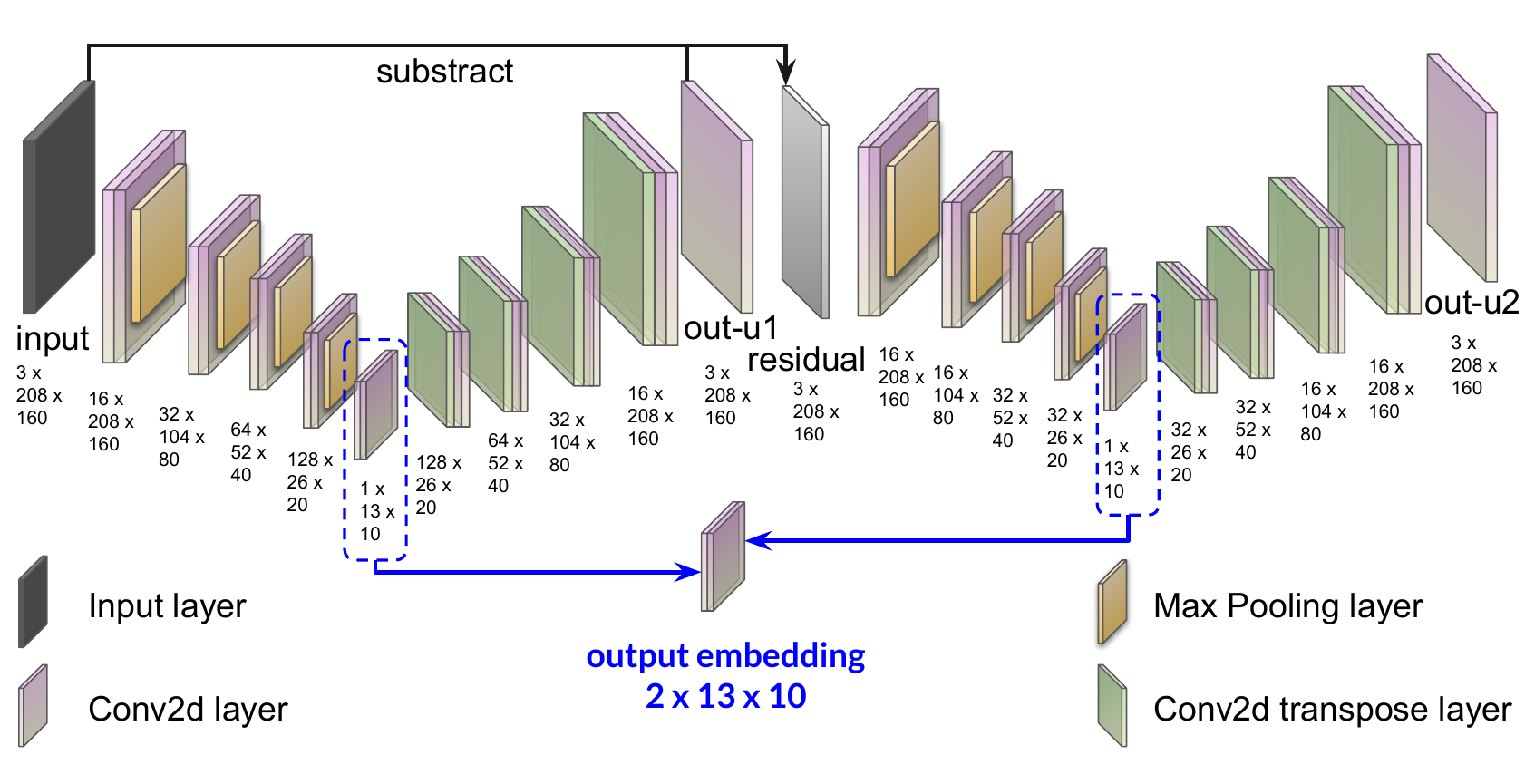}
     \caption{{\bf An illustration of the W-Net architecture.} In this architecture, the game frame is captured by two bottleneck embeddings: The current perceptual state and the difference, i.e., the residual, between the expected and current state, both adjoined into a $2 \times 13 \times 10$ feature map or output embedding, shown at the bottom, center.}
     \label{fig:residual_unet}
\end{figure}

As shown in Fig~\ref{fig:residual_unet}, W-Net consists of two U-Nets connected end to end.
Specifically, the first U-Net takes game frames as input, and its output is termed \texttt{out-u1} layer, as shown in the figure. 
Then, we define a \texttt{residual} layer, obtained by subtracting the \texttt{out-u1} layer from the input layer, as the input of the second U-Net. 
This operation enhances the occurrence of surprising states, given the current expected state. The output of the second U-Net is termed \texttt{out-u2}.
Therefore, the first U-Net learns to encode the expected state while the second U-Net learns to encode the residual parts.
The combination of the expected state and the deviation from the expected state will be used as the basis for learning the potential function. 
This is achieved by adjoining the two bottom layers of the two U-Nets, and we then represent the game image as a $2 \times 13 \times 10$ embedding, i.e., yielding $260$ dimensions.


Note that the W-Net is pre-trained for each target environment (an Atari game) before training the RL agent (the APF-WNet-DDQN agent in this paper). We show the visualization of W-Net's representation ability in Atari game Ms Pacman in Section~\nameref{sec:experiment}, and provide the visualizations of more Atari games in~\nameref{app:fig_wnets}.

\subsection*{Model Architecture}\label{sec:exp_dqn}

Combining an APF with a state representation encoder, we have the APF+ method.
To test APF+, we apply it to train a baseline RL agent to form the APF-X-Y algorithm, where X represents the state encoder and Y is a downstream RL agent.
We specifically propose APF-WNet-DDQN as follows.

In particular, we can see the APF-WNet-DDQN algorithm as the combination of three modules: (1) the DDQN module, (2) the W-Net module, and (3) the APF network $\Phi$ module. 
To describe the APF-WNet-DDQN algorithm clearly, we first depict the input for each component module of APF-WNet-DDQN. Then, we describe the workflow of APF-WNet-DDQN.

\subsubsection*{Input Spaces of APF, W-Net, and DDQN}
We clarify the input of each component module of APF-WNet-DDQN in the scenario of video game playing by which we evaluate our method.
In the video game setting, at each time point $t$, we fix the colored game image (practically, a $3\times 208 \times 160$ matrix in Atari games) as a {\em game frame}, denoted as $g_t$.
Let $G$ denote the game frame space. W-Net is trained on space $G$.

We then define the corresponding environmental state $s_t\in S$ as a stack of the latest four preprocessed game frames at time $t$, i.e., $s_t=(Pre(g_{t-3}),Pre(g_{t-2}),Pre(g_{t-1}),Pre(g_{t}))$, to capture the video motion information \cite{Mnih2015}.
The function $Pre$ takes a game frame as input and produces a resized and grayscale image (we particularly resize each game frame into dimensions of $84\times 84$ in experiments).
The environmental state $s \in S$ is collected in a replay buffer to train the DDQN networks.

To train the APF network, we use W-Net to project the game frames to embeddings.
Formally, let $z=W(g)$ be the embedding of game frame $g$, and we input $z$ into the APF network, where $W$ takes a game frame $g\in G$ and outputs the embedding $z$ encoded by the W-Net.
Let $Z$ denote all embeddings output by the W-Net taking each state $g\in G$.
Note that $|Z|\le |G|$, since multiple states may be encoded as the same embedding.

In summary, we use the environmental state space $S$ to train the DDQN, the game frame space $G$ to train the W-Net, and the embedding space $Z$ to train the APF network.

\subsubsection*{The APF-WNet-DDQN Framework}

\begin{figure}[!h]
     \includegraphics[width=.9\linewidth]{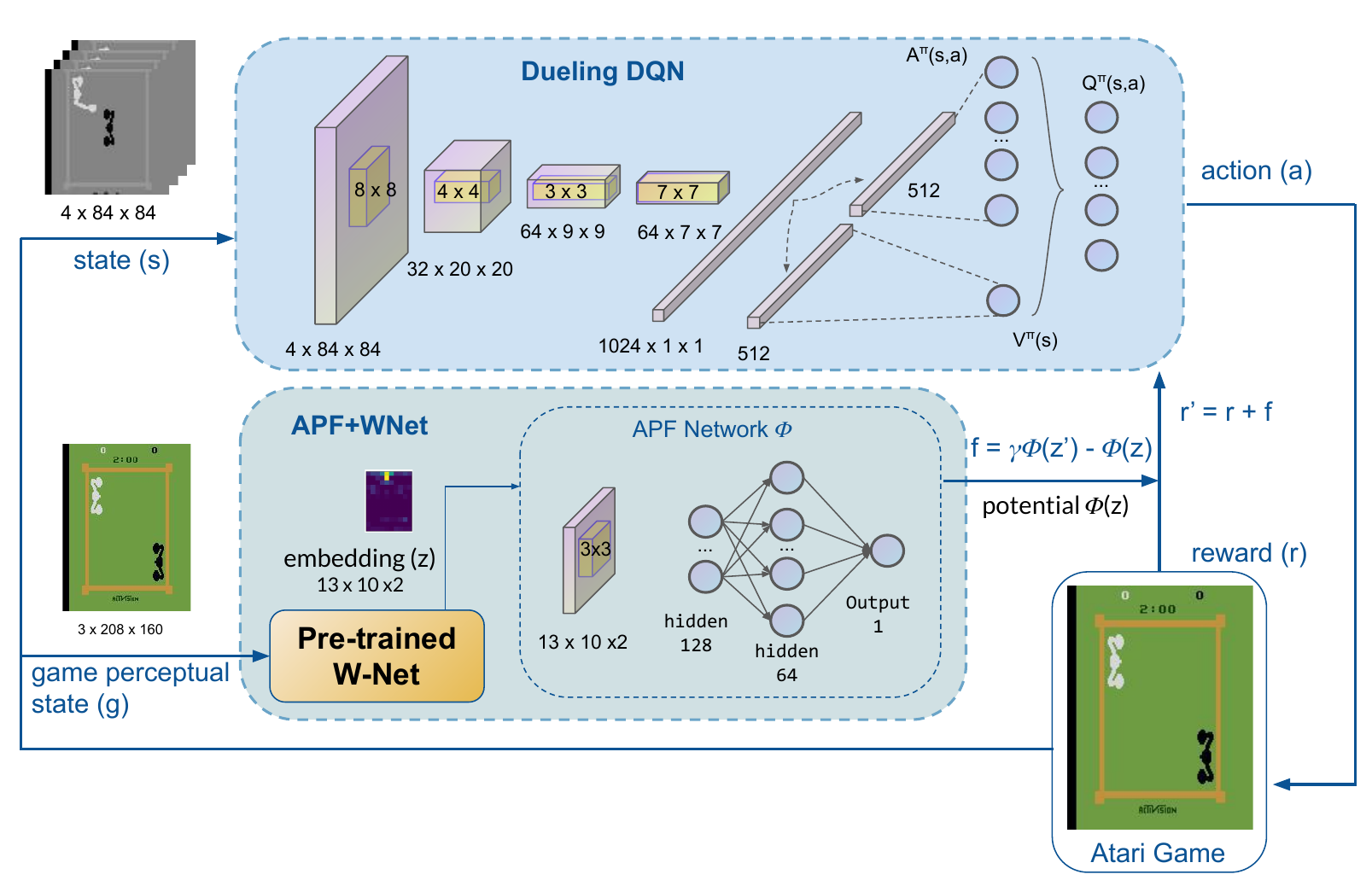}
     \caption{{\bf The model architecture of the APF-WNet-DDQN algorithm.} The W-Net is used to map game frames to embeddings. The APF network is trained on embedding space and learns to output potential values for shaping rewards. The downstream DDQN is trained on rewards shaped by the APF.}
     \label{fig:pf_dqn}
\end{figure}

This subsection introduces the workflow of the APF-WNet-DDQN algorithm.
As illustrated in Fig~\ref{fig:pf_dqn}, the APF-WNet-DDQN algorithm inferences the action $a$ taken at the state $s$ based on the $Q$ network of the DDQN module. 
Compared to the baseline DDQN, APF-WNet-DDQN updates the $Q$ network using rewards shaped by the APF network. In other words, APF-WNet-DDQN learns action values based on the shaped reward function $R' = R+F$ instead of the environmental reward function $R$.

To obtain the reward shaping function $F(s', s)=\gamma\Phi(s') - \Phi(s)$, the APF network $\Phi$ is concurrently trained with the $Q$ networks and learns to estimate potential values based on the agent's past experiences.
In particular, the APF network ($\Phi$) takes the embedding $z$ encoded by the W-Net (i.e., $z=W(g)$) as input, instead of directly using the game frame $g$, to achieve efficient computation of states' potential values, i.e., $\Phi(W(g))$.
Therefore, the APF network ($\Phi$) is trained in this embedding space $Z$ to obtain the shaped reward function $R'$ as Eq~\ref{eq:rf}.

\begin{eqnarray}\label{eq:rf}
\begin{aligned}
R'(s,a,s') = & R(s,a,s')+F(s,s')\\
= & R(s,a,s')+(\gamma\Phi(z'=W(g'))-\Phi(z=W(g))).
\end{aligned}
\end{eqnarray}

In summary, with the representation ability of the W-Net, it projects game frames to embeddings.
Then the APF learns an effective potential function to shape the environmental rewards in the embedding space.
Eventually, with the shaped reward function $R'$, the downstream DDQN converges faster to the optimal policy. 


\section*{Experiments}
\label{sec:experiment}
Now, we empirically show the efficacy of APF-WNet-DDQN in high-dimensional-state environments, specifically, in $20$ Atari 2600 video games.
We select these 20 games from the 22 games analyzed in the following baseline methods ST-DIM and ARI~\cite{anand2019unsupervised}, with the removal of two well-known hard exploration games Pitfall and Montezuma’s Revenge~\cite{ecoffet2021first,aytar2018playing}, in which APF-WNet-DDQN and all baseline methods show failure in training an effective policy.
The experiments are conducted in the environment provided by OpenAI Gym~\cite{brockman2016openai}.
To systematically evaluate the performance of our proposed APF+W-Net method in the DDQN framework, we compare it (APF-WNet-DDQN) with the following baseline methods:

\begin{enumerate}
\item The well-known baseline Dueling Deep Q-Network (DDQN~{\cite{wang16dueling}}) alone;
\item The APF-augmented DDQN that uses the existing ST-DIM~\cite{anand2019unsupervised} method as the game-state encoder (APF-STDIM-DDQN);
\item The APF-augmented DDQN that uses ARI~\cite{anand2019unsupervised} as the encoder (APF-ARI-DDQN), where the ARI generates state labels (embeddings) by consulting the ground-truth ALE RAM information from game-state internals.
\end{enumerate}

The expectation is that using APF, in general, improves on the bare DDQN performance. For the other methods, it is expected that ARI benefits from the fact that handcrafted game internals are used. In this paper, we want to demonstrate, that a usable game state can also be computed from the pixel-frame sequence using the proposed W-Net encoder.

\subsection*{Baseline state representation methods}
In the experiment, we compare our method, i.e., APF-WNet-DDQN, with three baseline methods, namely, the basic DDQN and two APF+ augmented methods called APF-STDIM-DDQN and APF-ARI-DDQN.

We compare the APF-WNet-DDQN to the basic DDQN to show the efficacy of employing the APF reward shaping method with the W-Net state representation network.

The APF-STDIM-DDQN method uses the same framework as APF-WNet-DDQN, with a replaced state representation network called ST-DIM \cite{anand2019unsupervised}.
ST-DIM is a state representation learning framework that tries to capture the key task features, e.g., the scores, and agents' and enemies' positions in Atari games, in both a local (convolutional) feature map and a dense latent space, by maximizing the mutual information of temporarily and spatially interconnected states.
It should be noted that both APF-STDIM-DDQN and APF-WNet-DDQN are realized by 'plugin' swapping the methods within exactly the identical APF-X-DDQN code framework: The pre-trained ST-DIM network can be directly exchanged with a pre-trained W-Net network for a performance comparison (see Fig. \ref{fig:pf_dqn}).

Notably, to further illustrate the efficacy of APF-WNet-DDQN, we also compare it with an APF-DDQN framework, replacing the state representation network (e.g., the W-Net) with the ground truth state information provided by the Atari Annotated RAM Interface (ARI) \cite{anand2019unsupervised}.
ARI is an interface extracting the underlying true game information of each game frame from the RAM in the Openai-gym environment.
It provides the true game features, such as the score, and agents' and enemies' positions, in each game frame.
We obtain the APF-ARI-DDQN framework by replacing the state encoder with the ARI interface.
Note that we do not use all game features provided by ARI, and instead, we select a subset of the features (see details in Appendix~\nameref{app:tab_arilabels}) for the APF-ARI-DDQN to achieve its best performance in our evaluation.

By comparing APF-WNet-DDQN with APF-ARI-DDQN, we hope to gain insights into how close the W-Net's (pixel-based) performance will be in capturing the key game information to the ground truth from the game internals.

\subsection*{Experimental Parameters}

For each Atari game, we repeatedly train an RL agent for $5$ times with different random seeds, and proceed each training with $30$ million game frames.
We then measure each agent's performance by the average of these $5$ repeated training results.

{\bf W-Net and ST-DIM Parameters:}
We use a random policy agent to collect $30000$ game frames for each Atari game and pre-train the W-Net on those game frames for $100$ epochs. A batch size of $64$ game frames is sampled in each update. 
There are $890676$ trainable parameters in the W-Net model and the parameters of the W-Net are fixed during the training of the APF-WNet-DDQN agent. 
The output embedding used for the perceptual state only has $260$ dimensions.
We also use a random agent to collect data for pre-training the ST-DIM model. The parameter settings are kept the same as described in~\cite{anand2019unsupervised} (See~\nameref{app:strepo} for details). The output embedding from ST-DIM has $256$ dimensions. It is worth mentioning that the ST-DIM is trained on data collected by $8$ differently initialized workers to maintain a low inter-dependency level of the data, while the W-Net only uses one data-collection worker.

{\bf DDQN Parameters:}
The hyperparameters of DDQN are mostly the same as the standard hyperparameters reported in~\cite{Mnih2015} except that we reduce the replay memory size to $50000$ to save memory usage and use a lower learning rate of $0.00001$. 

{\bf APF Parameters:}
In each experimental run of training the APF-augmented agents, we store $2000$ embedding trajectories in the trajectory replay buffer for updating the APF network. The APF update frequency is every episode.
To restrict memory usage, each trajectory contains a maximum of $1000$ most-recently visited embeddings. After each training episode, a batch size of $64$ trajectories is sampled to update the APF network using Eq~\ref{eq:loss_pf}. 

The APF network is implemented using a multi-layer perception (MLP) that consists of two fully connected hidden layers and a fully connected linear output layer. 
The input dimension of the APF network is the same as the output embedding from W-Net which is $13 \times 10\times 2$. 
The first and second hidden layers consist of $128$ and $64$ rectifier units and each is followed by a rectified linear unit (ReLU) activation function~\cite{glorot2011deep} and a dropout layer, respectively.
The output layer generates a single value, representing the estimated potential value at the given game frame (encoded as an embedding).
\medskip

Hereafter, we present the results of the experiment.
The result illustration is in two streams: we first provide a series of input, intermediary, and output pictures of the W-Net network to visualize the representation performance of the W-Net;
Then, we systematically show the performances of APF-WNet-DDQN and the baseline methods.

\subsection*{Encoding Illustration of W-Net}\label{sec:exp-wnet}

\begin{figure}[!h]
     \includegraphics[width=.9\linewidth]{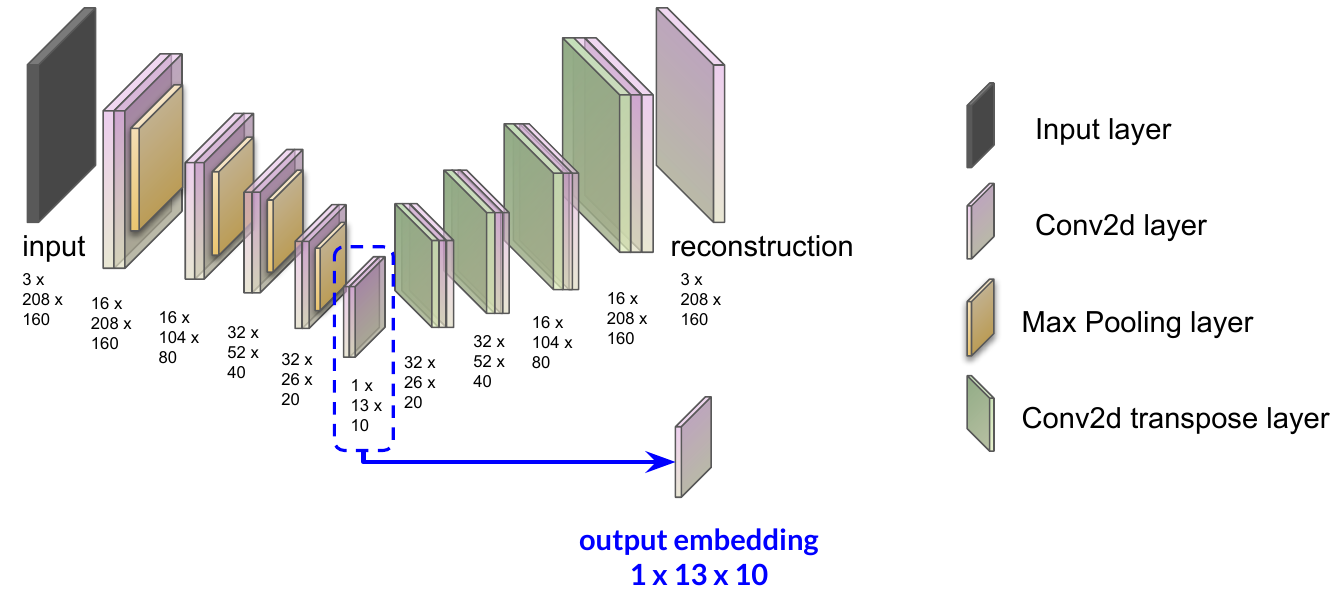}
     \caption{{\bf An illustration of generating embeddings using a single U-Net structure.} In this architecture, the game frame is captured by one bottleneck embedding: The output embedding is a $1 \times 13 \times 10$ feature map, shown at the bottom, center.}
     \label{fig:arc_unet}
\end{figure}

We first deliver the illustrative results showing the efficacy of the W-Net.
We use the Ms Pacman game as an example to show the performance (see Appendix~\nameref{app:fig_wnets} for the details of the complete game range), and compare the encoding and reconstruction results of W-Net with its baseline architecture, a single U-Net as shown in Fig~\ref{fig:arc_unet}. 
Note this paper uses the term U-Net for the U-Net structure with the removal of skip layers as described in Section~\nameref{sec:wnet}, but the term should be distinguished from the U-Nets in~\cite{ronneberger2015u}.

\begin{figure}[!h]
     \includegraphics[width=1.\linewidth]{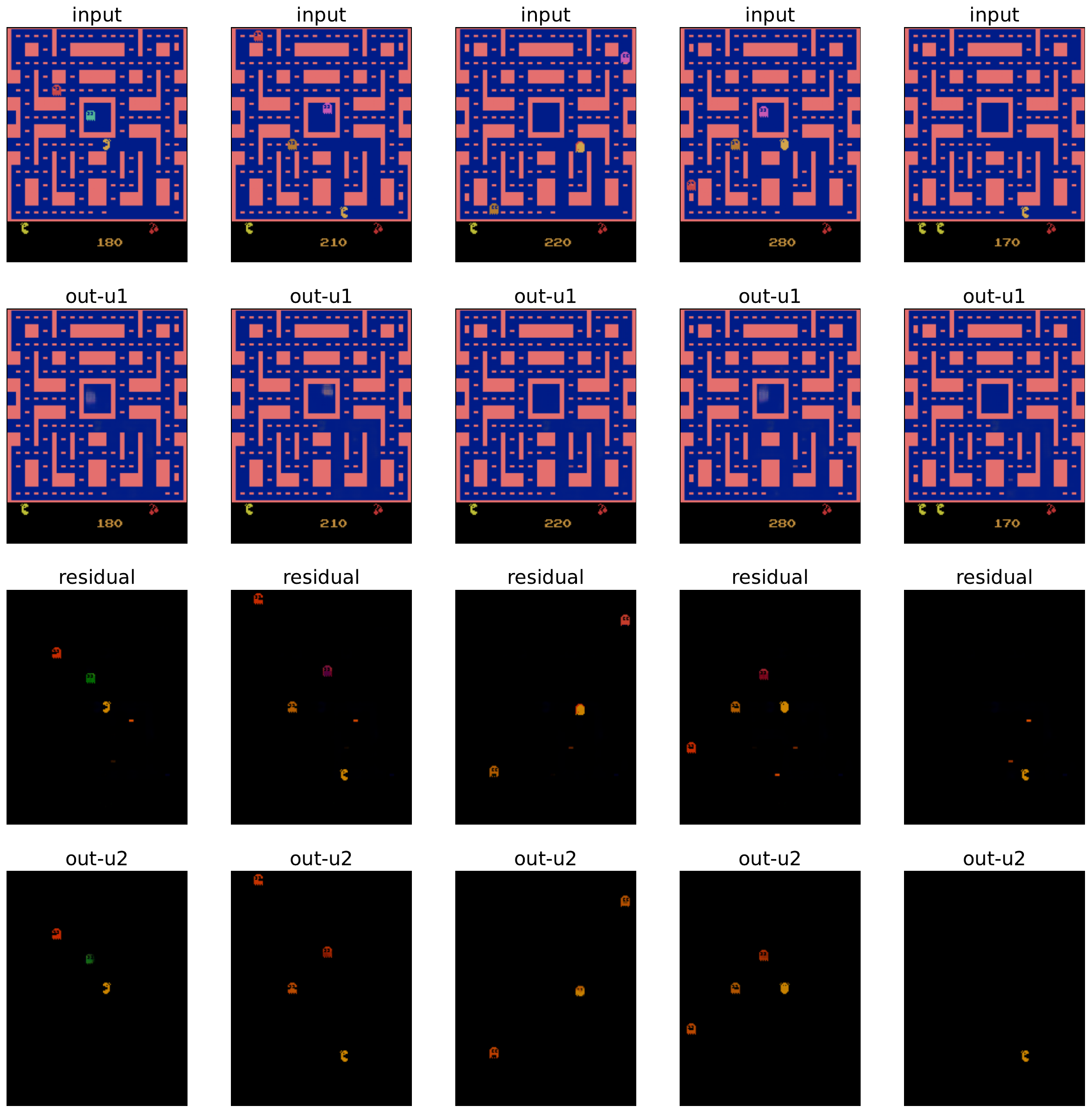}
     \caption{{\bf A visualization of the W-Net's performance in the Ms Pacman Atari game: The first U-Net reconstructs the static information (second row) and the second reconstructs the dynamic information(fourth row).} In specific, the first row shows input images to the W-Net. The second row shows the \texttt{out-u1} layer, the output layer of the first U-Net of the W-Net. The third row shows the \texttt{residual} layer which is the input of the second U-Net. The fourth row depicts the reconstruction of the \texttt{residual} layer, which is the \texttt{out-u2} layer of the second U-Net in the W-Net.}
     \label{fig:resiunet_vis}
\end{figure}

Fig. \ref{fig:resiunet_vis} shows the W-Net reconstructs both static and dynamic information of the game frames. From the top to the bottom row, it shows the input (first row), the output of the first U-Net (out-u1, the second row), the residual of input and out-u1 (residual, the third row), and the output of the second U-Net (out-u2, the fourth row), of W-Net.
Observe that the first U-Net outputs pictures that efficiently reconstruct the background of the game frame, however, it fails to capture the motion entities, i.e., the agent and the enemies, showing complete vanishedness or blurryness of such entities on out-u1.
Utilizing this feature, we instead extract rich information about the motion entities in the residual layer and encode such key information by the second U-Net.
Through the reconstruction of the second U-Net (out-u2), we clearly observe the architecture's encoding performance on the motion entities.

\begin{figure}[!h]
     \includegraphics[width=1.\linewidth]{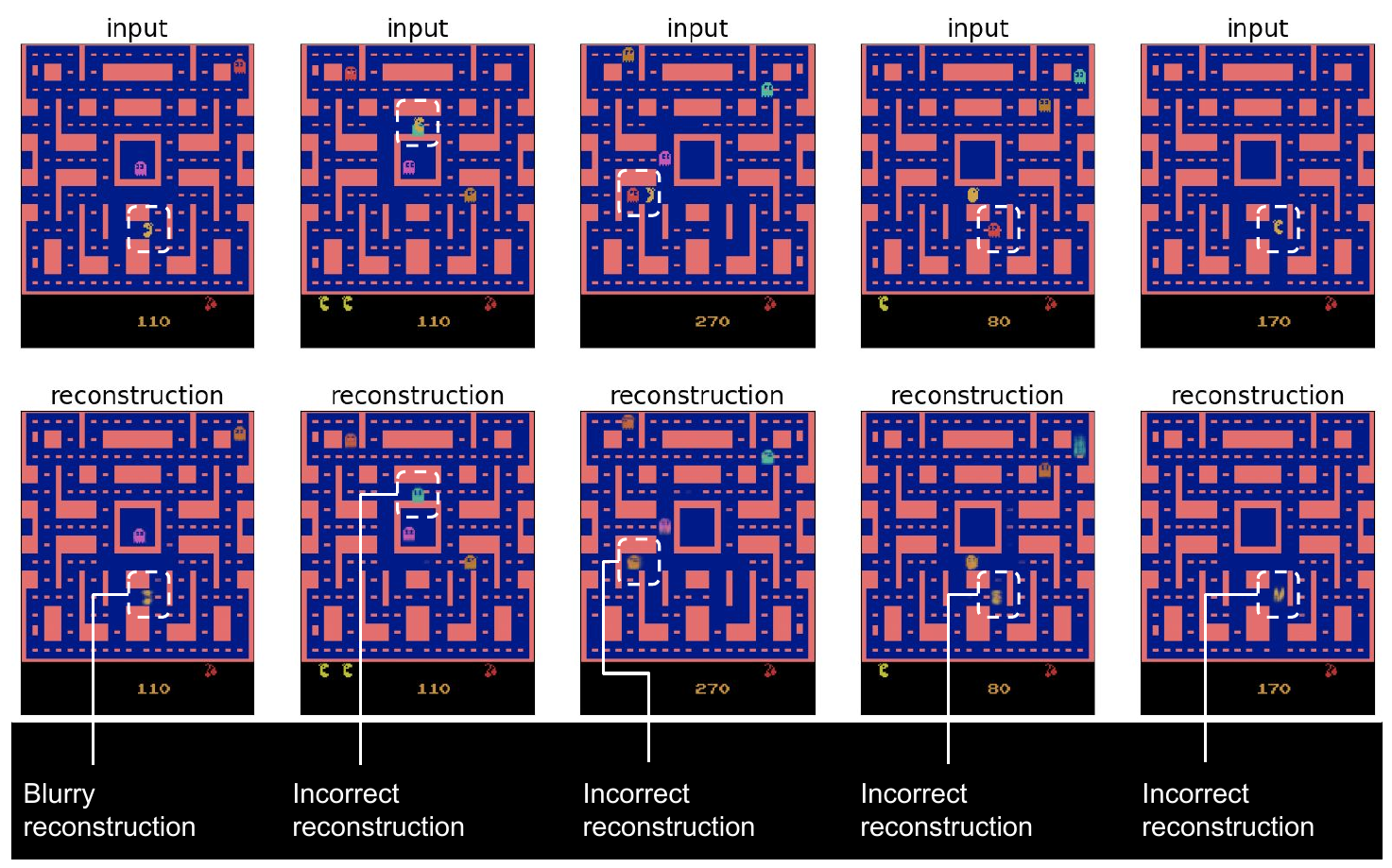}
     \caption{{\bf Reconstruction effect of the single U-Net structure.} The single U-Net reconstructs the average part in the dataset, which makes the reconstruction inaccurate on moving objects such as the important Ms. Pac-Man.}
     \label{fig:unet_loss}
\end{figure}

To compare with the U-Net fairly, we train a U-Net by the same sampled dataset as the training of the W-Net, and show its encoding performance in Fig. \ref{fig:unet_loss}.
Observe that in the reconstructed pictures (the second row), the specifically trained U-Net still suffers the loss of motion entities' feature.
Though it presents higher reconstruction quality than the out-u1 in Fig. \ref{fig:resiunet_vis}, it can still only deliver ambiguous motion entity information to the APF-DDQN framework, as the single U-Net only encodes distorted or blurry features for the important motion entities.

\begin{table}[h]
\centering
\caption{{\bf t-Test results on rewards obtained during training of DDQN versus APF-WNet-DDQN \\ on all $20$ Atari games.}}
\smallskip
\begin{tabular}{|c|cc|cc|cc|}
\hline
 & \multicolumn{2}{c|}{Initial}                                                            & \multicolumn{2}{c|}{Middle}                                                             & \multicolumn{2}{c|}{End}                                                                 \\ \cline{2-7} 
\multirow{-2}{*}{Game} & \multicolumn{1}{c|}{t-values}                        & p-values                         & \multicolumn{1}{c|}{t-values}                        & p-values                         & \multicolumn{1}{c|}{t-values}                        & p-values                          \\ \hline
Asteroids              & \multicolumn{1}{c|}{-0.239}                          & 8.11E-01                         & \multicolumn{1}{c|}{\cellcolor{APFGood}-5.777}  & \cellcolor{APFGood}8.76E-09 & \multicolumn{1}{c|}{\cellcolor{APFGood}-11.211} & \cellcolor{APFGood}2.31E-28  \\ \hline
Berzerk                & \multicolumn{1}{c|}{\cellcolor{APFGood}-44.715} & \cellcolor{APFGood}0.00E+00 & \multicolumn{1}{c|}{\cellcolor{APFGood}-3.448}  & \cellcolor{APFGood}5.69E-04 & \multicolumn{1}{c|}{\cellcolor{APFBad}3.086}   & \cellcolor{APFBad}2.03E-03  \\ \hline
Bowling                & \multicolumn{1}{c|}{1.131}                           & 2.59E-01                         & \multicolumn{1}{c|}{0.278}                           & 7.81E-01                         & \multicolumn{1}{c|}{-0.547}                          & 5.85E-01                          \\ \hline
Boxing                 & \multicolumn{1}{c|}{\cellcolor{APFGood}-4.073}  & \cellcolor{APFGood}4.81E-05 & \multicolumn{1}{c|}{\cellcolor{APFGood}-11.404} & \cellcolor{APFGood}6.52E-29 & \multicolumn{1}{c|}{\cellcolor{APFGood}-24.474} & \cellcolor{APFGood}3.63E-116 \\ \hline
Breakout               & \multicolumn{1}{c|}{\cellcolor{APFGood}-10.234} & \cellcolor{APFGood}2.50E-24 & \multicolumn{1}{c|}{\cellcolor{APFGood}-5.411}  & \cellcolor{APFGood}6.60E-08 & \multicolumn{1}{c|}{\cellcolor{APFGood}-4.495}  & \cellcolor{APFGood}7.14E-06  \\ \hline
DemonAttack            & \multicolumn{1}{c|}{0.136}                           & 8.92E-01                         & \multicolumn{1}{c|}{1.942}                           & 5.23E-02                         & \multicolumn{1}{c|}{\cellcolor{APFGood}-6.114}  & \cellcolor{APFGood}1.37E-09  \\ \hline
Freeway                & \multicolumn{1}{c|}{\cellcolor{APFGood}-8.018}  & \cellcolor{APFGood}6.14E-15 & \multicolumn{1}{c|}{\cellcolor{APFGood}-6.409}  & \cellcolor{APFGood}2.33E-10 & \multicolumn{1}{c|}{\cellcolor{APFGood}-10.799} & \cellcolor{APFGood}1.07E-25  \\ \hline
Frostbite              & \multicolumn{1}{c|}{\cellcolor{APFGood}-2.080}  & \cellcolor{APFGood}3.76E-02 & \multicolumn{1}{c|}{\cellcolor{APFGood}-13.859} & \cellcolor{APFGood}1.30E-42 & \multicolumn{1}{c|}{\cellcolor{APFBad}17.351}  & \cellcolor{APFBad}5.55E-63  \\ \hline
Hero                   & \multicolumn{1}{c|}{1.359}                           & 1.74E-01                         & \multicolumn{1}{c|}{\cellcolor{APFGood}-4.021}  & \cellcolor{APFGood}6.07E-05 & \multicolumn{1}{c|}{\cellcolor{APFGood}-15.921} & \cellcolor{APFGood}4.78E-53  \\ \hline
MsPacman               & \multicolumn{1}{c|}{-0.538}                          & 5.91E-01                         & \multicolumn{1}{c|}{\cellcolor{APFGood}-2.883}  & \cellcolor{APFGood}3.97E-03 & \multicolumn{1}{c|}{\cellcolor{APFGood}-5.619}  & \cellcolor{APFGood}2.13E-08  \\ \hline
Pong                   & \multicolumn{1}{c|}{-1.227}                          & 2.20E-01                         & \multicolumn{1}{c|}{-1.749}                          & 8.05E-02                         & \multicolumn{1}{c|}{-0.329}                          & 7.42E-01                          \\ \hline
PrivateEye             & \multicolumn{1}{c|}{1.257}                           & 2.09E-01                         & \multicolumn{1}{c|}{\cellcolor{APFBad}5.691}   & \cellcolor{APFBad}1.82E-08 & \multicolumn{1}{c|}{\cellcolor{APFBad}12.359}  & \cellcolor{APFBad}8.77E-32  \\ \hline
Qbert                  & \multicolumn{1}{c|}{\cellcolor{APFGood}-1.986}  & \cellcolor{APFGood}4.71E-02 & \multicolumn{1}{c|}{-1.529}                          & 1.26E-01                         & \multicolumn{1}{c|}{\cellcolor{APFGood}-9.549}  & \cellcolor{APFGood}2.34E-21  \\ \hline
Riverraid              & \multicolumn{1}{c|}{-0.538}                          & 5.91E-01                         & \multicolumn{1}{c|}{\cellcolor{APFGood}-2.995}  & \cellcolor{APFGood}2.77E-03 & \multicolumn{1}{c|}{\cellcolor{APFGood}-10.433} & \cellcolor{APFGood}7.08E-25  \\ \hline
Seaquest               & \multicolumn{1}{c|}{0.614}                           & 5.39E-01                         & \multicolumn{1}{c|}{2.221}                           & 2.66E-02                         & \multicolumn{1}{c|}{0.049}                           & 9.61E-01                          \\ \hline
SpaceInvaders          & \multicolumn{1}{c|}{1.030}                           & 3.03E-01                         & \multicolumn{1}{c|}{\cellcolor{APFGood}-3.656}  & \cellcolor{APFGood}2.64E-04 & \multicolumn{1}{c|}{\cellcolor{APFGood}-4.372}  & \cellcolor{APFGood}1.31E-05  \\ \hline
Tennis                 & \multicolumn{1}{c|}{0.362}                           & 7.18E-01                         & \multicolumn{1}{c|}{\cellcolor{APFBad}9.835}   & \cellcolor{APFBad}2.08E-19 & \multicolumn{1}{c|}{\cellcolor{APFGood}-4.100}  & \cellcolor{APFGood}5.19E-05  \\ \hline
Venture                & \multicolumn{1}{c|}{\cellcolor{APFGood}-15.827} & \cellcolor{APFGood}1.64E-47 & \multicolumn{1}{c|}{\cellcolor{APFGood}-20.242} & \cellcolor{APFGood}7.96E-77 & \multicolumn{1}{c|}{\cellcolor{APFGood}-49.707} & \cellcolor{APFGood}6.45E-275 \\ \hline
VideoPinball           & \multicolumn{1}{c|}{\cellcolor{APFBad}3.795}   & \cellcolor{APFBad}1.58E-04 & \multicolumn{1}{c|}{\cellcolor{APFGood}-11.443} & \cellcolor{APFGood}2.39E-28 & \multicolumn{1}{c|}{\cellcolor{APFGood}-16.281} & \cellcolor{APFGood}1.50E-50  \\ \hline
YarsRevenge            & \multicolumn{1}{c|}{\cellcolor{APFGood}-2.982}  & \cellcolor{APFGood}2.90E-03 & \multicolumn{1}{c|}{\cellcolor{APFGood}-6.388}  & \cellcolor{APFGood}2.07E-10 & \multicolumn{1}{c|}{\cellcolor{APFGood}-37.032} & \cellcolor{APFGood}7.00E-223 \\ \hline
\end{tabular}
\smallskip
\begin{flushleft} In the table, green, orange, and white cells denote APF-WNet-DDQN performing significantly better, worse, and no difference compared to DDQN, respectively. Out of $20$ games, $14$ performed better with APF-WNet-DDQN, $3$ performed similarly, and only $3$ games performed worse. A binomial test yields $p = 0.021$ for $14$ in $20$.
\end{flushleft}
\label{tab:wnet}
\end{table}

\paragraph{Results.}\hfill
\medskip

\subsection*{Compare APF-WNet-DDQN and The Baseline Methods}
We show the performance of the APF-WNet-DDQN agent in 20 Atari 2600 games, against the 3 baseline methods, namely the DDQN (Table \ref{tab:wnet}), the APF-STDIM-DDQN (Table \ref{tab:st_wnet}), and APF-ARI-DDQN (Table \ref{tab:ari_wnet}).
We also provide the training curves of APF-WNet-DDQN and the baseline methods on all 20 games, and additional results in Appendix~\nameref{app:curves}.

\paragraph{Metrics.}
In all of the tables, we provide the t-test results for the APF-WNet-DDQN agent's accumulated reward against the corresponding baseline agent's for each game in three training periods: the first $10$ million training game frames (the {\em Initial} columns), the $10-20$ million training game frames (the {\em Middle} columns), and the last $10$ million training game frames (the {\em End} columns).
Green items denote that APF-WNet-DDQN is significantly better than the baseline method, orange items denote that APF-WNet-DDQN is significantly worse than the baseline method, and white items denote no significant difference is detected.
We provide the three periods' performances separately to show the trend during the whole training process, and the End period provides the most important insights into the methods' final achieved performances.

As each row in all of the three tables shows a binary result: whether APF-WNet-DDQN outperforms the baseline method, we further conduct a binomial test to illustrate whether APF-WNet-DDQN outperforms the baseline methods over all of the 20 games.

\noindent {\bf APF-WNet-DDQN v.s. DDQN}
In Table \ref{tab:wnet}, we observe that APF-WNet-DDQN significantly outperforms bare-bones DDQN in terms of the final accumulated rewards (column {\em End}), with APF-WNet-DDQN performing significantly better ($p = 0.021$) than DDQN in 14 out of 20 games (indicated in green, also in subsequent tables).
It is noticeable that in half of the 14 games, APF-WNet-DDQN shows faster converge rates towards a better policy from a non-significant performance difference with DDQN during the {\em Initial} period towards a significantly better performance during the {\em End} period.


These results clearly show the efficacy of the APF framework using the W-Net as a state representation encoder.
The shaped reward guides the RL training process to efficiently converge to better policies than the bare-bones RL architecture using the original reward function.
\medskip

\begin{table}[b!]
\centering
\caption{{\bf T-test results on rewards obtained during training of APF-STDIM-DDQN versus APF-WNet-DDQN on all $20$ Atari games.}}
\smallskip
\begin{tabular}{|c|rr|rr|rr|rr|}
\hline
& \multicolumn{2}{c|}{Initial}                                                            & \multicolumn{2}{c|}{Middle}                                                             & \multicolumn{2}{c|}{End}                                                                 \\ \cline{2-7} 
\multirow{-2}{*}{Game} & \multicolumn{1}{c|}{t-values}                        & \multicolumn{1}{c|}{p-values}    & \multicolumn{1}{c|}{t-values}                        & \multicolumn{1}{c|}{p-values}    & \multicolumn{1}{c|}{t-values}                        & \multicolumn{1}{c|}{p-values}     \\ \hline
Asteroids              & \multicolumn{1}{r|}{\cellcolor{APFBad}9.69}    & \cellcolor{APFBad}9.59E-22 & \multicolumn{1}{r|}{\cellcolor{APFGood}-21.514} & \cellcolor{APFGood}5.30E-90 & \multicolumn{1}{r|}{\cellcolor{APFGood}-31.56}  & \cellcolor{APFGood}6.89E-177 \\ \hline
Berzerk                & \multicolumn{1}{r|}{-48.751}                         & 0.00E+00                         & \multicolumn{1}{r|}{-0.558}                          & 5.77E-01                         & \multicolumn{1}{r|}{\cellcolor{APFGood}-10.606} & \cellcolor{APFGood}4.61E-26  \\ \hline
Bowling                & \multicolumn{1}{r|}{\cellcolor{APFGood}-7.967}  & \cellcolor{APFGood}4.94E-15 & \multicolumn{1}{r|}{\cellcolor{APFGood}-17.261} & \cellcolor{APFGood}1.21E-57 & \multicolumn{1}{r|}{\cellcolor{APFGood}-15.926} & \cellcolor{APFGood}4.03E-49  \\ \hline
Boxing                 & \multicolumn{1}{r|}{-1.137}                          & 2.56E-01                         & \multicolumn{1}{r|}{-0.387}                          & 6.99E-01                         & \multicolumn{1}{r|}{\cellcolor{APFGood}-6.043}  & \cellcolor{APFGood}1.79E-09  \\ \hline
Breakout               & \multicolumn{1}{r|}{\cellcolor{APFBad}5.685}   & \cellcolor{APFBad}1.40E-08 & \multicolumn{1}{r|}{\cellcolor{APFBad}6.74}    & \cellcolor{APFBad}1.78E-11 & \multicolumn{1}{r|}{\cellcolor{APFGood}-2.441}  & \cellcolor{APFGood}1.47E-02  \\ \hline
DemonAttack            & \multicolumn{1}{r|}{1.158}                           & 2.47E-01                         & \multicolumn{1}{r|}{-0.866}                          & 3.87E-01                         & \multicolumn{1}{r|}{-1.811}                          & 7.03E-02                          \\ \hline
Freeway                & \multicolumn{1}{r|}{\cellcolor{APFBad}4.543}   & \cellcolor{APFBad}6.35E-06 & \multicolumn{1}{r|}{0.365}                           & 7.15E-01                         & \multicolumn{1}{r|}{-0.262}                          & 7.93E-01                          \\ \hline
Frostbite              & \multicolumn{1}{r|}{\cellcolor{APFBad}3.011}   & \cellcolor{APFBad}2.62E-03 & \multicolumn{1}{r|}{-1.605}                          & 1.09E-01                         & \multicolumn{1}{r|}{\cellcolor{APFBad}6.365}   & \cellcolor{APFBad}2.19E-10  \\ \hline
Hero                   & \multicolumn{1}{r|}{-0.773}                          & 4.40E-01                         & \multicolumn{1}{r|}{\cellcolor{APFGood}-3.758}  & \cellcolor{APFGood}1.78E-04 & \multicolumn{1}{r|}{\cellcolor{APFBad}4.523}   & \cellcolor{APFBad}6.58E-06  \\ \hline
MsPacman               & \multicolumn{1}{r|}{-0.468}                          & 6.40E-01                         & \multicolumn{1}{r|}{\cellcolor{APFBad}2.494}   & \cellcolor{APFBad}1.27E-02 & \multicolumn{1}{r|}{\cellcolor{APFGood}-8.667}  & \cellcolor{APFGood}7.72E-18  \\ \hline
Pong                   & \multicolumn{1}{r|}{\cellcolor{APFBad}2.682}   & \cellcolor{APFBad}7.43E-03 & \multicolumn{1}{r|}{\cellcolor{APFGood}-1.99}   & \cellcolor{APFGood}4.69E-02 & \multicolumn{1}{r|}{\cellcolor{APFGood}-1.98}   & \cellcolor{APFGood}4.79E-02  \\ \hline
PrivateEye             & \multicolumn{1}{r|}{\cellcolor{APFGood}-10.537} & \cellcolor{APFGood}1.01E-23 & \multicolumn{1}{r|}{\cellcolor{APFGood}-21.281} & \cellcolor{APFGood}3.00E-74 & \multicolumn{1}{r|}{\cellcolor{APFGood}-49.283} & \cellcolor{APFGood}1.72E-233 \\ \hline
Qbert                  & \multicolumn{1}{r|}{-1.016}                          & 3.10E-01                         & \multicolumn{1}{r|}{\cellcolor{APFGood}-4.991}  & \cellcolor{APFGood}6.30E-07 & \multicolumn{1}{r|}{\cellcolor{APFGood}-7.123}  & \cellcolor{APFGood}1.28E-12  \\ \hline
Riverraid              & \multicolumn{1}{r|}{1.371}                           & 1.70E-01                         & \multicolumn{1}{r|}{0.046}                           & 9.63E-01                         & \multicolumn{1}{r|}{\cellcolor{APFGood}-3.945}  & \cellcolor{APFGood}8.24E-05  \\ \hline
Seaquest               & \multicolumn{1}{r|}{\cellcolor{APFBad}5.157}   & \cellcolor{APFBad}2.95E-07 & \multicolumn{1}{r|}{\cellcolor{APFBad}2.227}   & \cellcolor{APFBad}2.61E-02 & \multicolumn{1}{r|}{\cellcolor{APFGood}-2.579}  & \cellcolor{APFGood}1.01E-02  \\ \hline
SpaceInvaders          & \multicolumn{1}{r|}{-0.266}                          & 7.90E-01                         & \multicolumn{1}{r|}{\cellcolor{APFGood}-3.733}  & \cellcolor{APFGood}1.96E-04 & \multicolumn{1}{r|}{-0.061}                          & 9.51E-01                          \\ \hline
Tennis                 & \multicolumn{1}{r|}{-0.67}                           & 5.03E-01                         & \multicolumn{1}{r|}{\cellcolor{APFBad}12.437}  & \cellcolor{APFBad}3.08E-28 & \multicolumn{1}{r|}{\cellcolor{APFGood}-8.133}  & \cellcolor{APFGood}1.54E-14  \\ \hline
Venture                & \multicolumn{1}{r|}{\cellcolor{APFBad}2.554}   & \cellcolor{APFBad}1.08E-02 & \multicolumn{1}{r|}{\cellcolor{APFGood}-8.516}  & \cellcolor{APFGood}5.20E-17 & \multicolumn{1}{r|}{\cellcolor{APFGood}-45.796} & \cellcolor{APFGood}7.47E-264 \\ \hline
VideoPinball           & \multicolumn{1}{r|}{\cellcolor{APFBad}1.964}   & \cellcolor{APFBad}4.98E-02 & \multicolumn{1}{r|}{\cellcolor{APFGood}-7.471}  & \cellcolor{APFGood}1.81E-13 & \multicolumn{1}{r|}{-1.148}                          & 2.51E-01                          \\ \hline
YarsRevenge            & \multicolumn{1}{r|}{\cellcolor{APFGood}-5.622}  & \cellcolor{APFGood}2.13E-08 & \multicolumn{1}{r|}{\cellcolor{APFGood}-2.849}  & \cellcolor{APFGood}4.43E-03 & \multicolumn{1}{r|}{-1.911}                          & 5.62E-02                          \\ \hline
\end{tabular}
\begin{flushleft} In the table, green, orange, and white cells denote APF-WNet-DDQN performing significantly better, worse, and no difference compared to APF-STDIM-DDQN, respectively. Out of $20$ games, $13$ performed better with APF-WNet-DDQN, $5$ performed similarly, and only $2$ games performed worse. A binomial test yields $p = 0.058$ for $13$ in $20$.
\end{flushleft}
\label{tab:st_wnet}
\end{table}

\bigskip
\noindent {\bf APF-WNet-DDQN v.s. APF-STDIM-DDQN}
Table \ref{tab:st_wnet} shows the results of the comparison between APF-WNet-DDQN and APF-STDIM-DDQN.
We first highlight the results for the End training period where APF-WNet-DDQN outperforms APF-STDIM-DDQN in $13$ out of $20$ games, with a p-value of 0.058 for the binomial test, almost significant with a difference of one game.
It is also worth noticing that in a large part of games, APF-WNet-DDQN performs worse (in $8/20$ games) than or indifferent (in $9/20$ games) with APF-STDIM-DDQN, i.e., in the {\em Initial} period.
However, APF-WNet-DDQN converges much faster than APF-STDIM-DDQN to better policies during the Middle period ($13/20$ games) and continues to surpass APF-STDIM-DDQN till the {\em End} period.

We would like to also find out that APF-WNet-DDQN even outperforms APF-STDIM-DDQN in games, such as Bowling and Private Eye, where according to~\cite{anand2019unsupervised}, ST-DIM can efficiently encode important features.
These findings show that the W-Net, as an unsupervised game-state representation network, can efficiently capture the key information in game frames to better serve the APF framework for improved value estimation.
\medskip

\begin{table}[b!]
\centering
\caption{{\bf T-test results on rewards obtained during training of APF-ARI-DDQN versus APF-WNet-DDQN on all $20$ Atari games.}}
\smallskip
\begin{tabular}{|c|rr|rr|rr|rr|}
\hline
& \multicolumn{2}{c|}{Initial}                                                           & \multicolumn{2}{c|}{Middle}                                                              & \multicolumn{2}{c|}{End}                                                                 \\ \cline{2-7} 
\multirow{-2}{*}{Game} & \multicolumn{1}{c|}{t-values}                       & \multicolumn{1}{c|}{p-values}    & \multicolumn{1}{c|}{t-values}                        & \multicolumn{1}{c|}{p-values}     & \multicolumn{1}{c|}{t-values}                        & \multicolumn{1}{c|}{p-values}     \\ \hline
Asteroids              & \multicolumn{1}{r|}{\cellcolor{APFBad}5.326}  & \cellcolor{APFBad}1.11E-07 & \multicolumn{1}{r|}{\cellcolor{APFGood}-22.938} & \cellcolor{APFGood}1.75E-101 & \multicolumn{1}{r|}{\cellcolor{APFGood}-18.79}  & \cellcolor{APFGood}5.60E-73  \\ \hline
Berzerk                & \multicolumn{1}{r|}{\cellcolor{APFBad}2.77}   & \cellcolor{APFBad}5.63E-03 & \multicolumn{1}{r|}{\cellcolor{APFBad}8.174}   & \cellcolor{APFBad}3.66E-16  & \multicolumn{1}{r|}{\cellcolor{APFGood}-18.819} & \cellcolor{APFGood}7.51E-77  \\ \hline
Bowling                & \multicolumn{1}{r|}{\cellcolor{APFGood}-3.572} & \cellcolor{APFGood}3.73E-04 & \multicolumn{1}{r|}{\cellcolor{APFGood}-8.923}  & \cellcolor{APFGood}4.65E-18  & \multicolumn{1}{r|}{\cellcolor{APFGood}-3.558}  & \cellcolor{APFGood}3.92E-04  \\ \hline
Boxing                 & \multicolumn{1}{r|}{-0.426}                         & 6.70E-01                         & \multicolumn{1}{r|}{0.314}                           & 7.54E-01                          & \multicolumn{1}{r|}{-1.218}                          & 2.23E-01                          \\ \hline
Breakout               & \multicolumn{1}{r|}{\cellcolor{APFGood}-6.542} & \cellcolor{APFGood}6.71E-11 & \multicolumn{1}{r|}{-1.078}                          & 2.81E-01                          & \multicolumn{1}{r|}{\cellcolor{APFGood}-2.118}  & \cellcolor{APFGood}3.42E-02  \\ \hline
DemonAttack            & \multicolumn{1}{r|}{\cellcolor{APFBad}6.416}  & \cellcolor{APFBad}2.01E-10 & \multicolumn{1}{r|}{\cellcolor{APFBad}7.195}   & \cellcolor{APFBad}1.18E-12  & \multicolumn{1}{r|}{-1.208}                          & 2.27E-01                          \\ \hline
Freeway                & \multicolumn{1}{r|}{\cellcolor{APFGood}-2.69}  & \cellcolor{APFGood}7.28E-03 & \multicolumn{1}{r|}{\cellcolor{APFGood}-8.374}  & \cellcolor{APFGood}2.05E-16  & \multicolumn{1}{r|}{\cellcolor{APFGood}-13.736} & \cellcolor{APFGood}3.32E-39  \\ \hline
Frostbite              & \multicolumn{1}{r|}{\cellcolor{APFGood}-4.724} & \cellcolor{APFGood}2.40E-06 & \multicolumn{1}{r|}{\cellcolor{APFGood}-10.95}  & \cellcolor{APFGood}1.74E-27  & \multicolumn{1}{r|}{\cellcolor{APFBad}33.038}  & \cellcolor{APFBad}3.91E-189 \\ \hline
Hero                   & \multicolumn{1}{r|}{\cellcolor{APFBad}3.132}  & \cellcolor{APFBad}1.77E-03 & \multicolumn{1}{r|}{\cellcolor{APFGood}-5.413}  & \cellcolor{APFGood}7.22E-08  & \multicolumn{1}{r|}{1.55}                            & 1.21E-01                          \\ \hline
MsPacman               & \multicolumn{1}{r|}{0.73}                           & 4.65E-01                         & \multicolumn{1}{r|}{\cellcolor{APFBad}4.552}   & \cellcolor{APFBad}5.56E-06  & \multicolumn{1}{r|}{\cellcolor{APFBad}2.117}   & \cellcolor{APFBad}3.43E-02  \\ \hline
Pong                   & \multicolumn{1}{r|}{1.481}                          & 1.39E-01                         & \multicolumn{1}{r|}{-0.649}                          & 5.16E-01                          & \multicolumn{1}{r|}{-0.987}                          & 3.24E-01                          \\ \hline
PrivateEye             & \multicolumn{1}{r|}{\cellcolor{APFBad}3.837}  & \cellcolor{APFBad}1.36E-04 & \multicolumn{1}{r|}{-1.555}                          & 1.20E-01                          & \multicolumn{1}{r|}{\cellcolor{APFBad}4.293}   & \cellcolor{APFBad}2.19E-05  \\ \hline
Qbert                  & \multicolumn{1}{r|}{\cellcolor{APFBad}5.236}  & \cellcolor{APFBad}1.73E-07 & \multicolumn{1}{r|}{\cellcolor{APFBad}2.872}   & \cellcolor{APFBad}4.10E-03  & \multicolumn{1}{r|}{-1.878}                          & 6.04E-02                          \\ \hline
Riverraid              & \multicolumn{1}{r|}{\cellcolor{APFBad}4.132}  & \cellcolor{APFBad}3.74E-05 & \multicolumn{1}{r|}{\cellcolor{APFBad}2.051}   & \cellcolor{APFBad}4.04E-02  & \multicolumn{1}{r|}{\cellcolor{APFGood}-5.172}  & \cellcolor{APFGood}2.53E-07  \\ \hline
Seaquest               & \multicolumn{1}{r|}{\cellcolor{APFBad}8.767}  & \cellcolor{APFBad}6.39E-18 & \multicolumn{1}{r|}{\cellcolor{APFBad}3.298}   & \cellcolor{APFBad}1.00E-03  & \multicolumn{1}{r|}{\cellcolor{APFBad}5.331}   & \cellcolor{APFBad}1.19E-07  \\ \hline
SpaceInvaders          & \multicolumn{1}{r|}{\cellcolor{APFGood}-3.22}  & \cellcolor{APFGood}1.31E-03 & \multicolumn{1}{r|}{\cellcolor{APFGood}-2.68}   & \cellcolor{APFGood}7.43E-03  & \multicolumn{1}{r|}{\cellcolor{APFBad}2.965}   & \cellcolor{APFBad}3.07E-03  \\ \hline
Tennis                 & \multicolumn{1}{r|}{\cellcolor{APFBad}5.729}  & \cellcolor{APFBad}2.04E-08 & \multicolumn{1}{r|}{\cellcolor{APFBad}11.045}  & \cellcolor{APFBad}4.58E-23  & \multicolumn{1}{r|}{\cellcolor{APFBad}3.399}   & \cellcolor{APFBad}7.48E-04  \\ \hline
Venture                & \multicolumn{1}{r|}{\cellcolor{APFBad}9.114}  & \cellcolor{APFBad}4.07E-19 & \multicolumn{1}{r|}{\cellcolor{APFGood}-7.961}  & \cellcolor{APFGood}5.28E-15  & \multicolumn{1}{r|}{\cellcolor{APFGood}-20.484} & \cellcolor{APFGood}1.72E-79  \\ \hline
VideoPinball           & \multicolumn{1}{r|}{\cellcolor{APFBad}4.166}  & \cellcolor{APFBad}3.39E-05 & \multicolumn{1}{r|}{\cellcolor{APFBad}10.492}  & \cellcolor{APFBad}2.13E-24  & \multicolumn{1}{r|}{\cellcolor{APFGood}-5.885}  & \cellcolor{APFGood}5.78E-09  \\ \hline
YarsRevenge            & \multicolumn{1}{r|}{\cellcolor{APFGood}-2.202} & \cellcolor{APFGood}2.78E-02 & \multicolumn{1}{r|}{1.264}                           & 2.06E-01                          & \multicolumn{1}{r|}{0.827}                           & 4.08E-01                          \\ \hline
\end{tabular}
\begin{flushleft} In the table, green, orange, and white cells denote APF-WNet-DDQN performing significantly better, worse, and no difference compared to APF-ARI-DDQN, respectively. Out of $20$ games, $8$ performed better with APF-WNet-DDQN, $6$ performed similarly, and $6$ games performed worse. A binomial test yields $p = 0.808$ for $8$ in $20$ indicating that WNet performs similarly with the ground-truth embedding method.
\end{flushleft}
\label{tab:ari_wnet}
\end{table}

\noindent {\bf APF-WNet-DDQN v.s. APF-ARI-DDQN}
Finally, Table \ref{tab:ari_wnet} shows the comparison between APF-WNet-DDQN and APF-ARI-DDQN.
In all three training periods, we cannot find statistical evidence showing a significant difference between APF-WNet-DDQN and APF-ARI-DDQN.
In the most important {\em End} period, APF-WNet-DDQN performs better in $8/20$ games, but worse in $6/20$ games. Therefore, there seems to be no
statistical difference between the two methods.
The reader is reminded that the ARI method uses game-internal information, whereas the W-Net framework only has access to the video frames and still effectively obtains a high performance.


\subsection*{Experiment Discussion}

We now provide our conjecture on the above observations on W-Net's representation performances and APF-WNet-DDQN's learning performances.

\paragraph{W-Net's representation performance.}
By the observation in Section~\nameref{sec:exp-wnet}, we learn that a single U-Net can only encode the average information of the input picture, with neglection or distortion of the picture details.
Compared to the reconstruction performance of the classic U-Net (i.e., the structure with the skip layers), we conjecture that this occurs because a U-Net alone only encodes the average expected state.
It is difficult for it to capture accurate details of the moving entities in a picture sequence, especially of the small entities.
We then utilize such U-Net's feature to concatenate a second U-Net that encodes the residual of the first U-net's encoding, which records rich details of the moving entities.
Therefore, such a double U-Net structure can precisely encode the average state information (e.g., the background of the game frames) and the moving entities' details (e.g., the agent's and the enemies' real-time information).

\paragraph{APF-WNet-DDQN v.s. the baseline methods.}
First, with solid evidence, we observe that it is efficacy to employ the APF reward shaping framework to enhance the performance and efficiency of the basic RL algorithm, i.e., the DDQN method in our experiments.
It shows significant improvement by the APF framework than the basic DDQN method with employing all investigated state representation methods, i.e., the W-Net (Table \ref{tab:st_wnet}), the ST-DIM, and the ARI (tables in~\nameref{app:ttests}).
With such informative state representation methods, APF can guide the RL algorithm to converge towards the optimal policy fast.

However, in the initial period, the APF's state representation method does not have enough historic trajectories to learn informative state potentials, and thus it possibly chaotically tracts the RL algorithm, ending up with poor performances during this period.
Nonetheless, the APF quickly changes to improve the RL algorithm once it starts to learn informative state potentials in the latter periods.

The average higher performance of APF-WNet-DDQN than APF-STDIM-DDQN reveals that the W-Net can better encode information to support the APF in Atari game-playing tasks.
It is worth noticing that it is not the best strategy to utilize all information contained in the game frames to train the APF network.
As introduced above, the ARI interface provides the ground state representations returned by the Atari RAM.
However, part of our APF-ARI-DDQN experiments shows higher performances by only using part of the ARI representation information (see details in~\nameref{app:tab_arilabels}).
We conjecture that some of the state representation information might be irrelevant to the state's potential, and that information may further undermine the APF network's capability of categorizing whether a state is good or bad for achieving high performances.
This is also a potential reason that APF-WNet-DDQN even outperforms the APF-STDIM-DDQN in games where the ST-DIM shows a high representation performance according to the report of \cite{anand2019unsupervised}.

\section*{Conclusion}
\label{sec:con}

This paper proposes a new game-state representation method, W-Net, to extend the application of our previously examined APF for improved adaptive value estimation to the high-dimensional domain.
Specifically, the foundation of APF is the PBRS framework, where the use of a potential function to form the reward-shaping function will guarantee policy invariance.
The key idea of our APF methods is to learn potential functions based on the agent's past experiences to remove the dependency on prior knowledge when designing the potential function. 
Our previous work has validated the beneficial effects of the APF method in environments with low-dimensional state spaces.
This paper demonstrates the feasibility of applying the APF method in environments with high-dimensional state space by using a state representation method. We denote the APF pluses a state encoder format as APF+.

To reduce the dimensionality of the game-state feature vector while retaining both static and event-related information of the current game state, we design a novel neural network structure, W-Net, as the state encoder for pixel-based environments.
Combining our proposed APF+W-Net module with a downstream DDQN agent, we formulate a new RL algorithm APF-WNet-DDQN.
We assess the APF-WNet-DDQN algorithm and compare its performance with the baseline DDQN algorithm and two APF+ augmented baselines (APF-STDIM-DDQN and APF-ARI-DDQN) in twenty Atari-game environments. 
In the evaluation of these problems, we propose to inspect three training zones along the training process and perform statistical testing in each period. The use of W-Net becomes gradually visible during the {\em Initial} and {\em Middle} training stages, with a convincing effect in the {\em End} period of training.
The results show that the APF-WNet-DDQN agent outperforms the bare-bones DDQN agent on $14$ out of $20$ Atari games, and outperforms the APF-STDIM-DDQN agent on $13$ out of $20$ games, considering the last $30\%$ of the training performances. Moreover, the image-frame-based APF-WNet-DDQN performs comparably to the informed APF-ARI-DDQN which extracts embeddings from the ground-truth internal ('RAM') game-state information.
 
It should be noted that DDQN was used as a reference downstream RL algorithm to measure the augmentation effect of using APF. However, it should be noted that the APF+W-Net module can also be used as an enhancement on other RL architectures. Future research may reveal what the benefits of our APF and W-Net-based state representation are in other architectures for high-dimensional state problems.

A possible limitation of the APF-WNet-DDQN algorithm is that, in training, it is dependent and limited to the memory capacity of computing hardware because the agent learns from trajectories of states. Higher memory capacity enables better performance.
Therefore, future directions may include investigating how to reduce the needed data amount, e.g., utilizing the knowledge of large language models in estimating the potential values in reinforcement learning.

In summary, the contributions of this paper are threefold. 
First, we propose the APF+ module that validates the extended use of APF within high-dimensional domains by using a state representation method. 
Second, to address the high-dimensionality and concomitant memory issues when storing agents' past experiences, we design a novel neural network structure, W-Net. Especially, W-Net captures not only the common features of the collected states but also the changing, i.e., surprising parts within current game states. 
Lastly, we assemble the APF+W-Net module within a baseline RL algorithm DDQN to form a new RL algorithm, APF-WNet-DDQN. We empirically compare the APF-WNet-DDQN agent's performance with three algorithms in $20$ Atari games. The results show the efficacy of APF+ in the large majority of task problems.

\section*{Supporting information}\label{sec:app}


\paragraph*{S1 Table}
\label{app:tab_arilabels}
{\bf ARI Labels that are used in training the APF-ARI-DDQN agent.} To achieve the best performance of APF-ARI-DDQN, we select the listed ARI labels to extract embeddings when training the APF-ARI-DDQN agent.

\paragraph*{S1 Supporting Information}
\label{app:strepo}
To implement the Spatiotemporal Deep Infomax (ST-DIM)~\cite{anand2019unsupervised} algorithm, we follow the author-provided code at: \url{https://github.com/mila-iqia/atari-representation-learning}

\paragraph*{S1 Appendix.}
\label{app:ttests}
{\bf T-test results on the DDQN agent versus APF-STDIM-DDQN and APF-ARI-DDQN.} This appendix shows more t-test results on (1) rewards achieved by the agent DDQN versus APF-STDIM-DDQN during training and (2) rewards achieved by the agent DDQN versus APF-ARI-DDQN during training.

\paragraph*{S1 Fig.}
\label{app:curves}
{\bf A comparison of training curves in all twenty games.} This figure showcases the comparison of training curves among APF-WNet-DDQN versus DDQN, APF-STDIM-DDQN and APF-ARI-DDQN in all $20$ Atari games.

\paragraph*{S2 Fig.}
\label{app:fig_wnets}
{\bf The state representation effect of W-Net in all $20$ Atari games.}
This figure showcases the W-Net's state representation ability by visualizing its four key layers while performing the trained W-Net in all $20$ Atari games. The four key layers are the \texttt{input} layer, the \texttt{out-u1} layer, the \texttt{residual} layer, and the \texttt{out-u2} layer. Games differ in the degree to which the output of the second U-Net (\texttt{out-u2}: deviation from expectancy) is visible to the human eye, i.e., without image enhancement. However, the effect of W-Net is clear overall.

\section*{Acknowledgments}
Yifei Chen credits Marco Wiering's role in polishing the idea of adaptive potential function. Yifei Chen acknowledges Yuzhe Zhang's help in improving the paper's writing and structure.

%
%
%


\section*{Appendix}
\subsection*{S1 Table.}


\begin{table}[h]
\begin{adjustwidth}{-2.25in}{0in} 
\centering
\caption*{
{\bf ARI Labels that are used in training the APF-ARI-DDQN agent.}}
\smallskip
\begin{tabular}{|c|l|}
\hline
Game          & Labels  \\ \hline
Asteroids     & num\_lives\_direction, player\_score\_high, player\_score\_low   \\ \hline
Berzerk       & num\_lives, robots\_killed\_count, game\_level, player\_score\_0, player\_score\_1, player\_score\_2         \\ \hline
Bowling       & \begin{tabular}[c]{@{}l@{}}pin\_existence\_0, pin\_existence\_1, pin\_existence\_2, pin\_existence\_3, \\ pin\_existence\_4, pin\_existence\_5, pin\_existence\_6, pin\_existence\_7, pin\_existence\_8, pin\_existence\_9\end{tabular}      \\ \hline
Boxing        & enemy\_score, clock, player\_score   \\ \hline
Breakout      & ball\_x, ball\_y, player\_x       \\ \hline
DemonAttack   & level, player\_x, missile\_y, num\_lives     \\ \hline
Freeway       & \begin{tabular}[c]{@{}l@{}}player\_y, score, enemy\_car\_x\_0, enemy\_car\_x\_1, enemy\_car\_x\_2, enemy\_car\_x\_3, \\ enemy\_car\_x\_4, enemy\_car\_x\_5, enemy\_car\_x\_6, enemy\_car\_x\_7, enemy\_car\_x\_8, enemy\_car\_x\_9\end{tabular}        \\ \hline
Frostbite     & num\_lives, igloo\_blocks\_count, score\_0, score\_1, score\_2      \\ \hline
Hero          & power\_meter, room\_number, level\_number, dynamite\_count, score\_0, score\_1   \\ \hline
MsPacman      & ghosts\_count, dots\_eaten\_count, player\_score, num\_lives      \\ \hline
Pong          & player\_y, player\_x, ball\_x, ball\_y        \\ \hline
PrivateEye    & clock\_0, clock\_1, score\_0, score\_1        \\ \hline
Qbert         & \begin{tabular}[c]{@{}l@{}}tile\_color\_0, tile\_color\_1, tile\_color\_2, tile\_color\_3, tile\_color\_4, tile\_color\_5, tile\_color\_6, \\ tile\_color\_7, tile\_color\_8, tile\_color\_9, tile\_color\_10, tile\_color\_11, tile\_color\_12, tile\_color\_13, \\ tile\_color\_14, tile\_color\_15, tile\_color\_16, tile\_color\_17, tile\_color\_18, tile\_color\_19, tile\_color\_20\end{tabular} \\ \hline
Riverraid     & fuel\_meter\_high, fuel\_meter\_low    \\ \hline
Seaquest      & oxygen\_meter\_value, divers\_collected\_count, num\_lives     \\ \hline
SpaceInvaders & invaders\_left\_count, player\_score, num\_lives, player\_x, enemies\_x, missiles\_y, enemies\_y                                                               \\ \hline
Tennis        & ball\_x, ball\_y, player\_x, player\_y       \\ \hline
Venture       & \begin{tabular}[c]{@{}l@{}}sprite0\_y, sprite1\_y, sprite2\_y, sprite3\_y, sprite4\_y, sprite5\_y, \\ sprite0\_x, sprite1\_x, sprite2\_x, sprite3\_x, sprite4\_x, sprite5\_x, player\_x, player\_y\end{tabular}                   \\ \hline
VideoPinball  & ball\_x, ball\_y         \\ \hline
YarsRevenge   & player\_x, player\_y, enemy\_x, enemy\_y, enemy\_missile\_x, enemy\_missile\_y     \\ \hline
\end{tabular}
\label{tab:arilabels}
\end{adjustwidth}
\end{table}

\newpage
\subsection*{S1 Appendix.}

Table~\ref{tab:stdim} shows t-test results on rewards achieved by the agent DDQN versus APF-STDIM-DDQN during training.
Table~\ref{tab:ari} shows t-test results on rewards achieved by the agent DDQN versus APF-ARI-DDQN during training.

\begin{table}[htpb]
\centering
\caption{{\bf T-test results on rewards obtained during training of DDQN versus APF-STDIM-DDQN on all $20$ Atari games.}}
\smallskip
\begin{tabular}{|c|rr|rr|rr|rr|}
\hline
& \multicolumn{2}{c|}{Initial}                                                            & \multicolumn{2}{c|}{Middle}                                                             & \multicolumn{2}{c|}{End}                                                                 \\ \cline{2-7} 
\multirow{-2}{*}{Game} & \multicolumn{1}{c|}{t-values}                        & \multicolumn{1}{c|}{p-values}    & \multicolumn{1}{c|}{t-values}                        & \multicolumn{1}{c|}{p-values}    & \multicolumn{1}{c|}{t-values}                        & \multicolumn{1}{c|}{p-values}     \\ \hline
Asteroids              & \multicolumn{1}{r|}{\cellcolor{APFGood}-10.223} & \cellcolor{APFGood}5.96E-24 & \multicolumn{1}{r|}{\cellcolor{APFBad}17.362}  & \cellcolor{APFBad}4.25E-62 & \multicolumn{1}{r|}{\cellcolor{APFBad}18.742}  & \cellcolor{APFBad}3.31E-72  \\ \hline
Berzerk                & \multicolumn{1}{r|}{\cellcolor{APFBad}3.463}   & \cellcolor{APFBad}5.39E-04 & \multicolumn{1}{r|}{\cellcolor{APFGood}-3.355}  & \cellcolor{APFGood}7.98E-04 & \multicolumn{1}{r|}{\cellcolor{APFBad}12.872}  & \cellcolor{APFBad}1.98E-37  \\ \hline
Bowling                & \multicolumn{1}{r|}{\cellcolor{APFBad}9.11}    & \cellcolor{APFBad}5.28E-19 & \multicolumn{1}{r|}{\cellcolor{APFBad}10.377}  & \cellcolor{APFBad}1.38E-23 & \multicolumn{1}{r|}{\cellcolor{APFBad}18.225}  & \cellcolor{APFBad}1.07E-62  \\ \hline
Boxing                 & \multicolumn{1}{r|}{\cellcolor{APFGood}-2.946}  & \cellcolor{APFGood}3.25E-03 & \multicolumn{1}{r|}{\cellcolor{APFGood}-10.589} & \cellcolor{APFGood}2.14E-25 & \multicolumn{1}{r|}{\cellcolor{APFGood}-15.348} & \cellcolor{APFGood}3.64E-50  \\ \hline
Breakout               & \multicolumn{1}{r|}{\cellcolor{APFGood}-14.187}  & \cellcolor{APFGood}1.23E-44 & \multicolumn{1}{r|}{\cellcolor{APFGood}-13.119} & \cellcolor{APFGood}1.25E-38 & \multicolumn{1}{r|}{\cellcolor{APFGood}-2.103}  & \cellcolor{APFGood}3.56E-02  \\ \hline
DemonAttack            & \multicolumn{1}{r|}{-1.055}  & 2.92E-01 & \multicolumn{1}{r|}{\cellcolor{APFBad}2.834}   & \cellcolor{APFBad}4.68E-03 & \multicolumn{1}{r|}{\cellcolor{APFGood}-4.657}  & \cellcolor{APFGood}3.58E-06  \\ \hline
Freeway                & \multicolumn{1}{r|}{\cellcolor{APFGood}-11.066} & \cellcolor{APFGood}9.90E-26 & \multicolumn{1}{r|}{\cellcolor{APFGood}-6.928}  & \cellcolor{APFGood}8.05E-12 & \multicolumn{1}{r|}{\cellcolor{APFGood}-10.538} & \cellcolor{APFGood}1.30E-24  \\ \hline
Frostbite              & \multicolumn{1}{r|}{\cellcolor{APFGood}-4.944}  & \cellcolor{APFGood}8.01E-07 & \multicolumn{1}{r|}{\cellcolor{APFGood}-12.162} & \cellcolor{APFGood}2.21E-33 & \multicolumn{1}{r|}{\cellcolor{APFBad}15.205}  & \cellcolor{APFBad}1.84E-49  \\ \hline
Hero                   & \multicolumn{1}{r|}{\cellcolor{APFBad}2.144}   & \cellcolor{APFBad}3.22E-02 & \multicolumn{1}{r|}{-0.123}  & 9.02E-01 & \multicolumn{1}{r|}{\cellcolor{APFGood}-18.565} & \cellcolor{APFGood}6.26E-69  \\ \hline
MsPacman               & \multicolumn{1}{r|}{-0.067}                          & 9.47E-01                         & \multicolumn{1}{r|}{\cellcolor{APFGood}-5.441}  & \cellcolor{APFGood}5.78E-08 & \multicolumn{1}{r|}{\cellcolor{APFBad}3.403}   & \cellcolor{APFBad}6.77E-04  \\ \hline
Pong                   & \multicolumn{1}{r|}{\cellcolor{APFGood}-3.979}  & \cellcolor{APFGood}7.37E-05 & \multicolumn{1}{r|}{0.237}   & 8.12E-01 & \multicolumn{1}{r|}{1.612}   & 1.07E-01  \\ \hline
PrivateEye             & \multicolumn{1}{r|}{\cellcolor{APFBad}11.399}  & \cellcolor{APFBad}4.89E-27 & \multicolumn{1}{r|}{\cellcolor{APFBad}25.212}  & \cellcolor{APFBad}5.64E-94 & \multicolumn{1}{r|}{\cellcolor{APFBad}53.619}  & \cellcolor{APFBad}2.98E-254 \\ \hline
Qbert                  & \multicolumn{1}{r|}{-0.95}   & 3.42E-01 & \multicolumn{1}{r|}{\cellcolor{APFBad}3.512}   & \cellcolor{APFBad}4.50E-04 & \multicolumn{1}{r|}{\cellcolor{APFGood}-2.423}  & \cellcolor{APFGood}1.54E-02  \\ \hline
Riverraid              & \multicolumn{1}{r|}{-1.921}  & 5.49E-02 & \multicolumn{1}{r|}{\cellcolor{APFGood}-3.141}  & \cellcolor{APFGood}1.71E-03 & \multicolumn{1}{r|}{\cellcolor{APFGood}-7.259}  & \cellcolor{APFGood}5.44E-13  \\ \hline
Seaquest               & \multicolumn{1}{r|}{\cellcolor{APFGood}-4.864}  & \cellcolor{APFGood}1.31E-06 & \multicolumn{1}{r|}{0.127}   & 8.99E-01 & \multicolumn{1}{r|}{\cellcolor{APFBad}2.862}   & \cellcolor{APFBad}4.29E-03  \\ \hline
SpaceInvaders          & \multicolumn{1}{r|}{1.338}   & 1.81E-01 & \multicolumn{1}{r|}{0.007}   & 9.94E-01 & \multicolumn{1}{r|}{\cellcolor{APFGood}-3.812}  & \cellcolor{APFGood}1.43E-04  \\ \hline
Tennis                 & \multicolumn{1}{r|}{1.12}    & 2.63E-01 & \multicolumn{1}{r|}{\cellcolor{APFGood}-4.108}  & \cellcolor{APFGood}4.73E-05 & \multicolumn{1}{r|}{\cellcolor{APFBad}4.795}   & \cellcolor{APFBad}2.24E-06  \\ \hline
Venture                & \multicolumn{1}{r|}{\cellcolor{APFGood}-16.143} & \cellcolor{APFGood}2.73E-49 & \multicolumn{1}{r|}{\cellcolor{APFGood}-13.481} & \cellcolor{APFGood}3.15E-38 & \multicolumn{1}{r|}{\cellcolor{APFGood}-5.657}  & \cellcolor{APFGood}2.02E-08  \\ \hline
VideoPinball           & \multicolumn{1}{r|}{1.607}   & 1.08E-01 & \multicolumn{1}{r|}{\cellcolor{APFGood}-3.539}  & \cellcolor{APFGood}4.22E-04 & \multicolumn{1}{r|}{\cellcolor{APFGood}-14.505} & \cellcolor{APFGood}1.03E-41  \\ \hline
YarsRevenge            & \multicolumn{1}{r|}{\cellcolor{APFBad}2.639}   & \cellcolor{APFBad}8.39E-03 & \multicolumn{1}{r|}{\cellcolor{APFGood}-3.185}  & \cellcolor{APFGood}1.47E-03 & \multicolumn{1}{r|}{\cellcolor{APFGood}-26.406} & \cellcolor{APFGood}3.49E-125 \\ \hline
\end{tabular}
\begin{flushleft} In the table, green, orange, and white cells denote APF-STDIM-DDQN performing significantly better, worse, and no difference compared to DDQN, respectively. Out of $20$ games, $11$ performed better with APF-STDIM-DDQN, $1$ performed similarly, and only $8$ games performed worse. A binomial test yields $p = 0.252$ for $11$ in $20$.
\end{flushleft}
\label{tab:stdim}
\end{table}

\begin{table}[htpb]
\centering
\caption{{\bf T-test results on rewards obtained during training of DDQN versus APF-ARI-DDQN on all $20$ Atari games.}}
\smallskip
\begin{tabular}{|c|cc|cc|cc|cc|}
\hline
& \multicolumn{2}{c|}{Initial}                                                            & \multicolumn{2}{c|}{Middle}                                                              & \multicolumn{2}{c|}{End}                                                                 \\ \cline{2-7} 
\multirow{-2}{*}{Game} & \multicolumn{1}{c|}{init-t}                          & init-p                           & \multicolumn{1}{c|}{mid-t}                           & mid-p                             & \multicolumn{1}{c|}{end-t}                           & end-p                             \\ \hline
Asteroids              & \multicolumn{1}{c|}{\cellcolor{APFGood}-5.764}  & \cellcolor{APFGood}9.44E-09 & \multicolumn{1}{c|}{\cellcolor{APFBad}18.374}  & \cellcolor{APFBad}3.68E-69  & \multicolumn{1}{c|}{\cellcolor{APFBad}7.641}   & \cellcolor{APFBad}3.23E-14  \\ \hline
Berzerk                & \multicolumn{1}{c|}{\cellcolor{APFGood}-40.115} & \cellcolor{APFGood}0.00E+00 & \multicolumn{1}{c|}{\cellcolor{APFGood}-11.692} & \cellcolor{APFGood}3.29E-31  & \multicolumn{1}{c|}{\cellcolor{APFBad}20.404}  & \cellcolor{APFBad}1.80E-89  \\ \hline
Bowling                & \multicolumn{1}{c|}{\cellcolor{APFBad}4.725}   & \cellcolor{APFBad}2.66E-06 & \multicolumn{1}{c|}{\cellcolor{APFBad}6.748}   & \cellcolor{APFBad}2.65E-11  & \multicolumn{1}{c|}{\cellcolor{APFBad}3.407}   & \cellcolor{APFBad}6.87E-04  \\ \hline
Boxing                 & \multicolumn{1}{c|}{\cellcolor{APFGood}-3.660}  & \cellcolor{APFGood}2.59E-04 & \multicolumn{1}{c|}{\cellcolor{APFGood}-11.393} & \cellcolor{APFGood}6.42E-29  & \multicolumn{1}{c|}{\cellcolor{APFGood}-21.793} & \cellcolor{APFGood}1.93E-94  \\ \hline
Breakout               & \multicolumn{1}{c|}{\cellcolor{APFGood}-3.796}  & \cellcolor{APFGood}1.49E-04 & \multicolumn{1}{c|}{\cellcolor{APFGood}-4.422}  & \cellcolor{APFGood}1.00E-05  & \multicolumn{1}{c|}{\cellcolor{APFGood}-2.403}  & \cellcolor{APFGood}1.63E-02  \\ \hline
DemonAttack            & \multicolumn{1}{c|}{\cellcolor{APFGood}-6.405}  & \cellcolor{APFGood}2.19E-10 & \multicolumn{1}{c|}{\cellcolor{APFGood}-5.804}  & \cellcolor{APFGood}8.65E-09  & \multicolumn{1}{c|}{\cellcolor{APFGood}-4.888}  & \cellcolor{APFGood}1.18E-06  \\ \hline
Freeway                & \multicolumn{1}{c|}{\cellcolor{APFGood}-6.456}  & \cellcolor{APFGood}2.29E-10 & \multicolumn{1}{c|}{1.616}                           & 1.06E-01                          & \multicolumn{1}{c|}{\cellcolor{APFBad}3.076}   & \cellcolor{APFBad}2.16E-03  \\ \hline
Frostbite              & \multicolumn{1}{c|}{\cellcolor{APFBad}2.458}   & \cellcolor{APFBad}1.40E-02 & \multicolumn{1}{c|}{\cellcolor{APFGood}-2.249}  & \cellcolor{APFGood}2.46E-02  & \multicolumn{1}{c|}{\cellcolor{APFGood}-19.837} & \cellcolor{APFGood}3.45E-82  \\ \hline
Hero                   & \multicolumn{1}{c|}{-1.714}                          & 8.68E-02                         & \multicolumn{1}{c|}{1.276}                           & 2.02E-01                          & \multicolumn{1}{c|}{\cellcolor{APFGood}-13.794} & \cellcolor{APFGood}1.90E-40  \\ \hline
MsPacman               & \multicolumn{1}{c|}{-1.239}                          & 2.16E-01                         & \multicolumn{1}{c|}{\cellcolor{APFGood}-7.454}  & \cellcolor{APFGood}1.24E-13  & \multicolumn{1}{c|}{\cellcolor{APFGood}-7.843}  & \cellcolor{APFGood}6.44E-15  \\ \hline
Pong                   & \multicolumn{1}{c|}{\cellcolor{APFGood}-2.777}  & \cellcolor{APFGood}5.57E-03 & \multicolumn{1}{c|}{-1.096}                          & 2.73E-01                          & \multicolumn{1}{c|}{0.637}                           & 5.24E-01                          \\ \hline
PrivateEye             & \multicolumn{1}{c|}{\cellcolor{APFGood}-2.601}  & \cellcolor{APFGood}9.48E-03 & \multicolumn{1}{c|}{\cellcolor{APFBad}6.720}   & \cellcolor{APFBad}3.81E-11  & \multicolumn{1}{c|}{1.622}                           & 1.06E-01                          \\ \hline
Qbert                  & \multicolumn{1}{c|}{\cellcolor{APFGood}-7.002}  & \cellcolor{APFGood}3.01E-12 & \multicolumn{1}{c|}{\cellcolor{APFGood}-4.540}  & \cellcolor{APFGood}5.82E-06  & \multicolumn{1}{c|}{\cellcolor{APFGood}-7.962}  & \cellcolor{APFGood}2.26E-15  \\ \hline
Riverraid              & \multicolumn{1}{c|}{\cellcolor{APFGood}-4.697}  & \cellcolor{APFGood}2.80E-06 & \multicolumn{1}{c|}{\cellcolor{APFGood}-5.262}  & \cellcolor{APFGood}1.56E-07  & \multicolumn{1}{c|}{\cellcolor{APFGood}-6.035}  & \cellcolor{APFGood}1.87E-09  \\ \hline
Seaquest               & \multicolumn{1}{c|}{\cellcolor{APFGood}-8.643}  & \cellcolor{APFGood}1.91E-17 & \multicolumn{1}{c|}{-0.827}                          & 4.08E-01                          & \multicolumn{1}{c|}{\cellcolor{APFGood}-5.508}  & \cellcolor{APFGood}4.62E-08  \\ \hline
SpaceInvaders          & \multicolumn{1}{c|}{\cellcolor{APFBad}4.310}   & \cellcolor{APFBad}1.73E-05 & \multicolumn{1}{c|}{-0.722}                          & 4.70E-01                          & \multicolumn{1}{c|}{\cellcolor{APFGood}-6.482}  & \cellcolor{APFGood}1.22E-10  \\ \hline
Tennis                 & \multicolumn{1}{c|}{\cellcolor{APFGood}-5.907}  & \cellcolor{APFGood}8.18E-09 & \multicolumn{1}{c|}{0.409}                           & 6.83E-01                          & \multicolumn{1}{c|}{\cellcolor{APFGood}-8.376}  & \cellcolor{APFGood}8.69E-16  \\ \hline
Venture                & \multicolumn{1}{c|}{\cellcolor{APFGood}-19.431} & \cellcolor{APFGood}6.35E-67 & \multicolumn{1}{c|}{\cellcolor{APFGood}-18.103} & \cellcolor{APFGood}2.48E-61  & \multicolumn{1}{c|}{\cellcolor{APFGood}-17.166} & \cellcolor{APFGood}3.11E-58  \\ \hline
VideoPinball           & \multicolumn{1}{c|}{-0.367}                          & 7.14E-01                         & \multicolumn{1}{c|}{\cellcolor{APFGood}-21.989} & \cellcolor{APFGood}3.05E-85  & \multicolumn{1}{c|}{\cellcolor{APFGood}-12.853} & \cellcolor{APFGood}1.66E-34  \\ \hline
YarsRevenge            & \multicolumn{1}{c|}{-0.776}                          & 4.38E-01                         & \multicolumn{1}{c|}{\cellcolor{APFGood}-7.188}  & \cellcolor{APFGood}9.36E-13  & \multicolumn{1}{c|}{\cellcolor{APFGood}-38.402} & \cellcolor{APFGood}2.36E-231 \\ \hline
\end{tabular}
\begin{flushleft} In the table, green, orange, and white cells denote APF-ARI-DDQN performing significantly better, worse, and no difference compared to DDQN, respectively. Out of $20$ games, $14$ performed better with APF-ARI-DDQN, $2$ performed similarly, and only $4$ games performed worse. A binomial test yields $p = 0.039$ for $14$ in $20$.
\end{flushleft}
\label{tab:ari}
\end{table}

\newpage
\subsection*{S1 Fig.}

{\bf A comparison of training curves in all twenty games.} The green curve represents our proposed APF-WNet-DDQN algorithm. The blue curve is the bare baseline DDQN. The red curve denotes APF-STDIM-DDQN and the orange curve shows the average score for APF-ARI-DDQN during training.

\begin{figure}[htbp]
\begin{adjustwidth}{-2.25in}{0in}
     \centering
     \includegraphics[width=.3\linewidth]{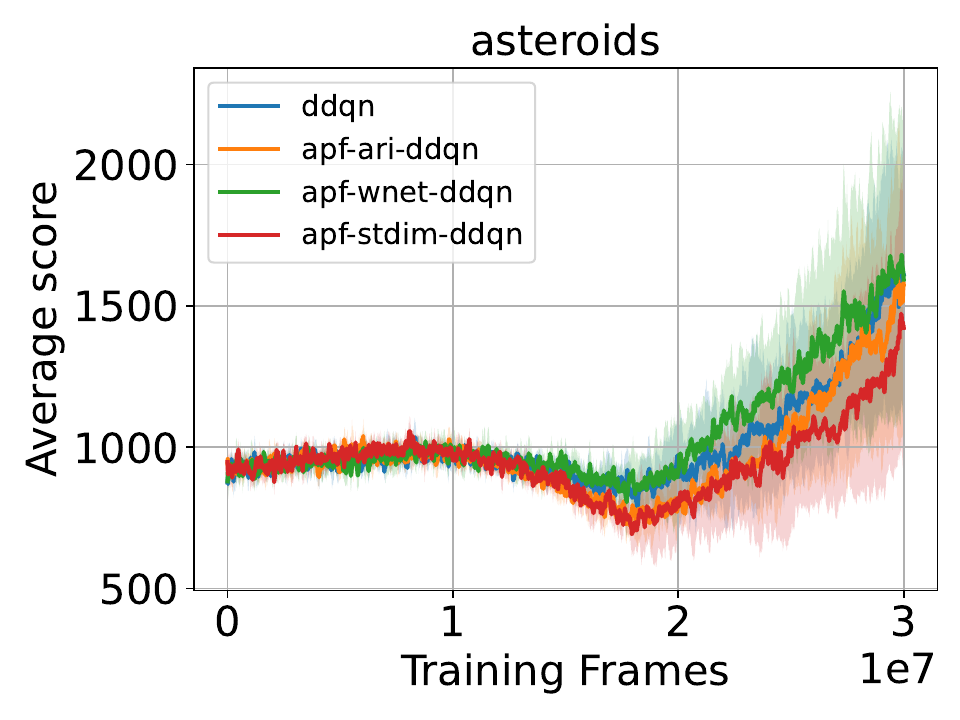}
     \includegraphics[width=.3\linewidth]{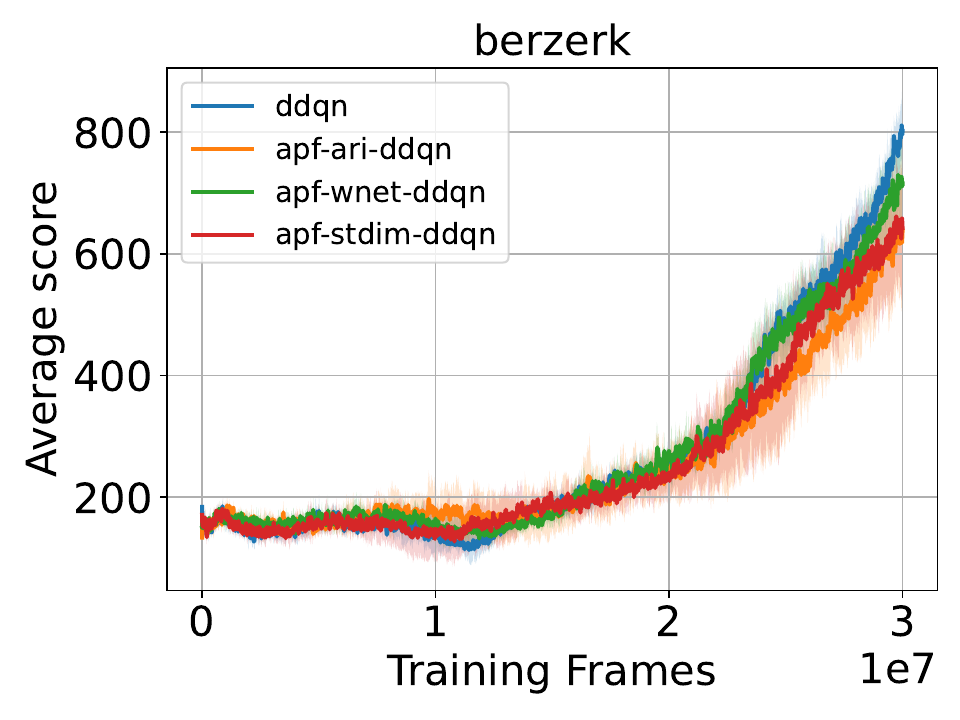}
     \includegraphics[width=.3\linewidth]{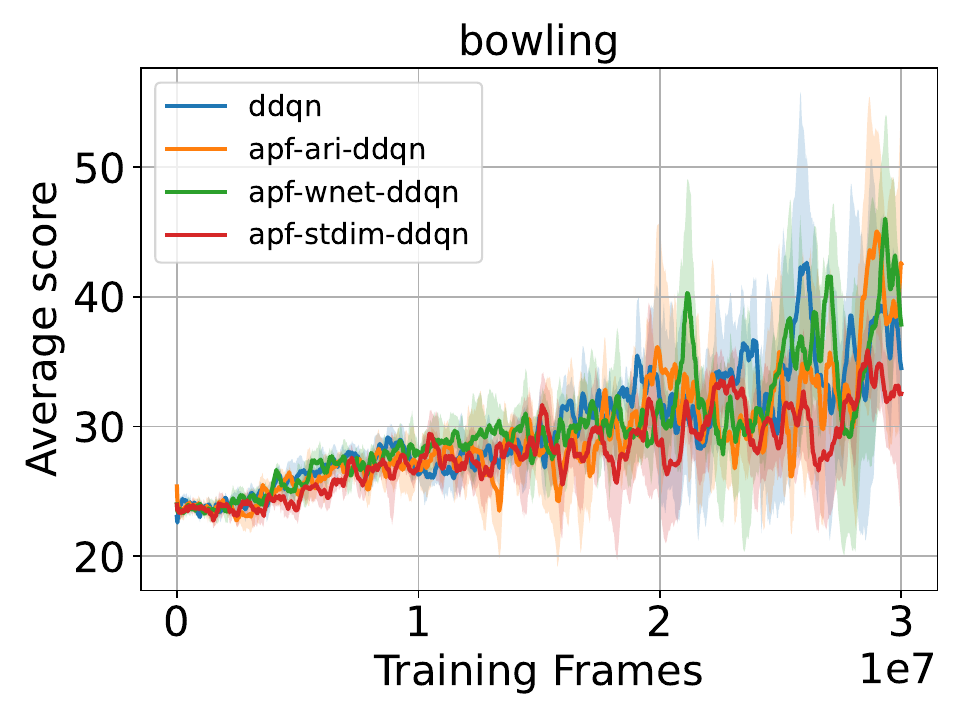}\\
     \includegraphics[width=.3\linewidth]{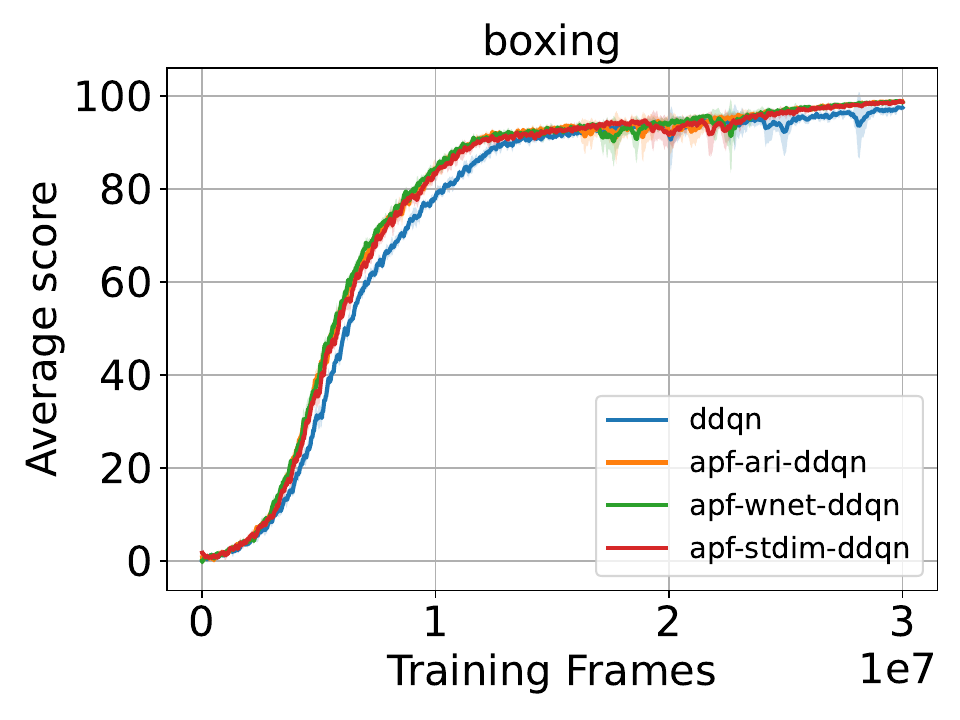}
     \includegraphics[width=.3\linewidth]{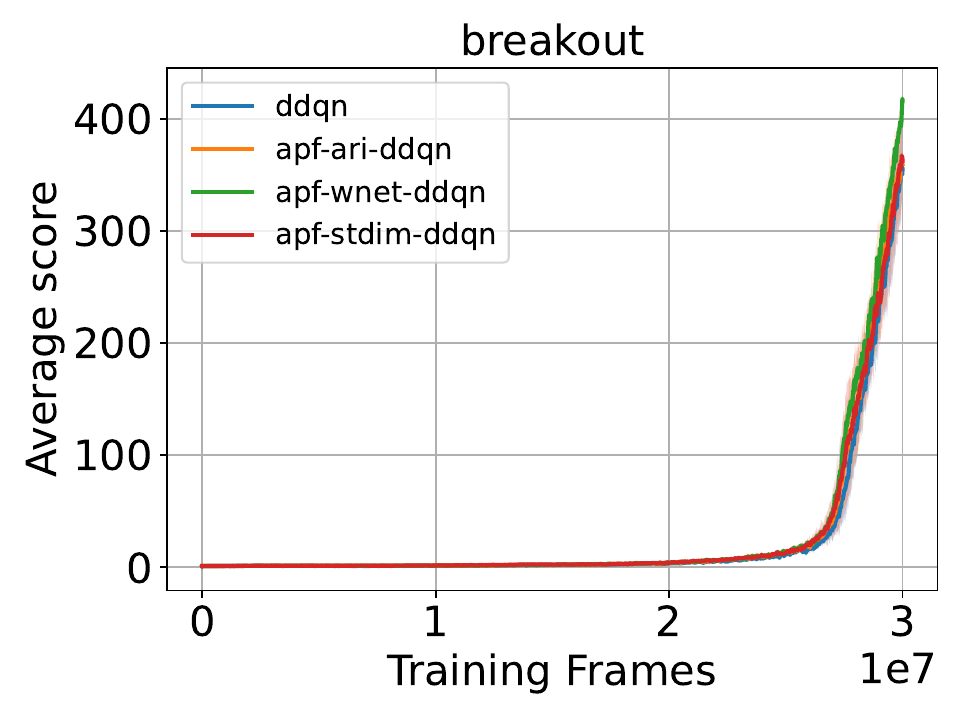}
     \includegraphics[width=.3\linewidth]{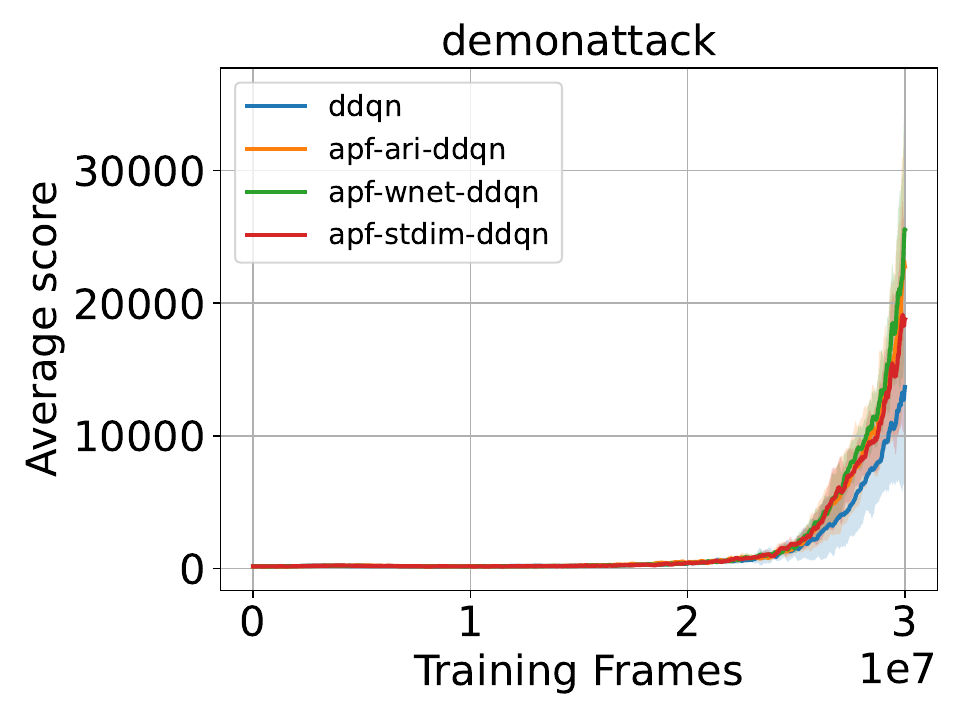}\\
     \includegraphics[width=.3\linewidth]{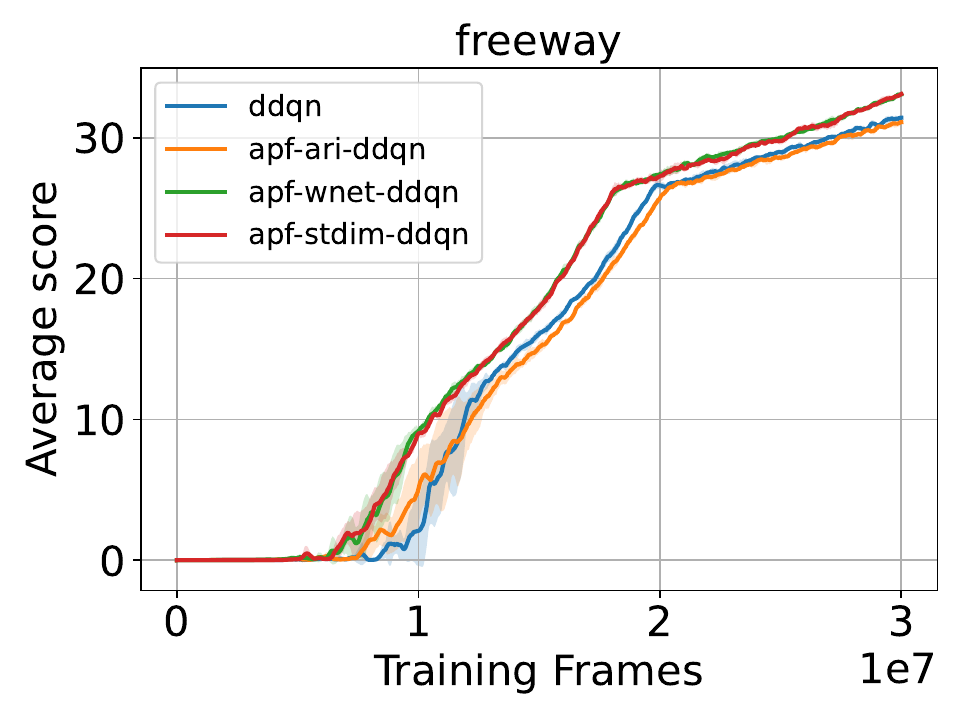}
     \includegraphics[width=.3\linewidth]{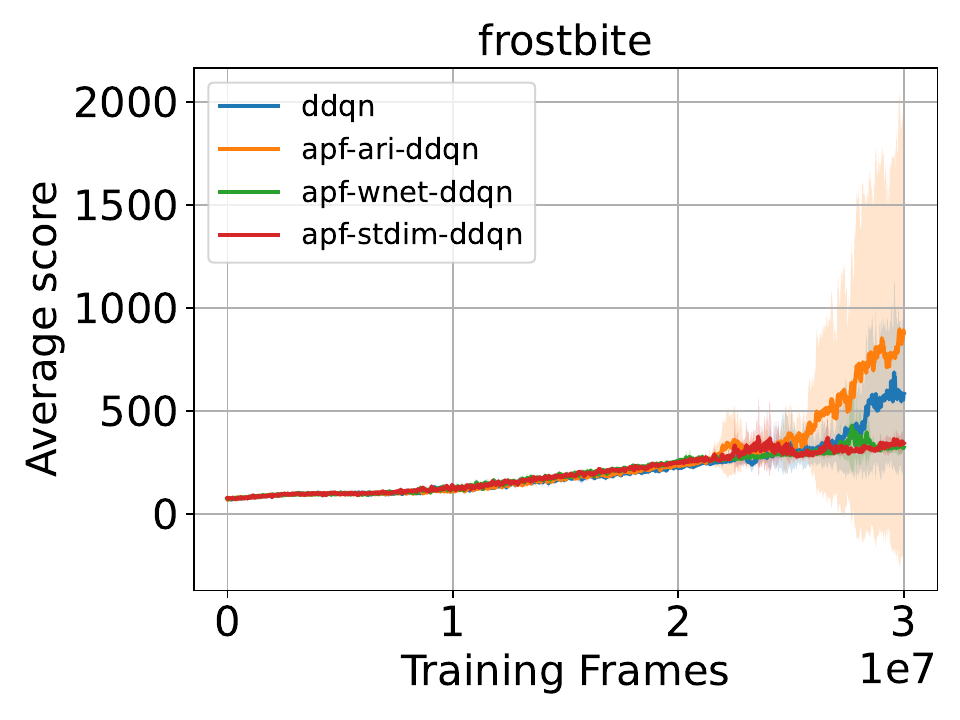}
     \includegraphics[width=.3\linewidth]{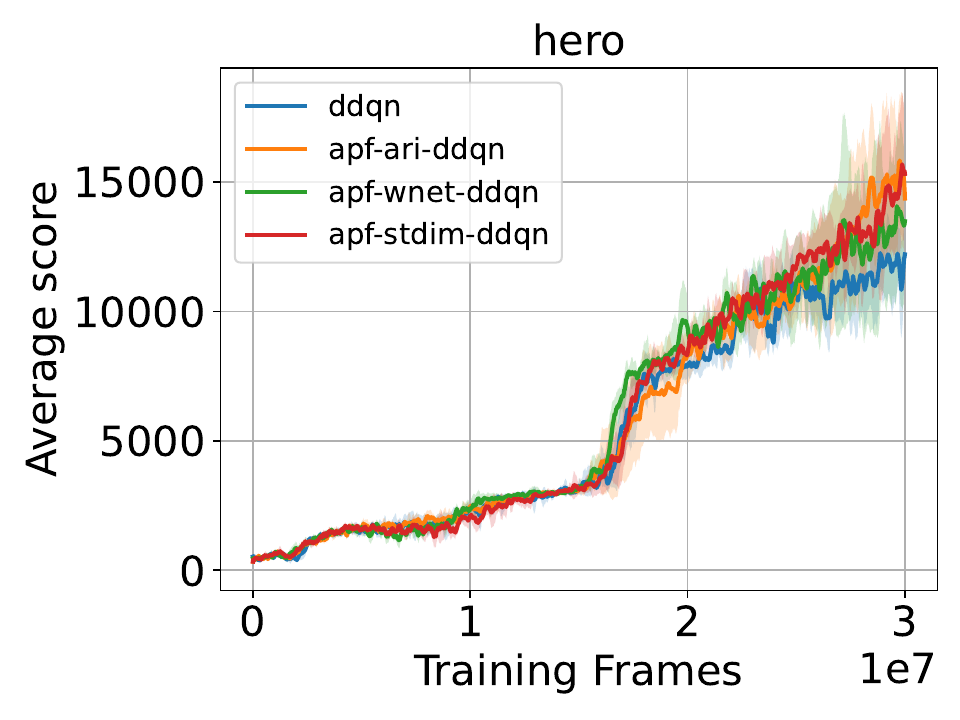}

\end{adjustwidth}
\end{figure}

\clearpage

\begin{figure}[!h]
\begin{adjustwidth}{-2.25in}{0in}
     \centering
     \includegraphics[width=.3\linewidth]{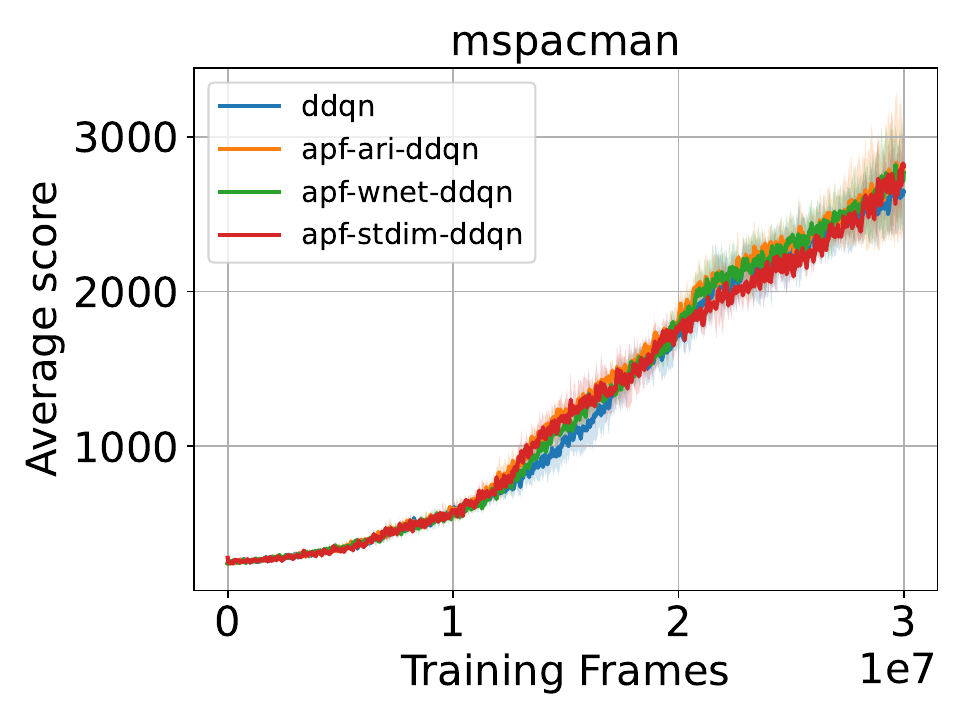}
     \includegraphics[width=.3\linewidth]{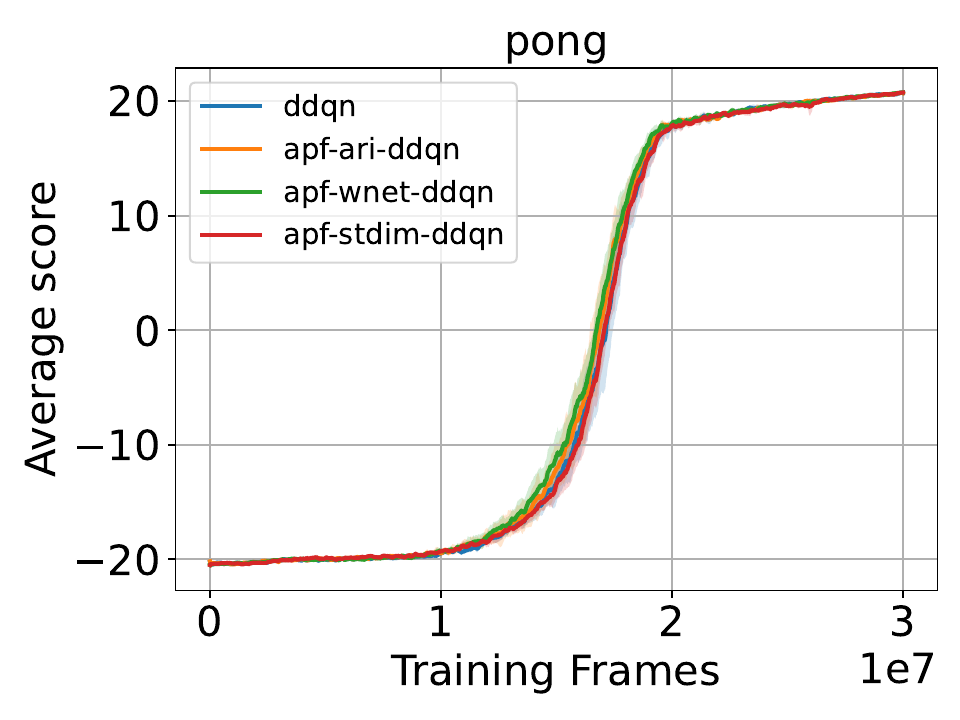}\\
     \includegraphics[width=.3\linewidth]{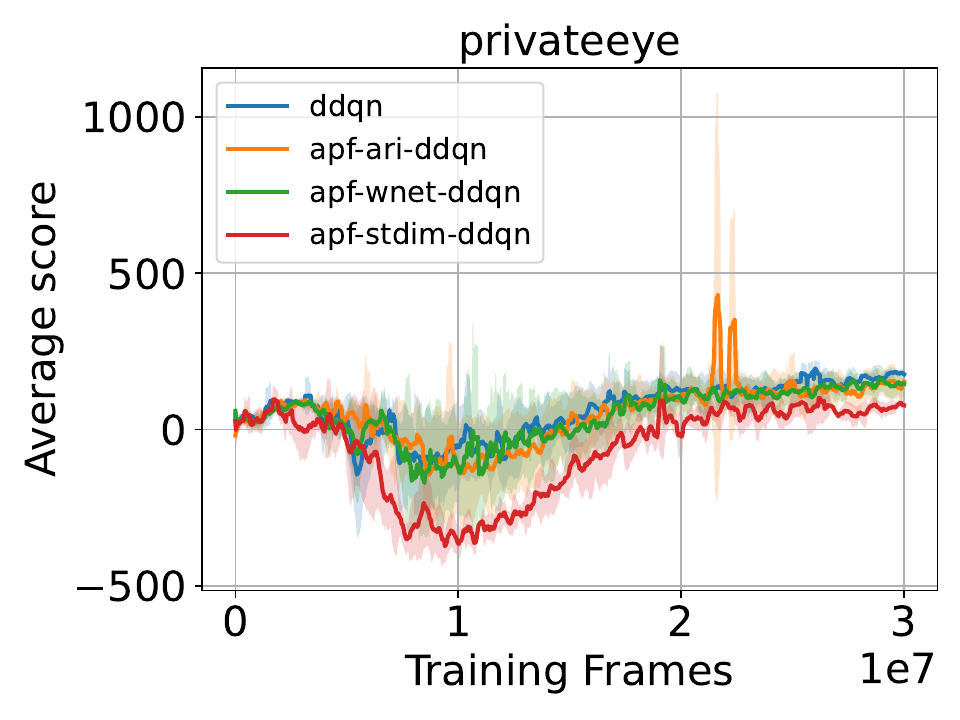}
     \includegraphics[width=.3\linewidth]{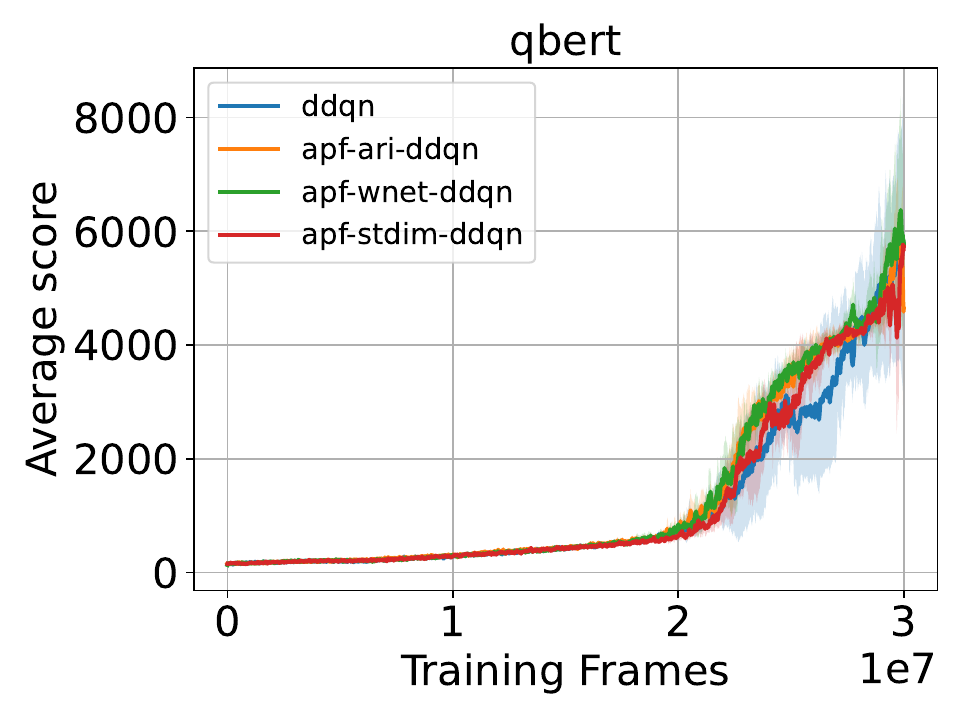}
     \includegraphics[width=.3\linewidth]{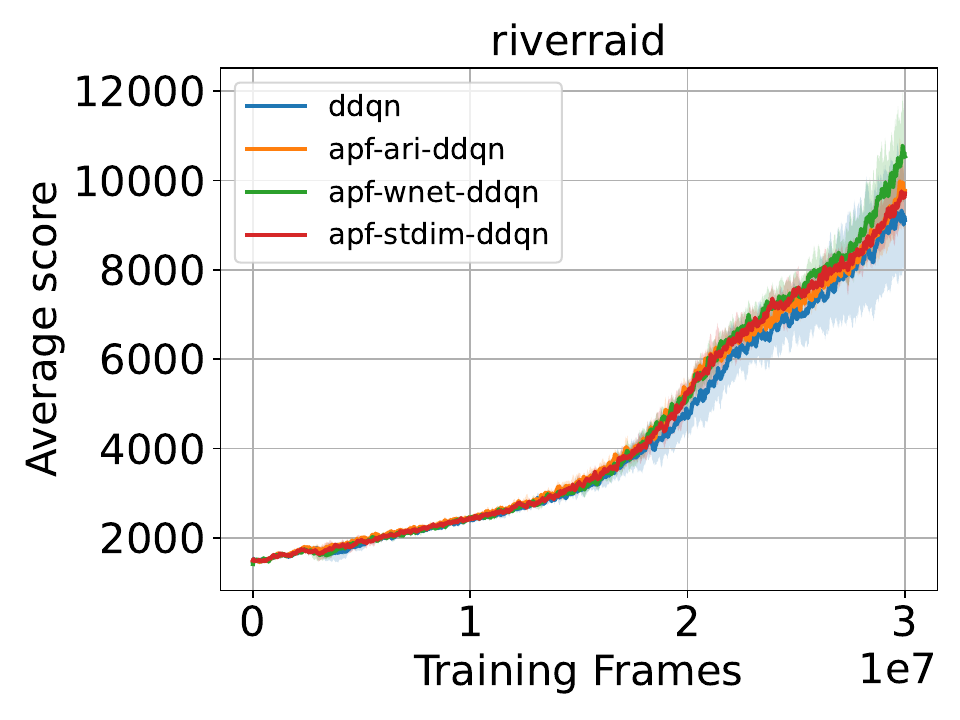}\\
     \includegraphics[width=.3\linewidth]{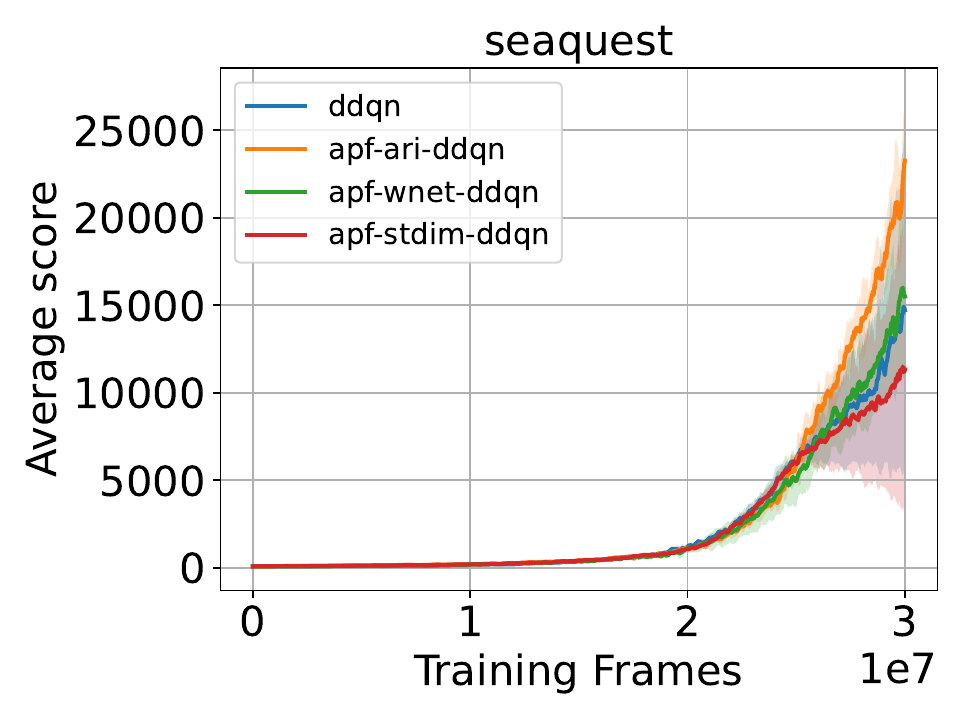}
     \includegraphics[width=.3\linewidth]{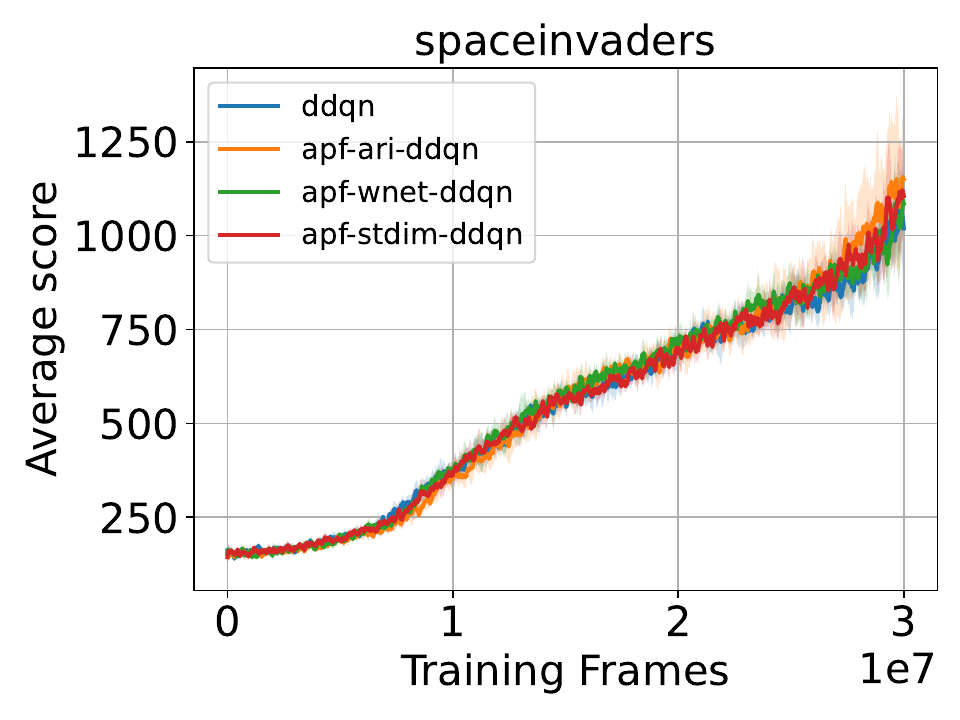}
     \includegraphics[width=.3\linewidth]{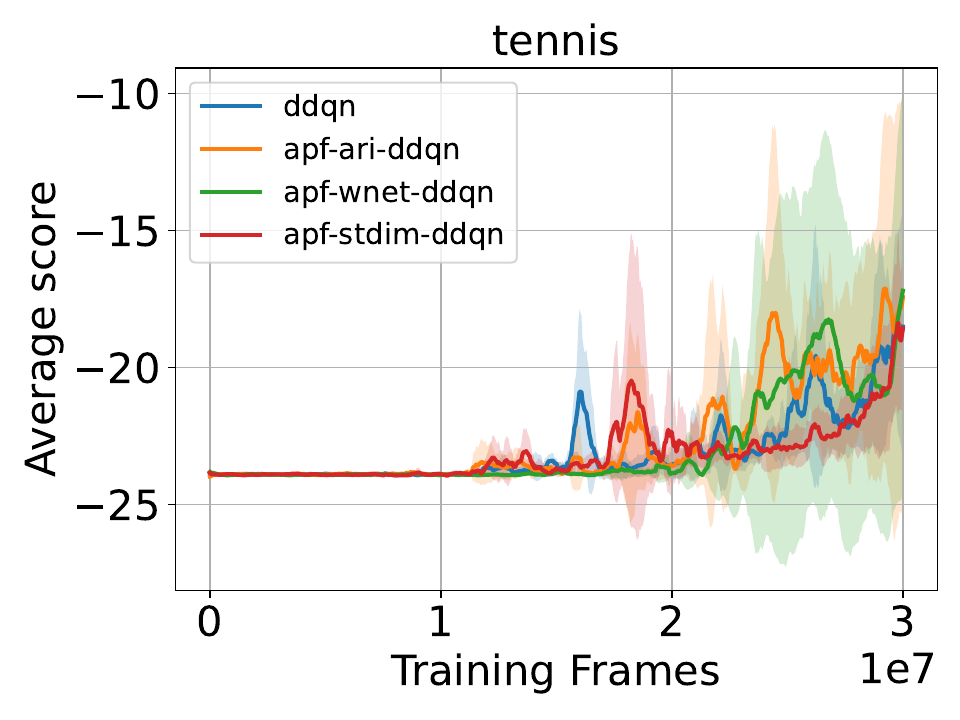}\\
     \includegraphics[width=.3\linewidth]{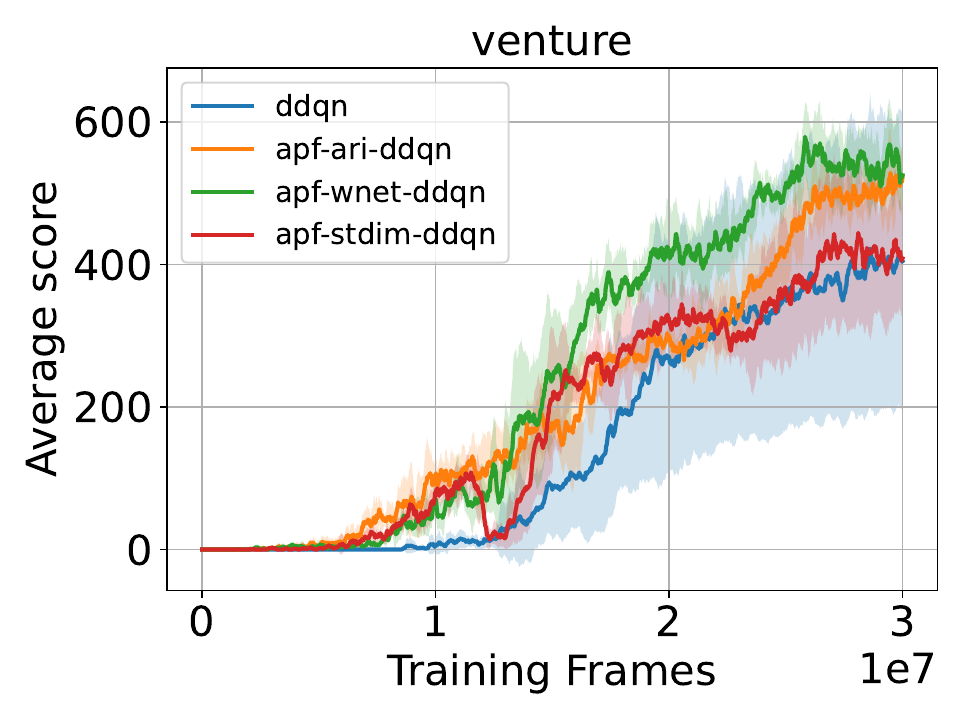}
     \includegraphics[width=.3\linewidth]{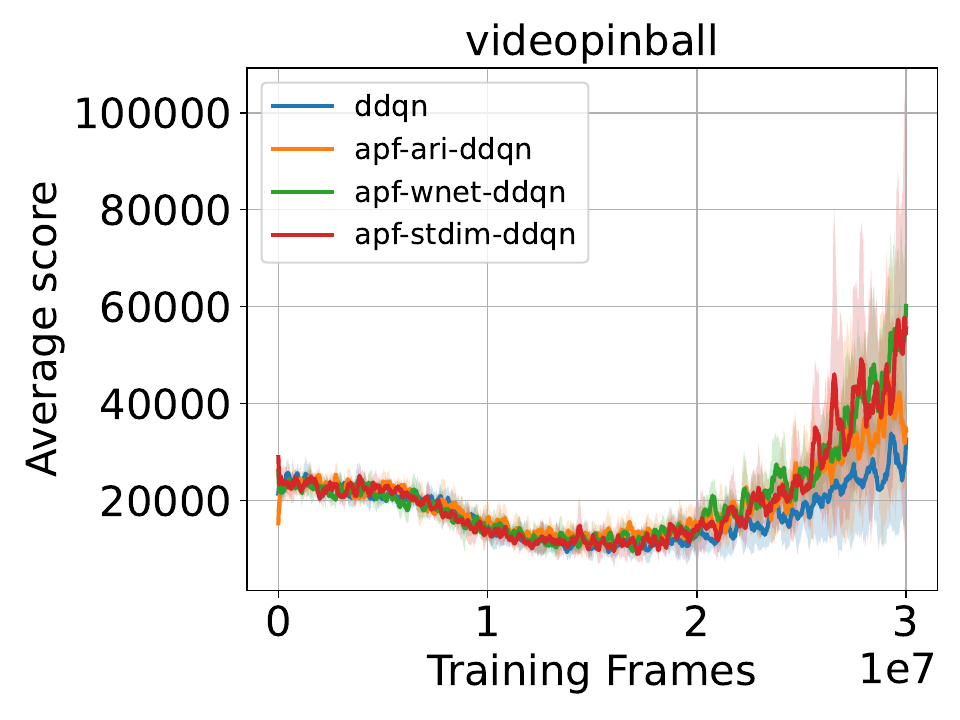}
     \includegraphics[width=.3\linewidth]{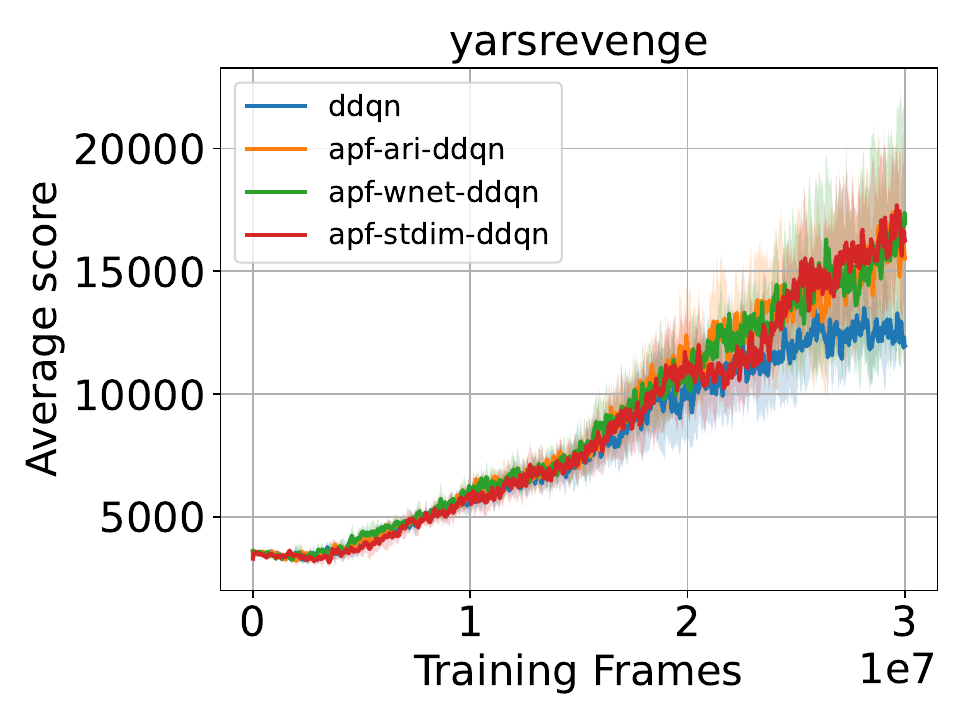}

\label{fig:res_all}
\end{adjustwidth} 
\end{figure}

\subsection*{S2 Fig.}

{\bf The state representation effect of W-Net in all $20$ Atari games.} In each figure, the four rows from top to bottom are visualizations of the input layer, the \texttt{out-u1} layer, the \texttt{residual} layer, and the \texttt{out-u2} layer of the W-Net, respectively.

\begin{figure}[htbp]
\begin{adjustwidth}{-2.25in}{0in}
\includegraphics[width=.99\linewidth]{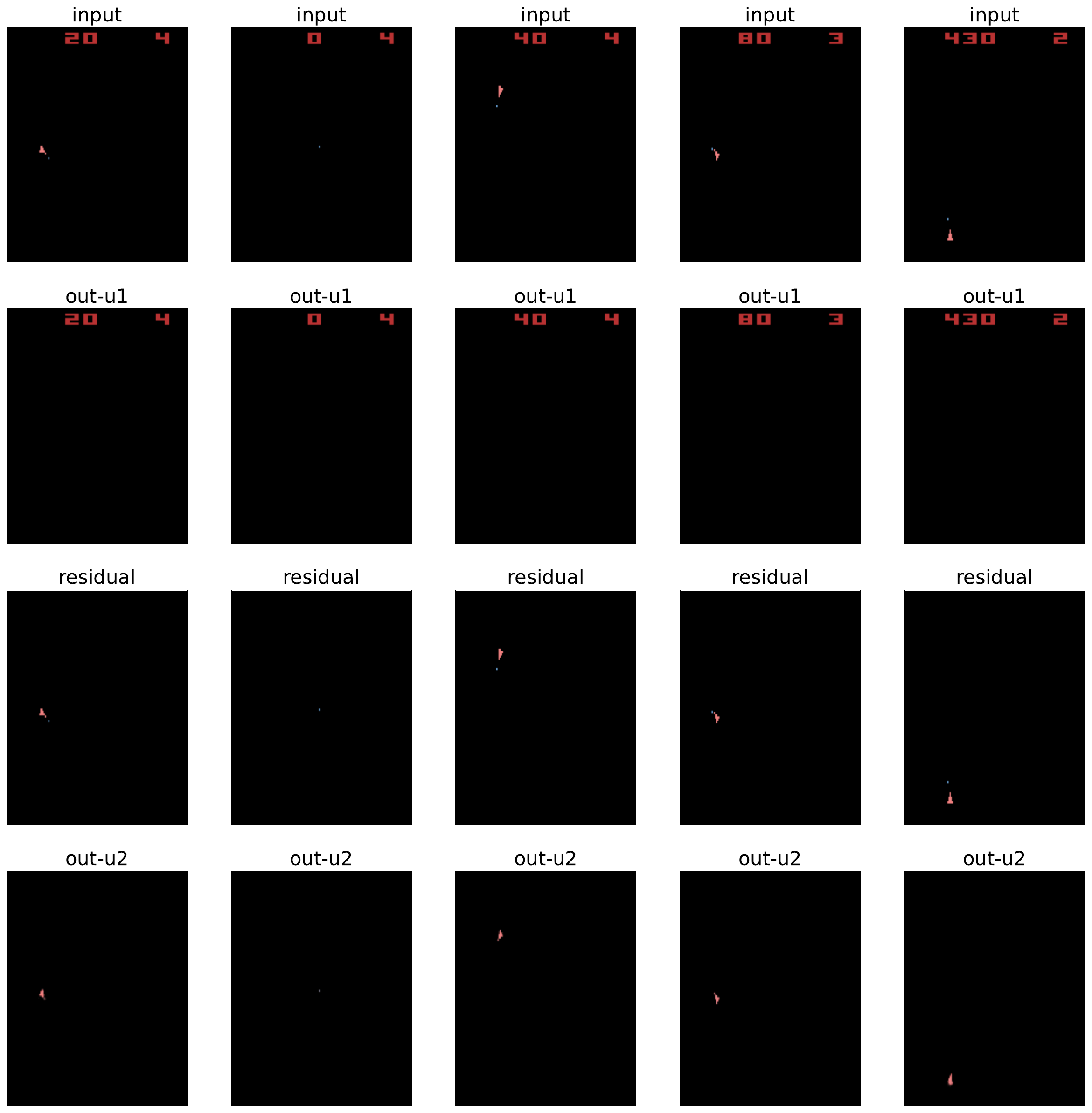}
\caption*{Asteroids}
\end{adjustwidth}
\end{figure}

\begin{figure}[htbp]
\begin{adjustwidth}{-2.25in}{0in}
\centering
\includegraphics[width=.99\linewidth]{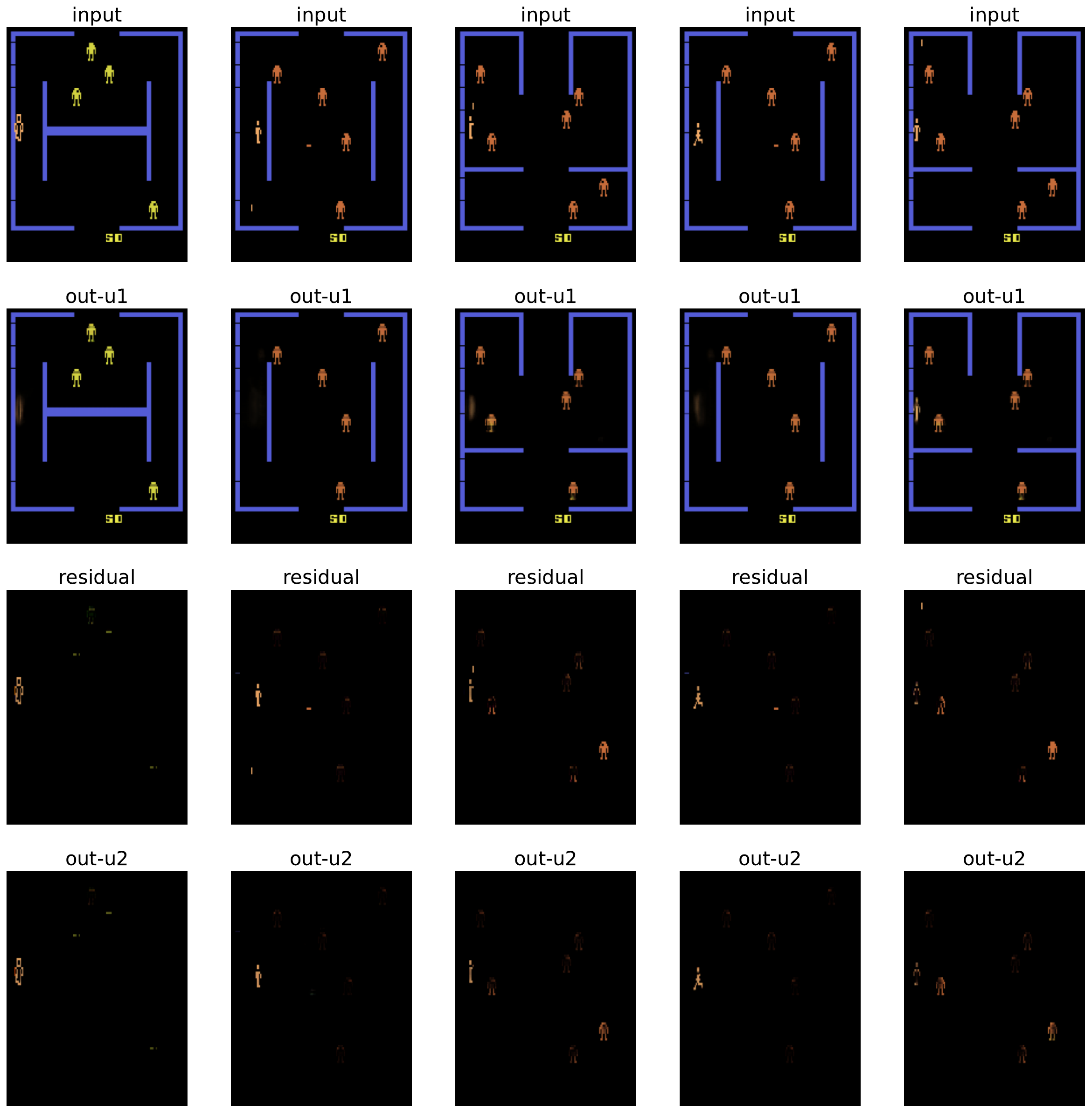}
\caption*{Berzerk}
\end{adjustwidth}
\end{figure}

\begin{figure}[htbp]
\begin{adjustwidth}{-2.25in}{0in}
\centering
\includegraphics[width=.99\linewidth]{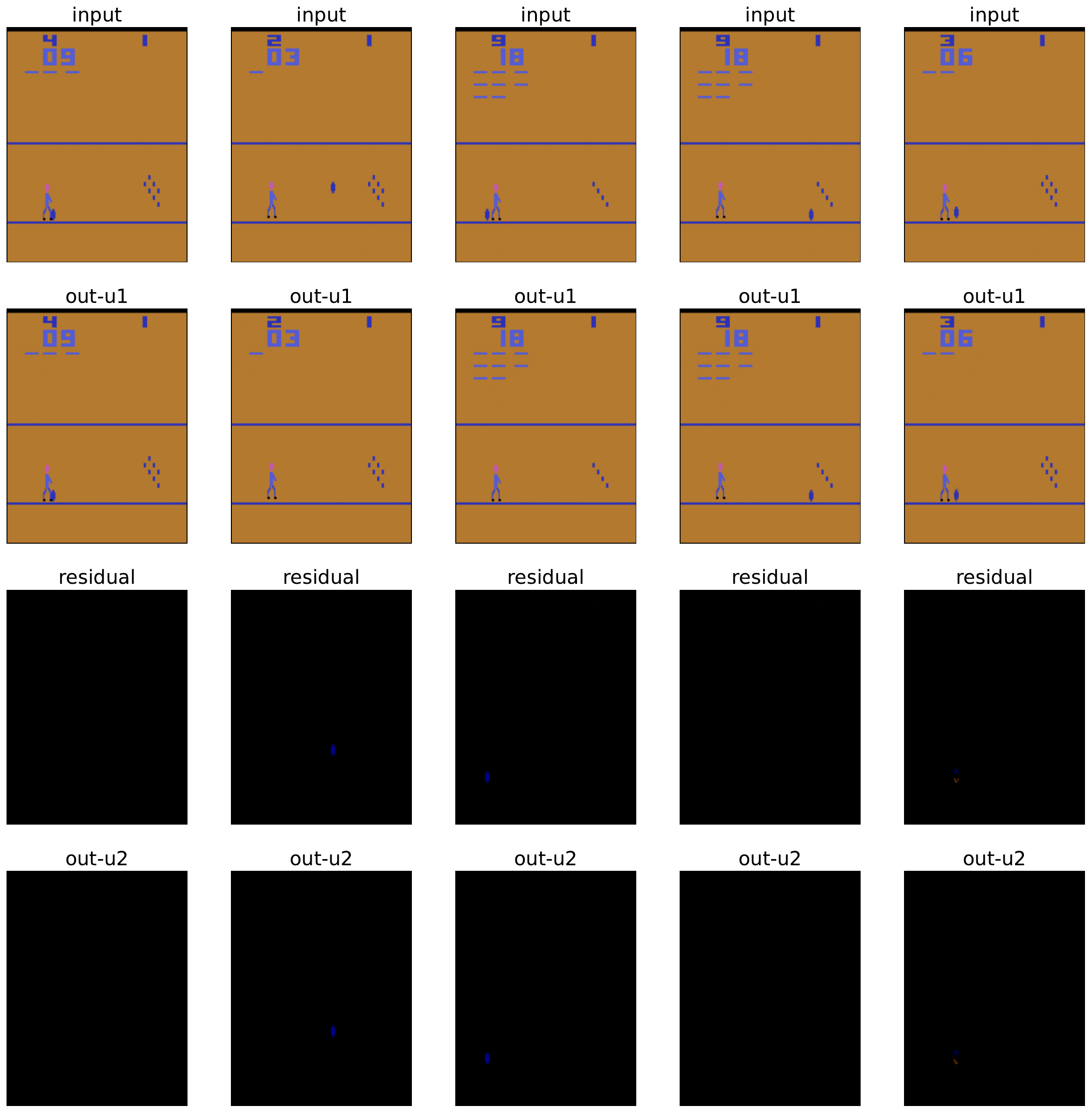}
\caption*{Bowling}
\end{adjustwidth}
\end{figure}

\begin{figure}[htbp]
\begin{adjustwidth}{-2.25in}{0in}
\centering
\includegraphics[width=.99\linewidth]{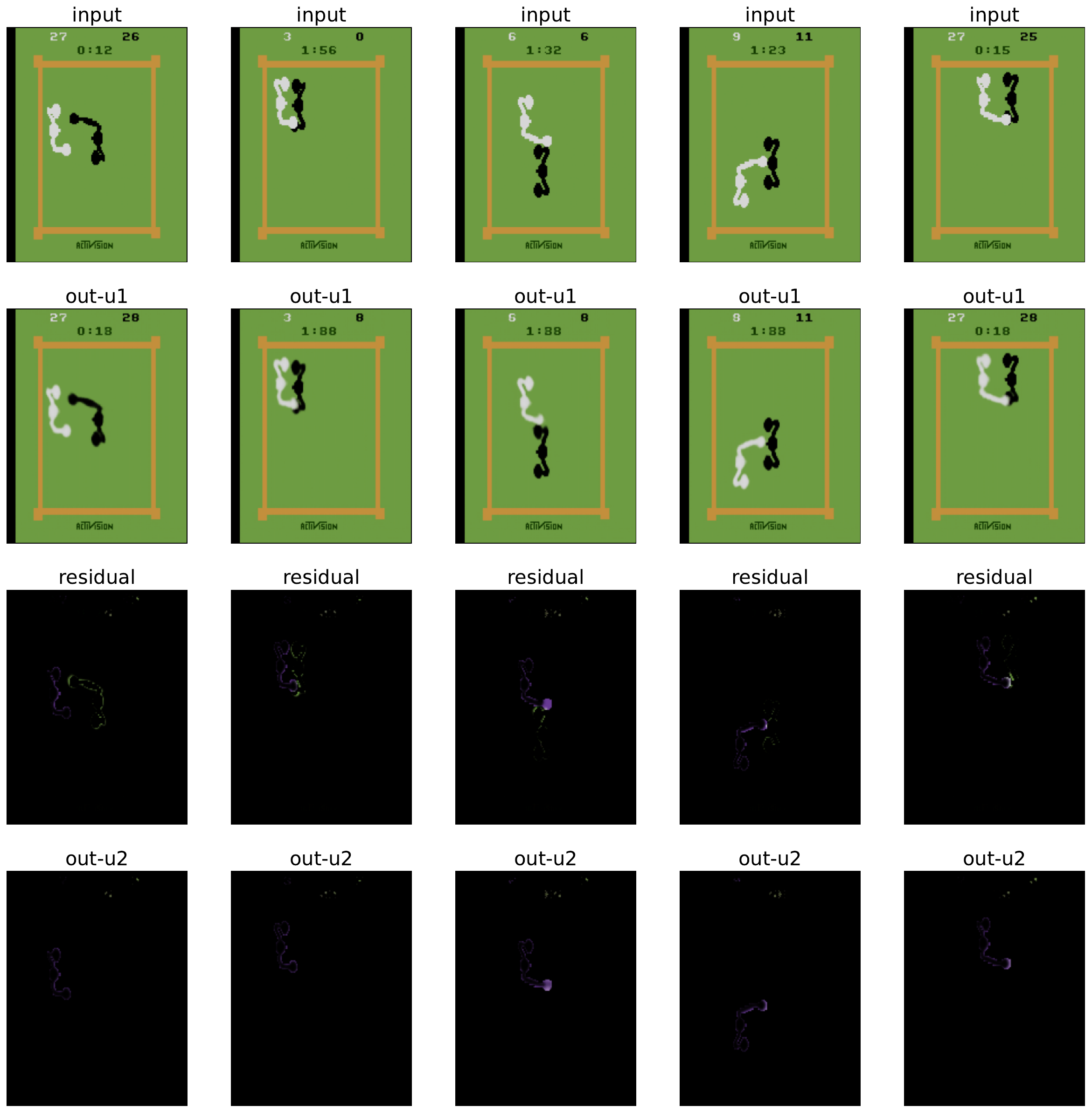}
\caption*{Boxing}
\end{adjustwidth}
\end{figure}

\begin{figure}[htbp]
\begin{adjustwidth}{-2.25in}{0in}
\centering
\includegraphics[width=.99\linewidth]{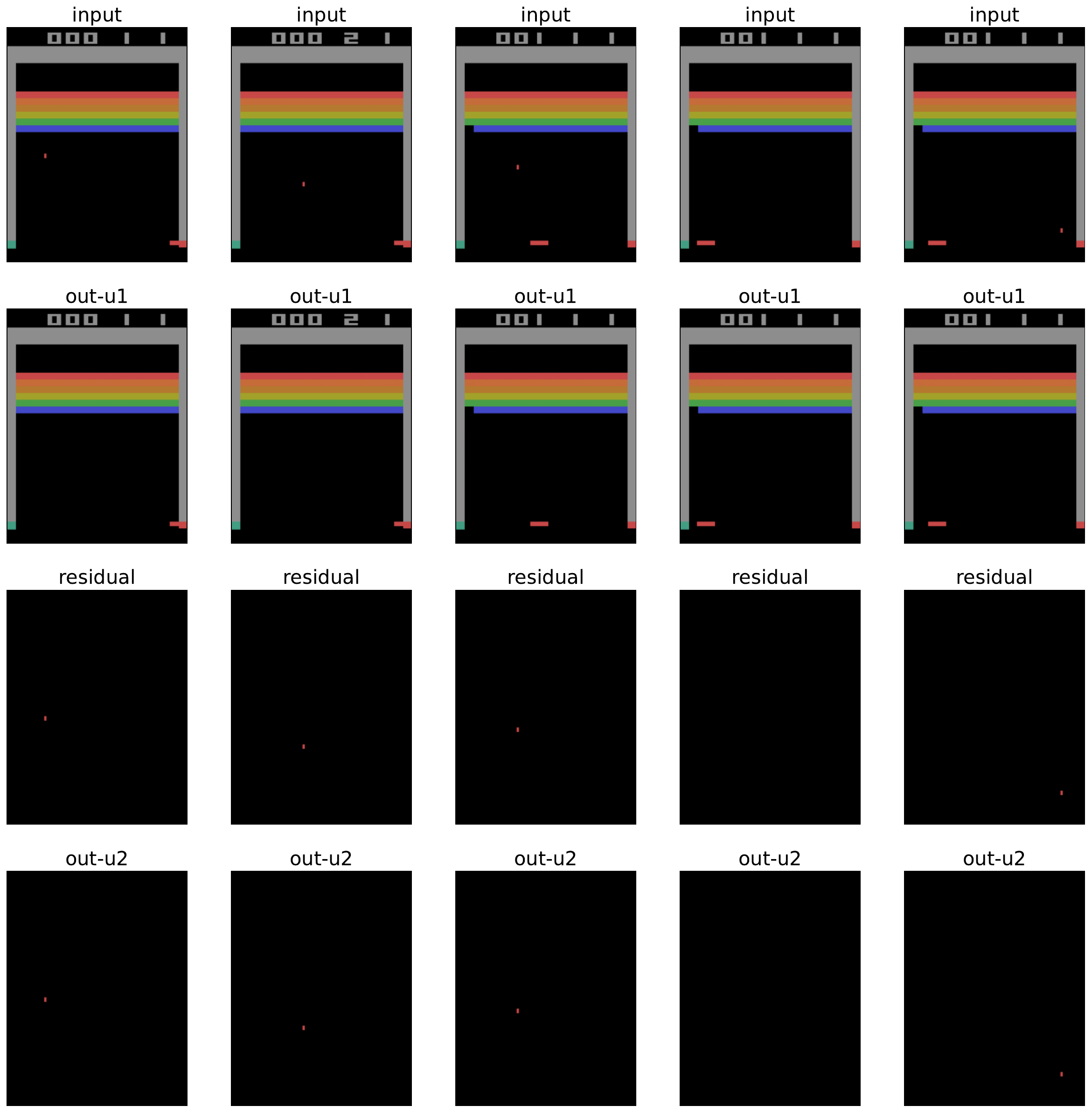}
\caption*{Breakout}
\end{adjustwidth}
\end{figure}

\begin{figure}[htbp]
\begin{adjustwidth}{-2.25in}{0in}
\centering
\includegraphics[width=.99\linewidth]{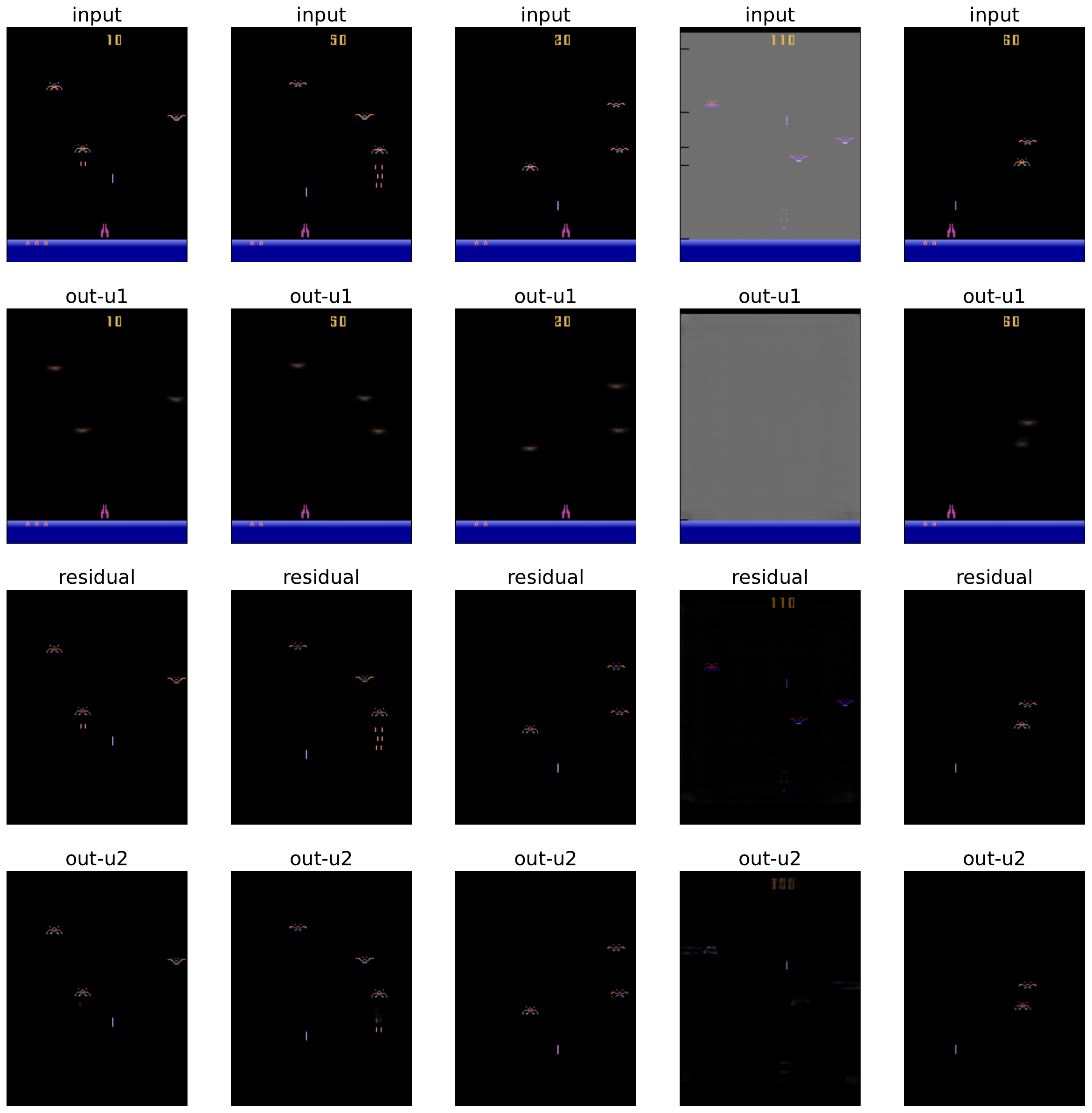}
\caption*{DemonAttack}
\end{adjustwidth}
\end{figure}

\begin{figure}[htbp]
\begin{adjustwidth}{-2.25in}{0in}
\centering
\includegraphics[width=.99\linewidth]{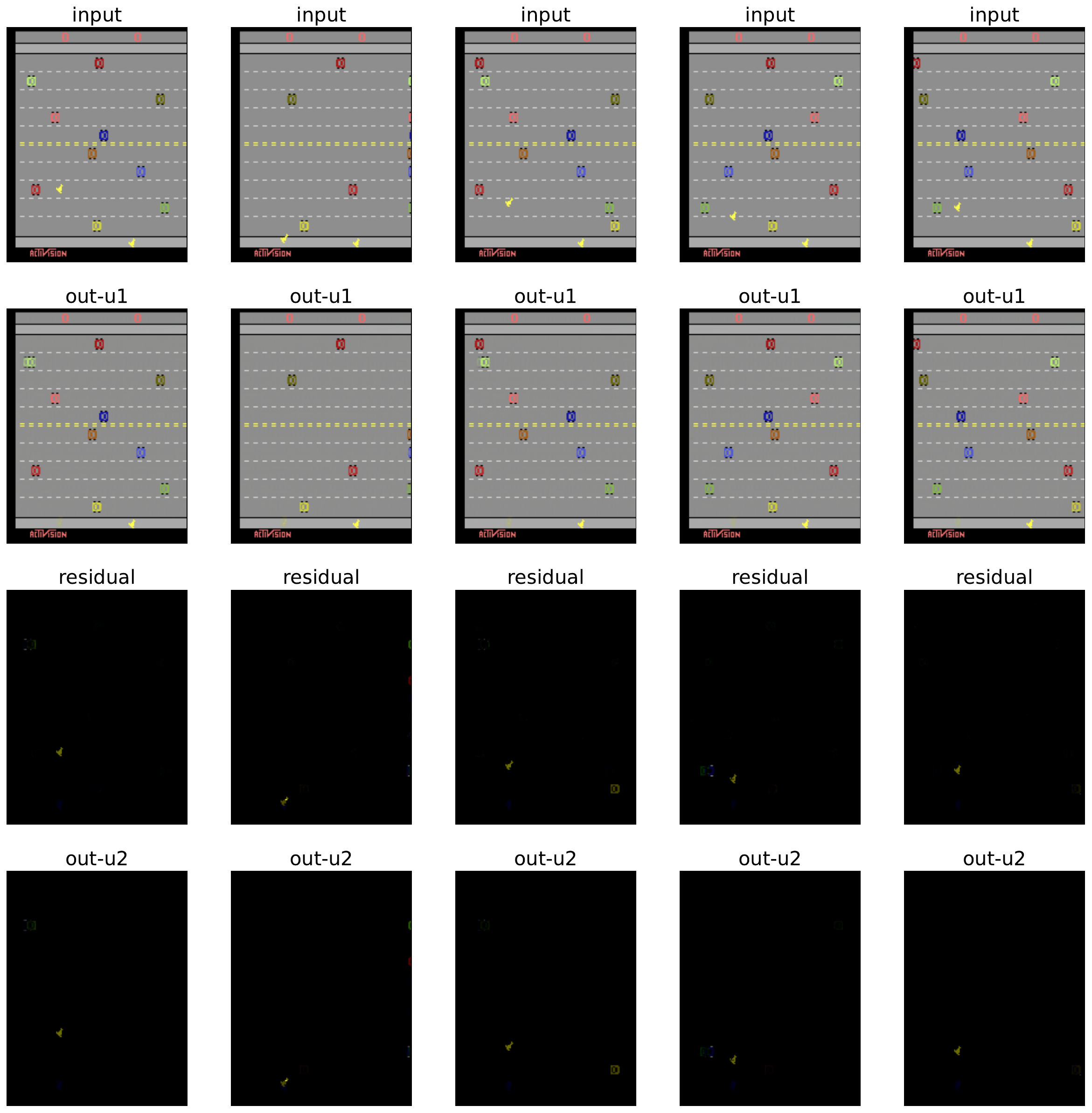}
\caption*{Freeway}
\end{adjustwidth}
\end{figure}

\begin{figure}[htbp]
\begin{adjustwidth}{-2.25in}{0in}
\centering
\includegraphics[width=.99\linewidth]{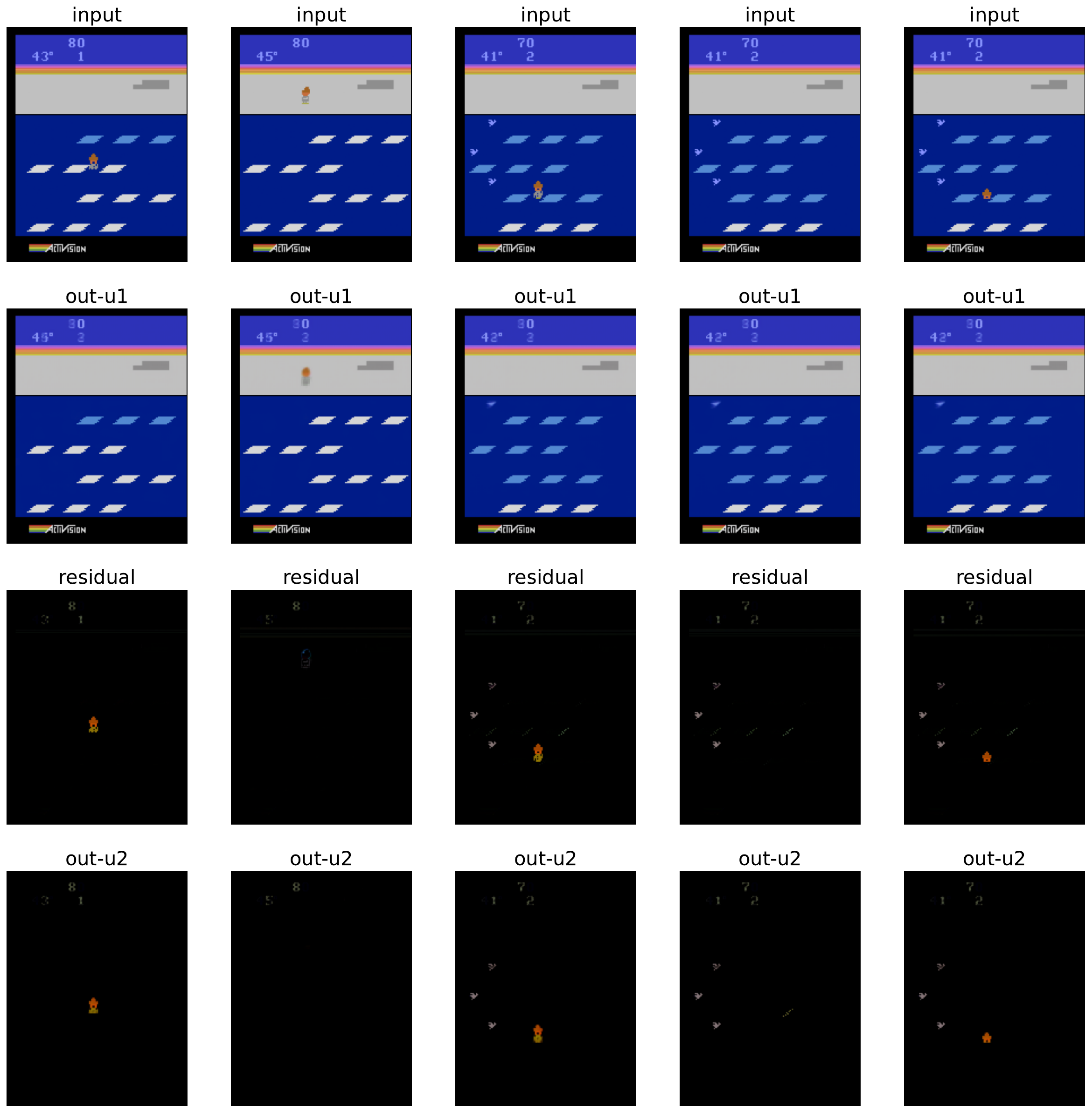}
\caption*{Frostbite}
\end{adjustwidth}
\end{figure}

\begin{figure}[htbp]
\begin{adjustwidth}{-2.25in}{0in}
\centering
\includegraphics[width=.99\linewidth]{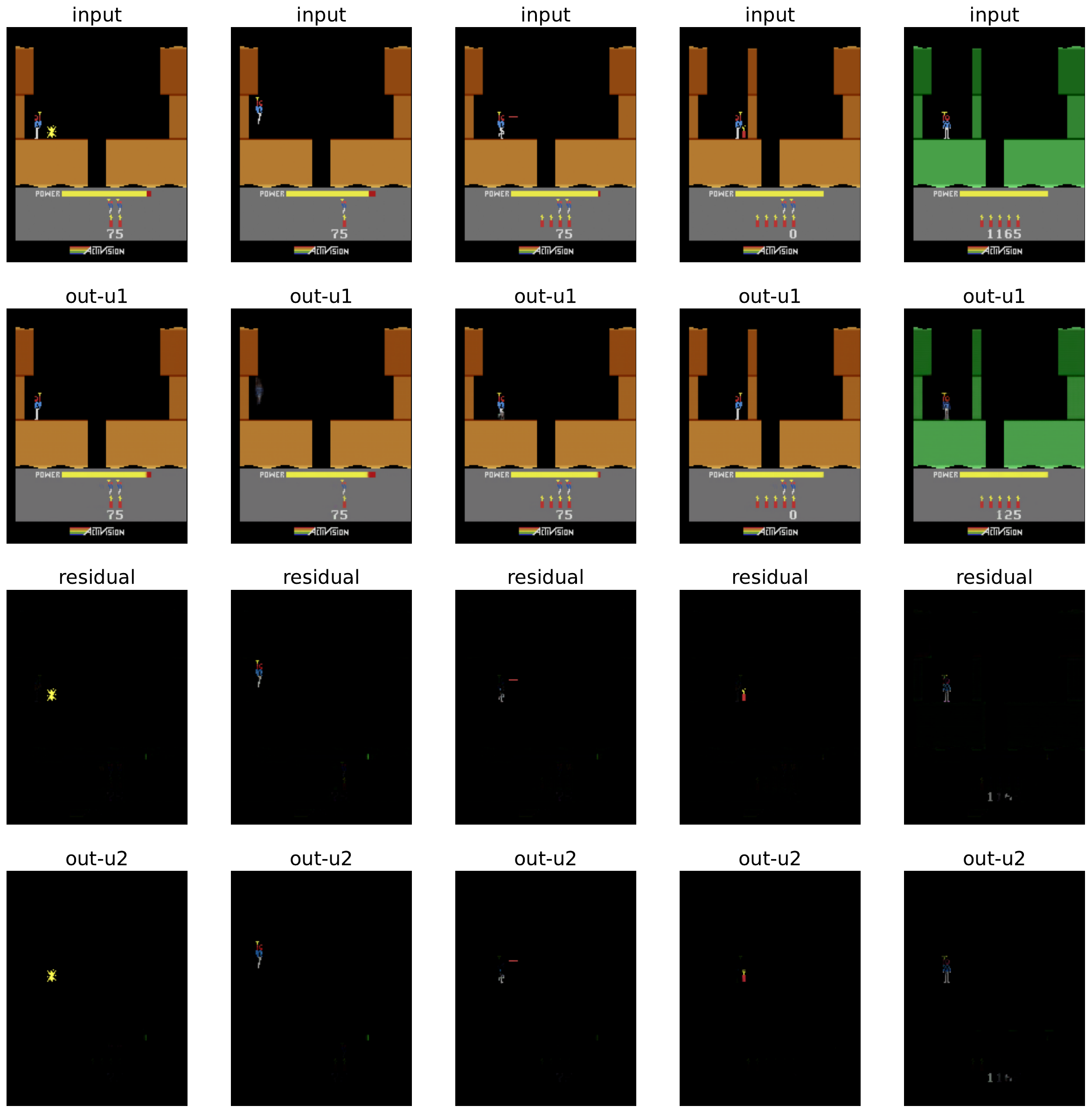}
\caption*{Hero}
\end{adjustwidth}
\end{figure}

\begin{figure}[htbp]
\begin{adjustwidth}{-2.25in}{0in}
\centering
\includegraphics[width=.99\linewidth]{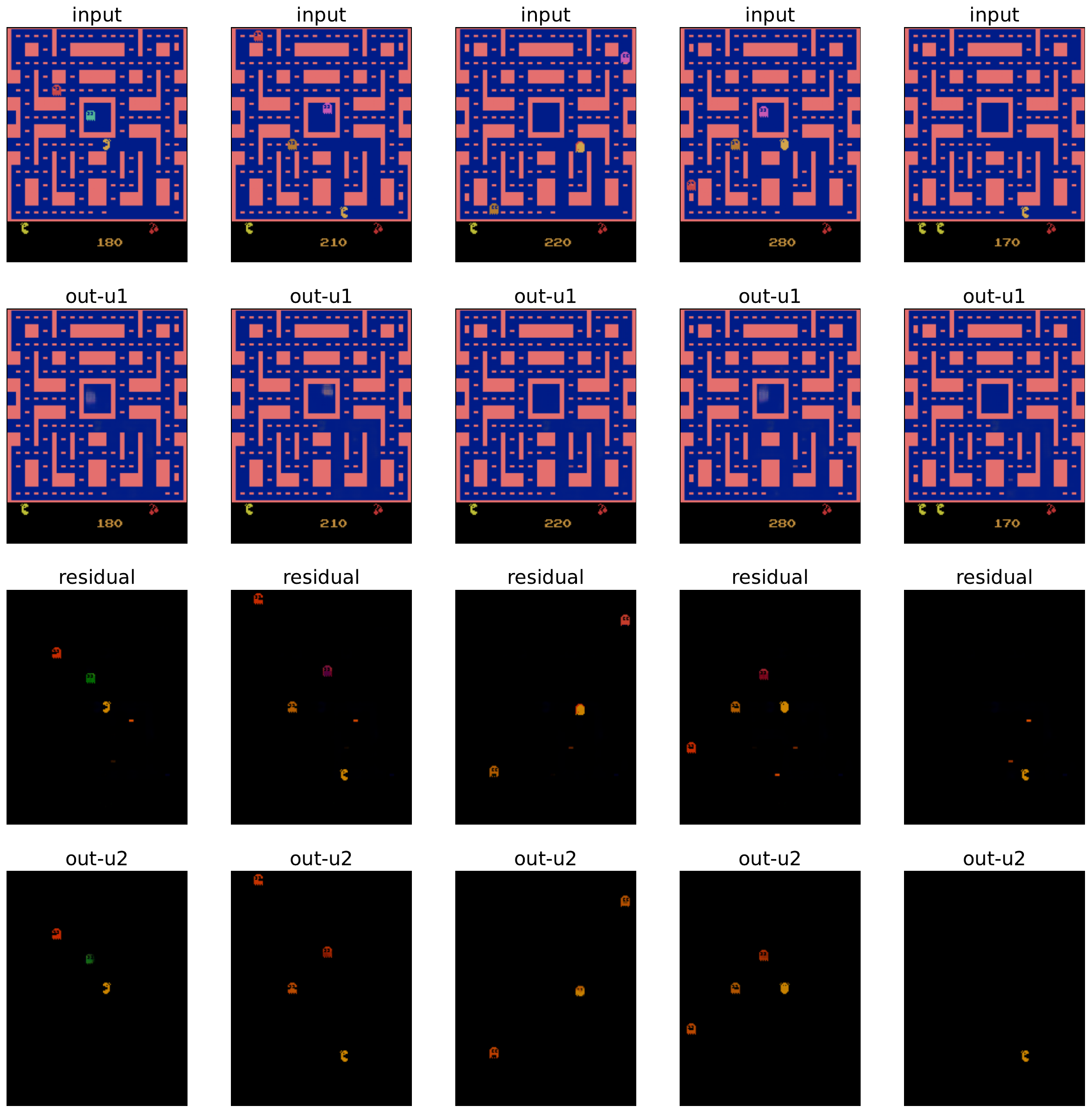}
\caption*{MsPacman}
\end{adjustwidth}
\end{figure}

\begin{figure}[htbp]
\begin{adjustwidth}{-2.25in}{0in}
\centering
\includegraphics[width=.99\linewidth]{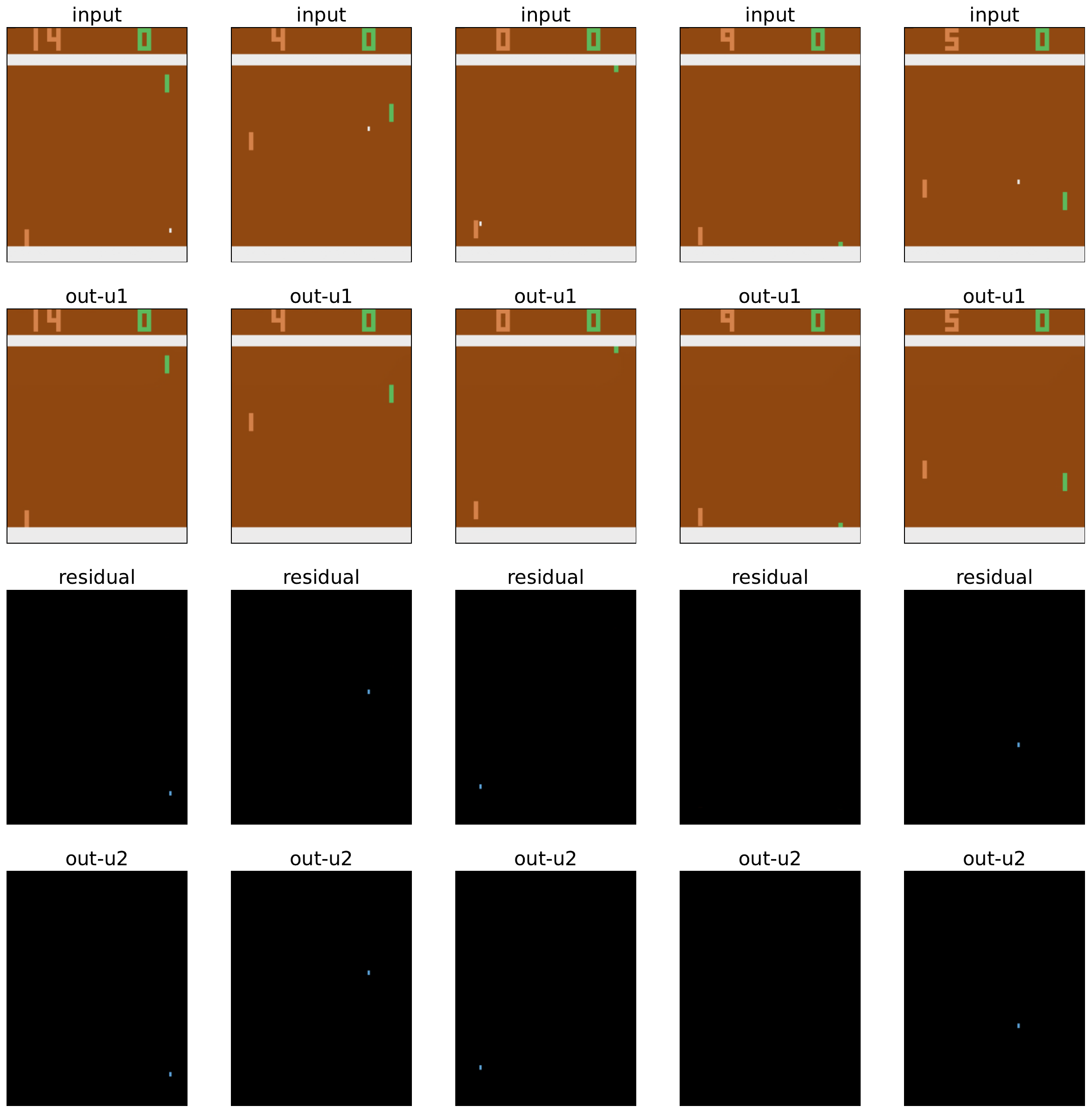}
\caption*{Pong}
\end{adjustwidth}
\end{figure}

\begin{figure}[htbp]
\begin{adjustwidth}{-2.25in}{0in}
\centering
\includegraphics[width=.99\linewidth]{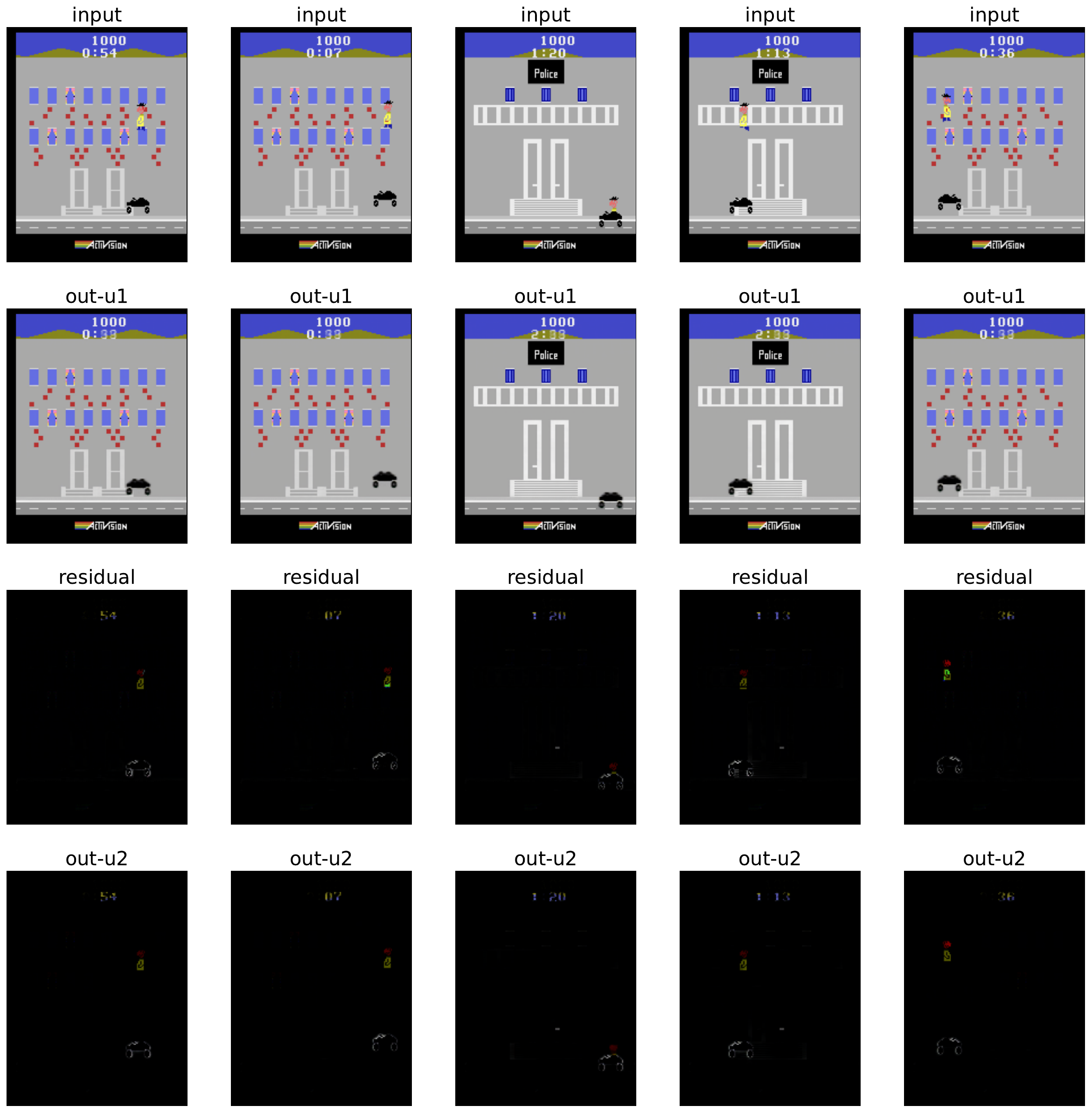}
\caption*{PrivateEye}
\end{adjustwidth}
\end{figure}

\begin{figure}[htbp]
\begin{adjustwidth}{-2.25in}{0in}
\centering
\includegraphics[width=.99\linewidth]{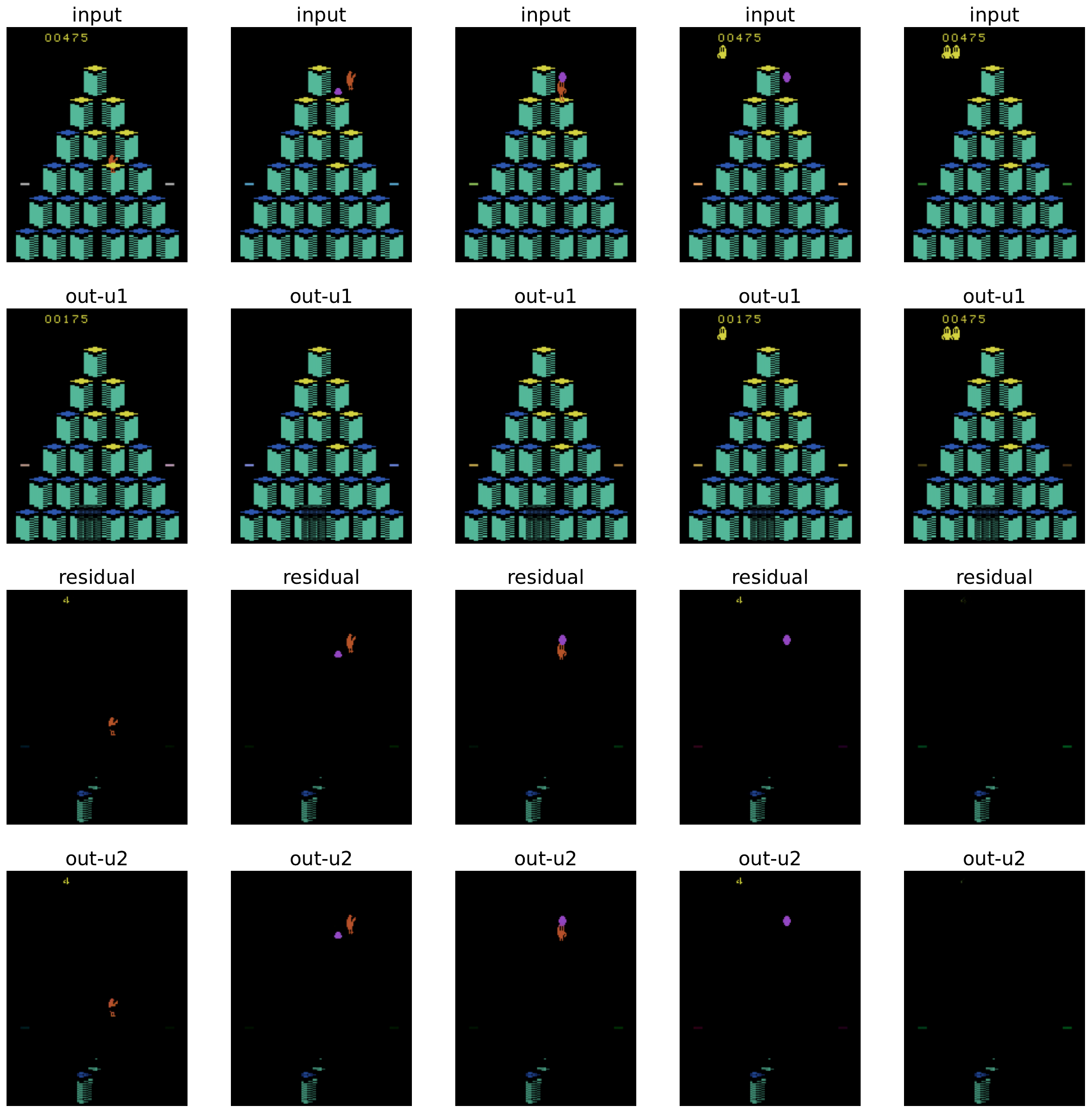}
\caption*{Qbert}
\end{adjustwidth}
\end{figure}

\begin{figure}[htbp]
\begin{adjustwidth}{-2.25in}{0in}
\centering
\includegraphics[width=.99\linewidth]{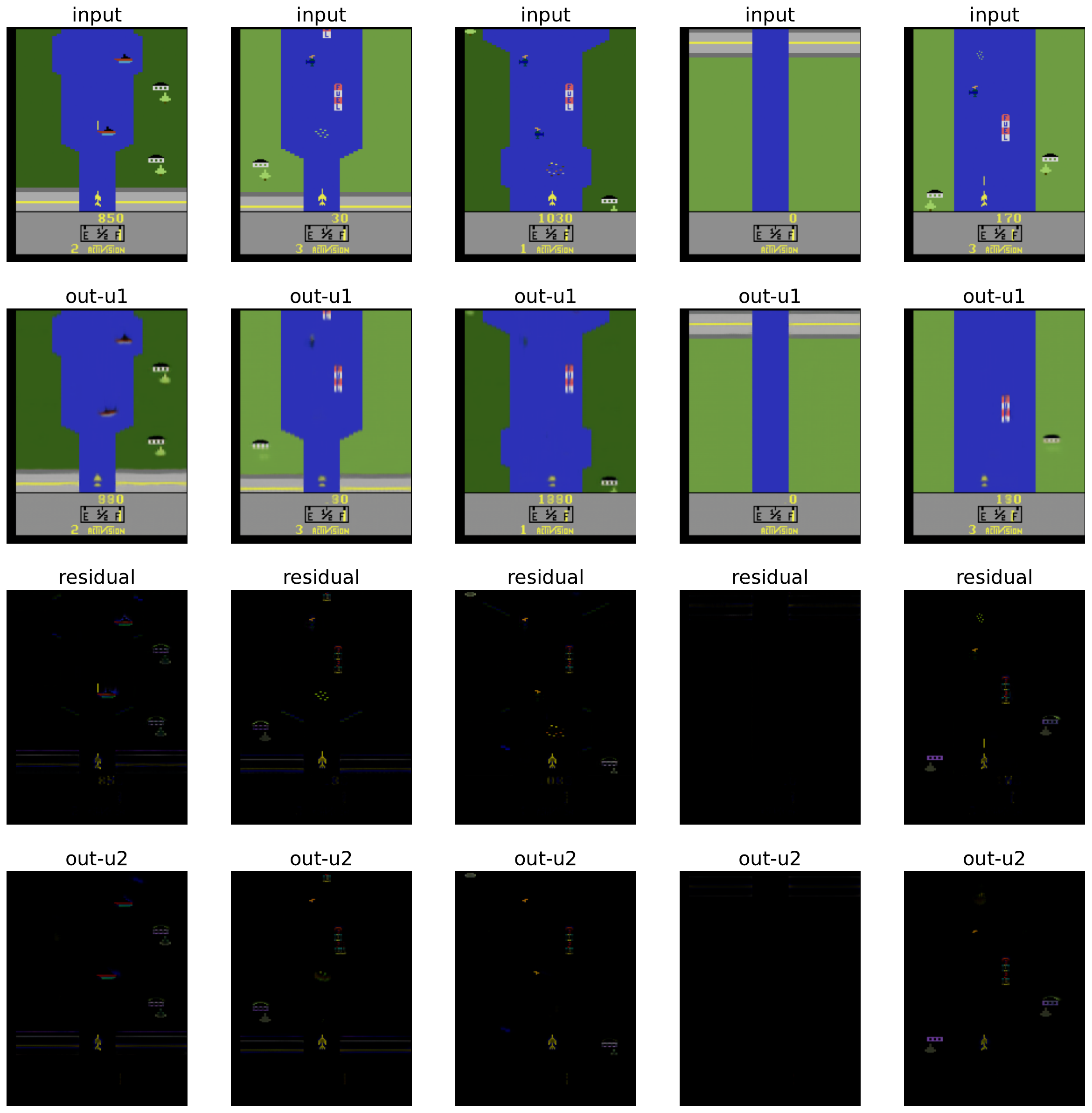}
\caption*{Riverraid}
\end{adjustwidth}
\end{figure}

\begin{figure}[htbp]
\begin{adjustwidth}{-2.25in}{0in}
\centering
\includegraphics[width=.99\linewidth]{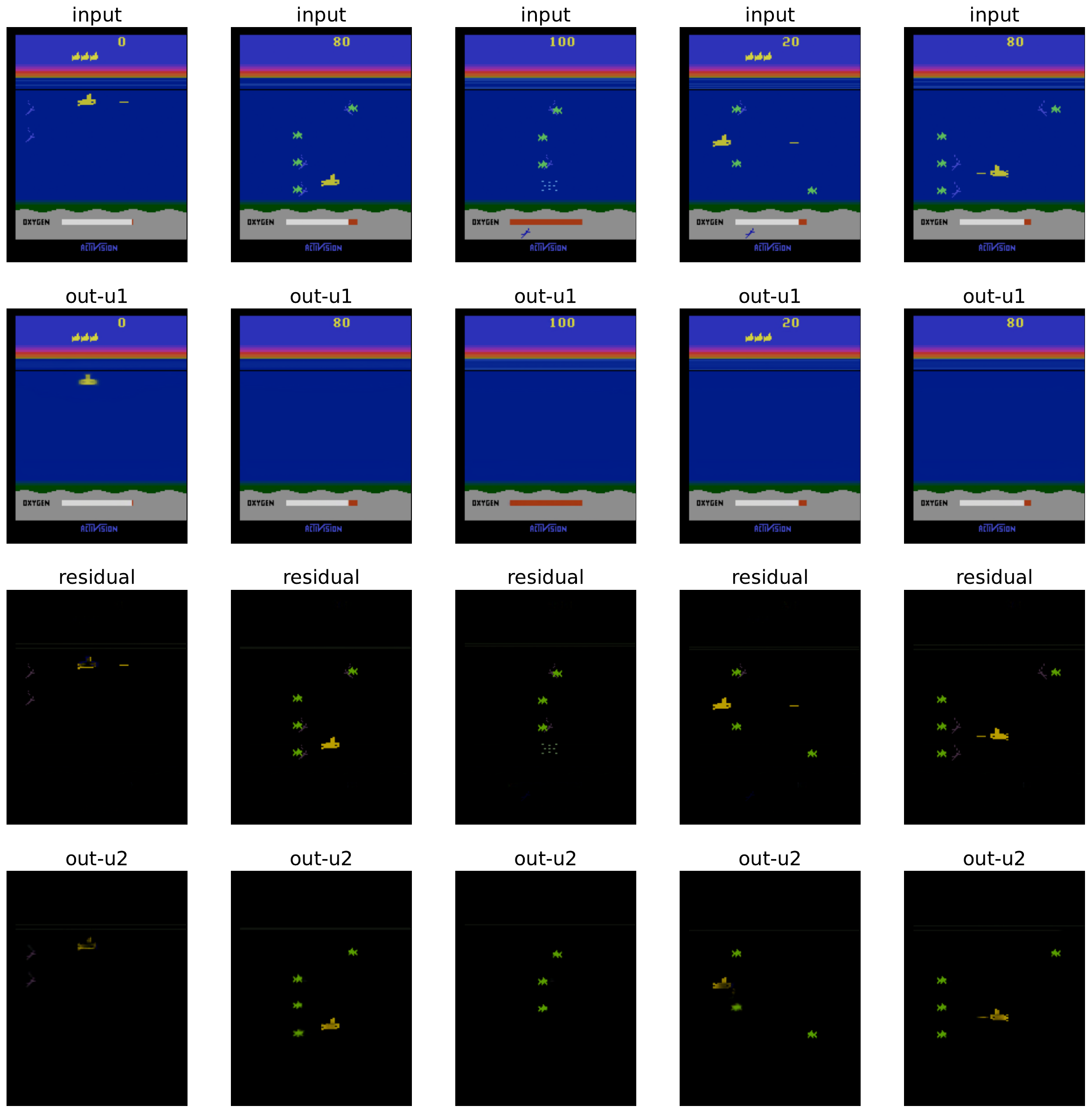}
\caption*{Seaquest}
\end{adjustwidth}
\end{figure}

\begin{figure}[htbp]
\begin{adjustwidth}{-2.25in}{0in}
\centering
\includegraphics[width=.99\linewidth]{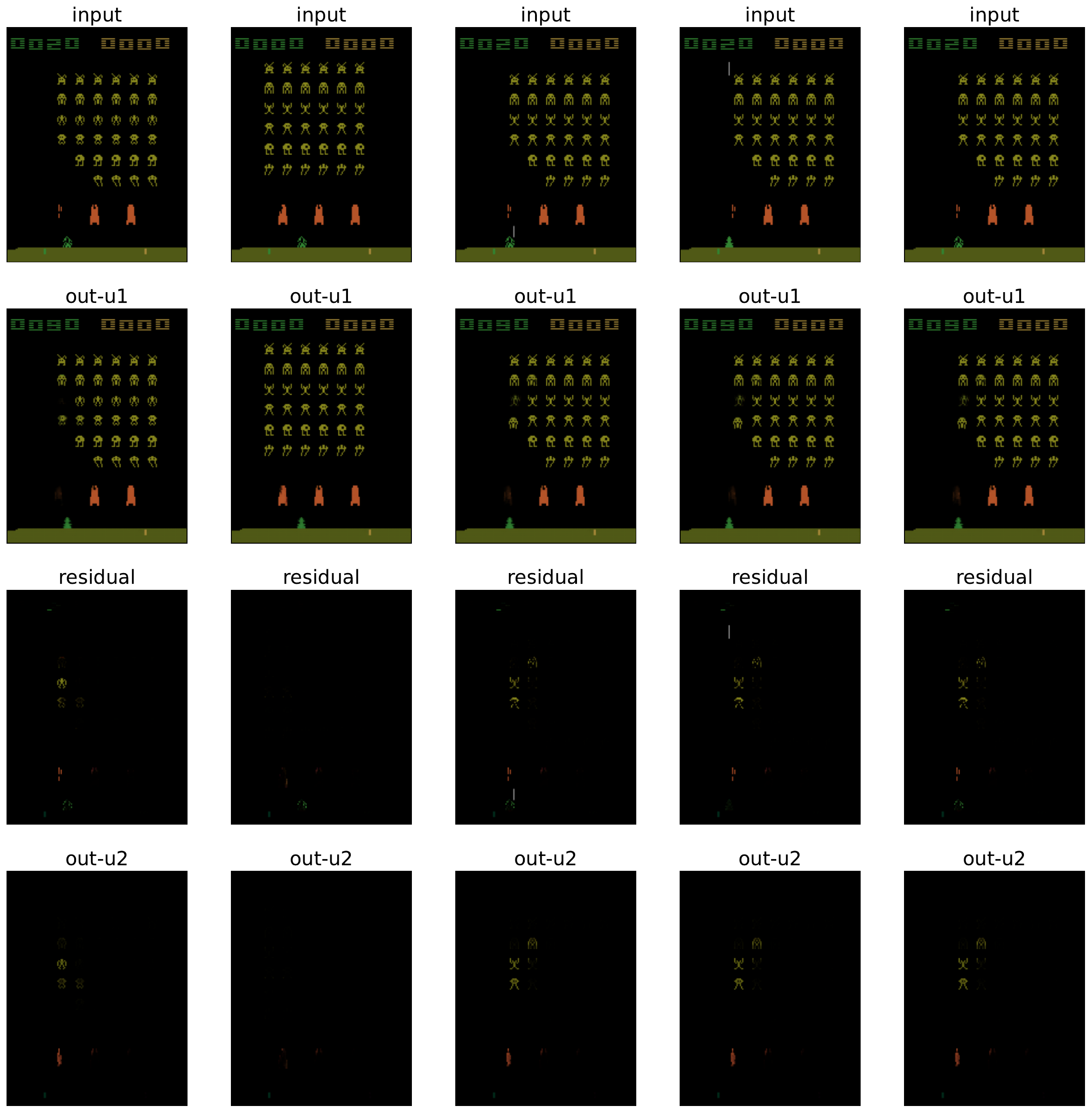}
\caption*{SpaceInvaders}
\end{adjustwidth}
\end{figure}

\begin{figure}[htbp]
\begin{adjustwidth}{-2.25in}{0in}
\centering
\includegraphics[width=.99\linewidth]{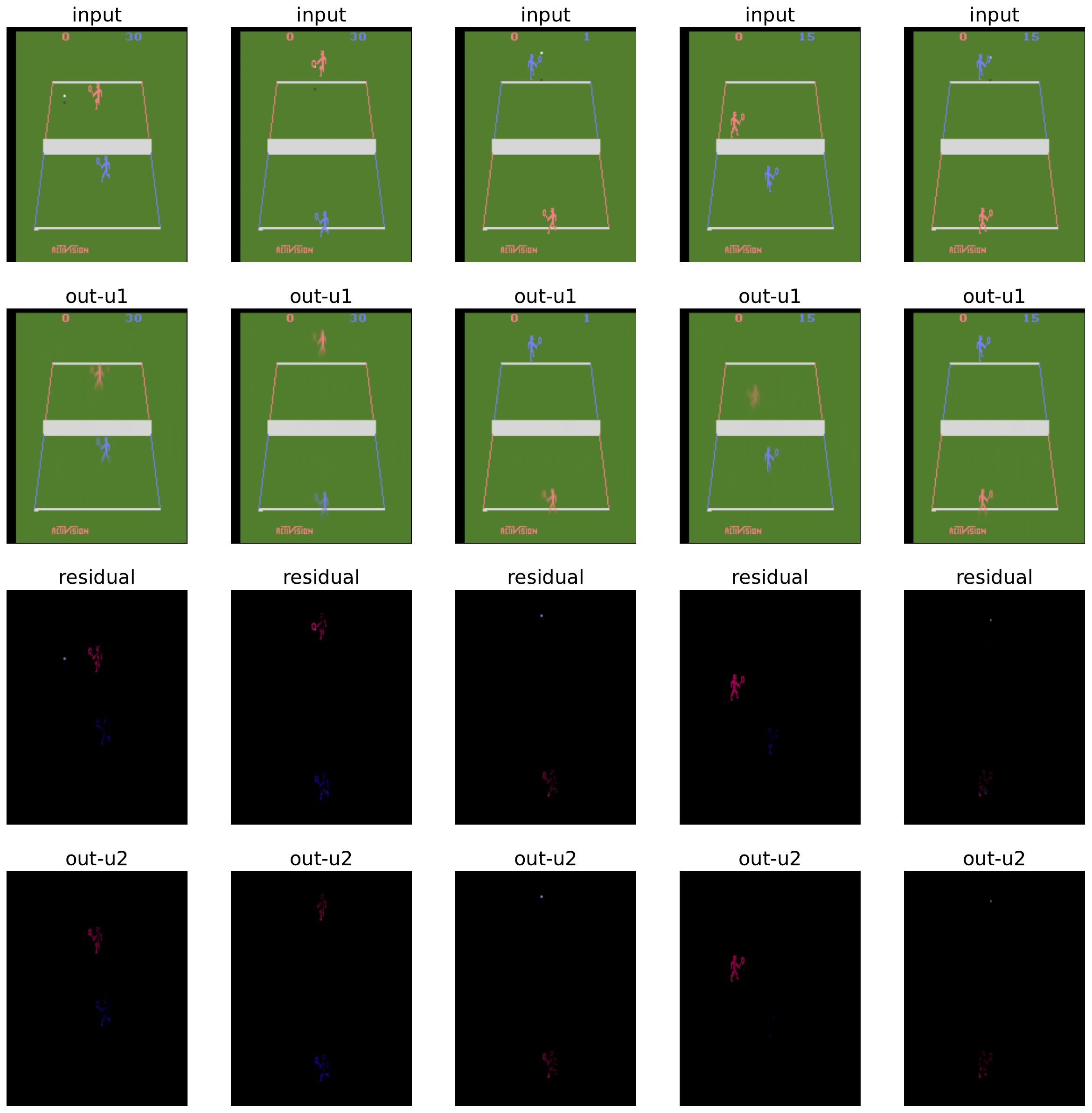}
\caption*{Tennis}
\end{adjustwidth}
\end{figure}
     
\begin{figure}[htbp]
\begin{adjustwidth}{-2.25in}{0in}
\centering
\includegraphics[width=.99\linewidth]{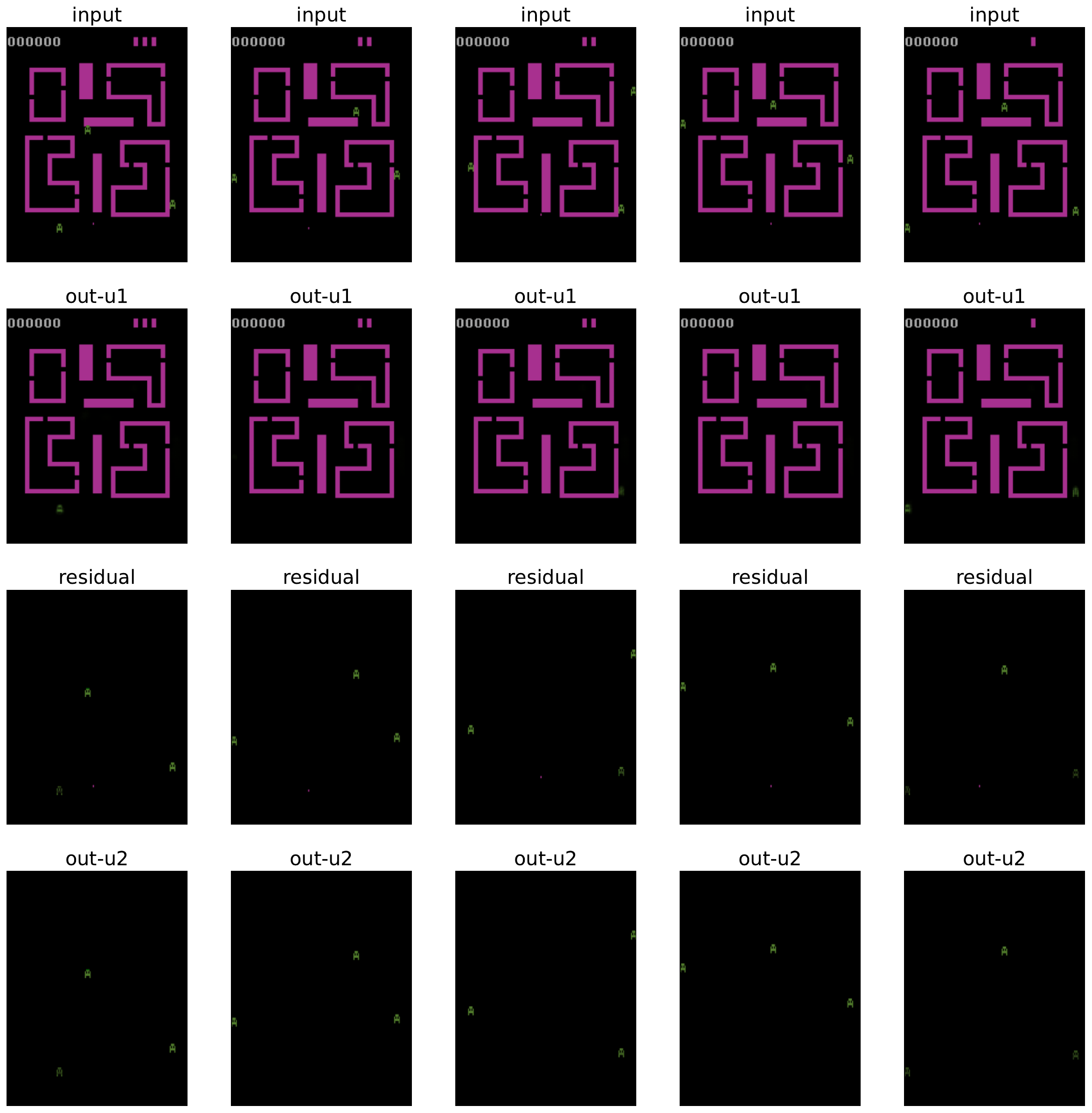}
\caption*{Venture}
\end{adjustwidth}
\end{figure}

\begin{figure}[htbp]
\begin{adjustwidth}{-2.25in}{0in}
\centering
\includegraphics[width=.99\linewidth]{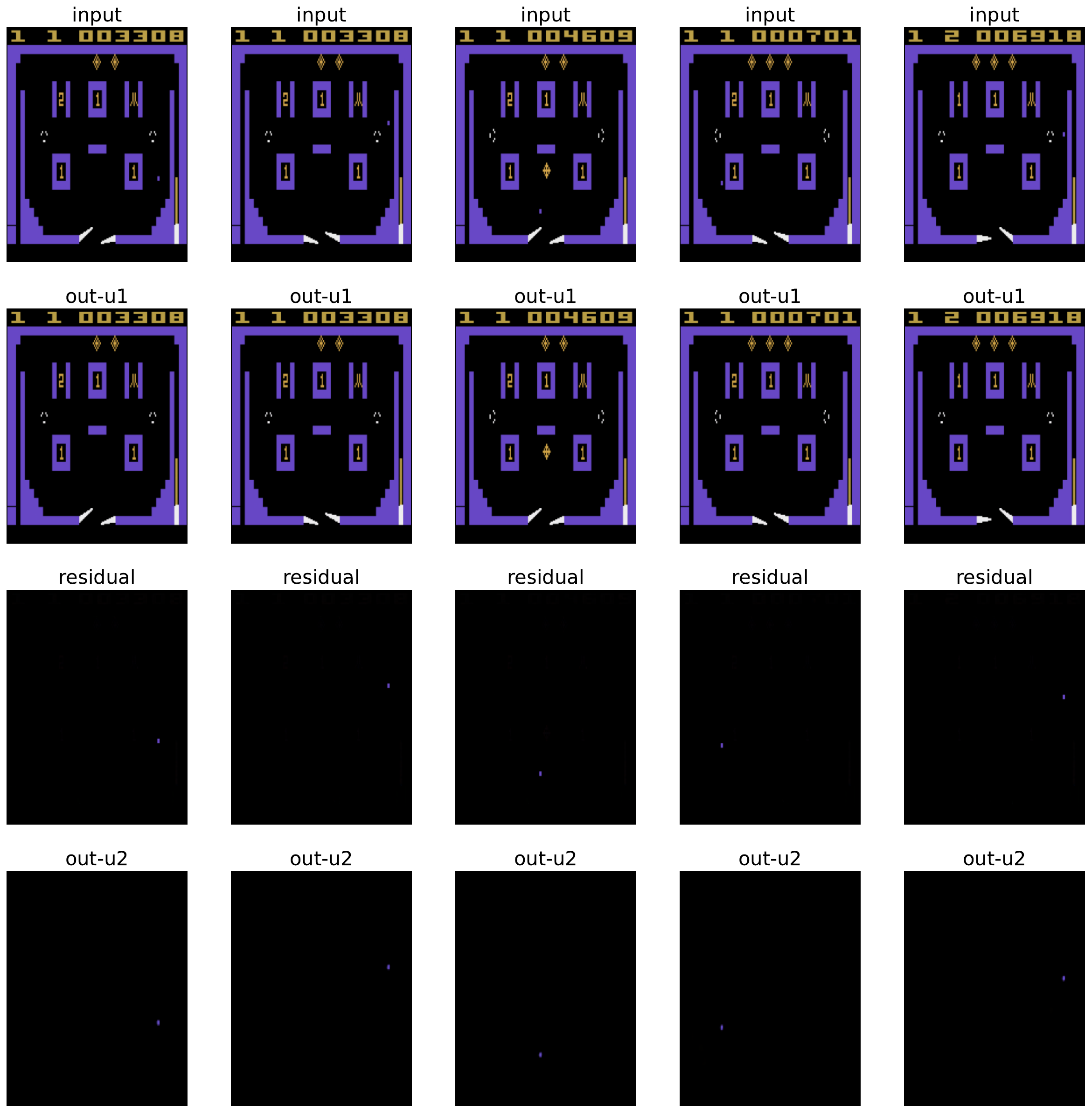}
\caption*{VideoPinball}
\end{adjustwidth}
\end{figure}
     
\begin{figure}[htbp]
\begin{adjustwidth}{-2.25in}{0in}
\centering
\includegraphics[width=.99\linewidth]{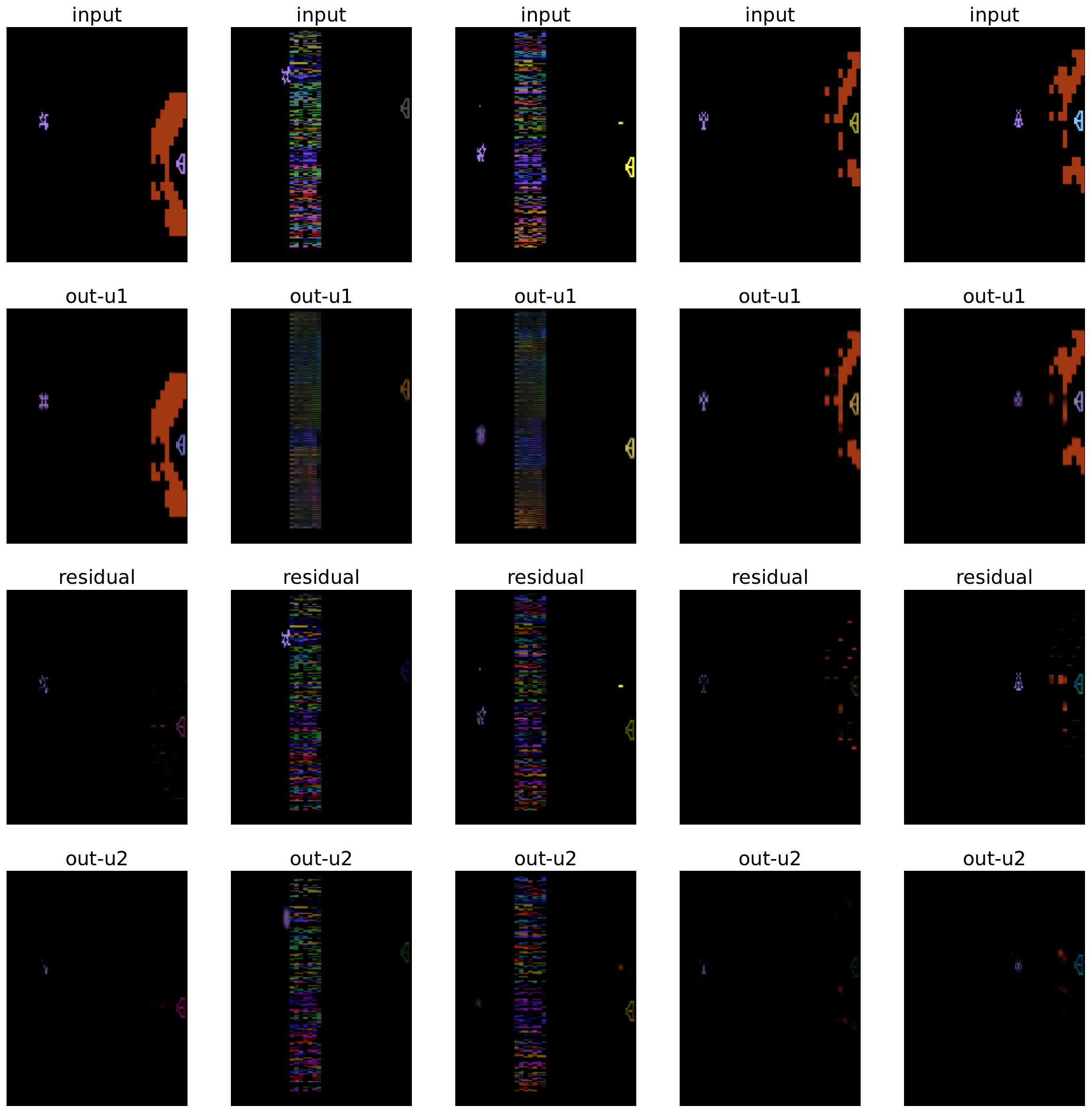}
\caption*{YarsRevenge}
\end{adjustwidth}
\end{figure}

\end{document}